\documentclass[11pt]{article}
\usepackage{amsmath,amsfonts,amssymb}
\usepackage{graphicx,psfrag,epsf,url}
\usepackage{enumerate,color,centernot}
\usepackage{natbib}
\usepackage{adjustbox,booktabs,arydshln}
\usepackage{caption}
\usepackage{subcaption}

\usepackage{graphicx}
\newcommand{\indep}{\rotatebox[origin=c]{90}{$\models$}}

\usepackage[letterpaper,margin=1in]{geometry}
\usepackage{algorithm}
\usepackage{algorithmic}

\newcommand{\bH}{{\boldsymbol H}}

\newcommand{\mX}{\mathcal{X}}
\newtheorem{theorem}{Theorem}[section]
\newtheorem{lemma}{Lemma}[section]

\newtheorem{corollary}[theorem]{Corollary}

\newtheorem{remark}{Remark}
\newtheorem{assumption}{Assumption}
\newenvironment{proof}{\trivlist\item[\hskip \labelsep{\sc Proof:}]}
 {\unskip\nobreak\ \lower.3ex\hbox{$\Box$}\endtrivlist}

\begin{document}

\title{A New Paradigm for Generative Adversarial Networks based on Randomized Decision Rules}

\author{Sehwan Kim, Qifan Song, and Faming Liang
\thanks{To whom correspondence should be addressed: Faming Liang (email: fmliang@purdue.edu).
  S. Kim, Q. Song and F. Liang are with Department of Statistics,
  Purdue University, West Lafayette, IN 47907. }}

\maketitle

\begin{abstract}
    The Generative Adversarial Network (GAN) was recently introduced in the literature as a novel machine learning method for training generative models. It has many applications in statistics such as nonparametric clustering and nonparametric conditional independence tests. However, training the GAN is notoriously difficult due to the issue of mode collapse, which refers to the lack of diversity among generated data.
 In this paper, we identify the reasons why the GAN suffers from this issue, and to address it, we propose a new formulation for the GAN based on randomized decision rules.
In the new formulation, the discriminator converges to a fixed point while the generator converges to a distribution at the Nash equilibrium. 
\textcolor{black}{
We propose to train the GAN by an empirical Bayes-like method by treating the discriminator as a hyper-parameter of the posterior distribution of the generator. Specifically, we simulate generators from its posterior distribution conditioned on the discriminator using a stochastic gradient Markov chain Monte Carlo (MCMC) algorithm, and update the discriminator using stochastic gradient descent along with simulations of the generators.} \textcolor{black}{We establish convergence of the proposed method to the Nash equilibrium.} \textcolor{black}{Apart from image generation, we apply the proposed method to nonparametric clustering and nonparametric conditional independence tests.} A portion of the numerical results is presented in the supplementary material. 
   \vspace{2mm}
   
\end{abstract}

\section{Introduction}

The Generative Adversarial Network (GAN) \citep{Goodfellow2014} provides a novel way for 
training generative models which seek to generate new data with the same statistics as the training data. Other than image generation, the GAN has been used in many nonparametric statistical tasks, such as clustering \citep{Mukher19clustergan},  conditional independent test \citep{bellot2019CItest}, and density estimation \citep{Singh2018NonparametricDE, LiuWong2021GANdensity}. In this paper, we call the training data real samples, and those generated by the GAN fake samples.

In its original design, the GAN is trained by competing two neural networks, namely
generator and discriminator, in a game. However, due to the instability issues such as 
{\it mode collapse} (i.e., lack of diversity among fake samples), 
non-convergence, and vanishing or exploding gradients, 
the GAN is notoriously hard to train \citep{Wiatrak2019StabilizingGA}. 
\textcolor{black}{In this paper, we identify the reasons} why the GAN suffers from the mode collapse issue: {\it (i) The GAN evaluates fake samples at an individual level, lacking a mechanism for enhancing the diversity of fake samples; and (ii) the GAN tends to get trapped into a sub-optimal solution, lacking a mechanism for escaping from local traps} (see Remark \ref{rem2022a} for more explanations). \textcolor{black}{
To address this issue, we propose a new formulation for the GAN based on randomized decision rules.} 
\textcolor{black}{In this formulation, 
the similarity between the fake and real samples can be evaluated at the population level;   
and the generator is simulated from its posterior distribution conditioned on the discriminator using a stochastic gradient MCMC algorithm, thereby mitigating the difficulty  of  getting trapped in local optima. }

{\bf Our contribution.} The main contribution of this paper is three-fold: (i) we have provided a new formulation for the GAN based on statistical randomized decision theory, which allows the mode collapse issue to be fully addressed; 
(ii) we have proposed a training algorithm associated with the new formulation, and shown that its convergence to the Nash equilibrium is asymptotically guaranteed, or said differently, the proposed algorithm is immune to mode collapse as the number of iterations becomes large; 
(iii) we have developed a \textcolor{black}{Kullback-Leibler divergence-based prior} for the generator, which enhances the diversity of fake samples   and 
further strengthens the effectiveness of the proposed method in overcoming the issue of mode collapse.
The proposed method is tested on image generation, nonparametric clustering,  and nonparametric conditional independence tests (in the supplementary material). Our numerical results \textcolor{black}{suggest} that the proposed method significantly outperforms the existing ones in overcoming the mode collapse issue.


{\bf Related Works.} To tackle the mode collapse issue, a variety of methods have been proposed in the literature, see \cite{Wiatrak2019StabilizingGA} for a recent survey. These methods can be roughly grouped to two categories, namely,  
 metric-based methods 
 and mixture generator methods.

The methods in the first category strive to find a more stable and informative metric to guide the training process of the GAN. For example, \cite{nowozin2016f} suggested
$f$-divergence, \cite{MaoGAN2017} suggested  $\chi^2$-divergence,  \cite{ArjovskyGAN2017} 
suggested Wasserstein distance,  \textcolor{black}{ \cite{Binkowski2018MMD} suggested maximum mean discrepancy, and \cite{che2017mode} and \cite{zhou2019lipschitz} 
suggested some regularized objective functions. } 
As mentioned previously, the GAN evaluates fake samples at the individual level and 
tends to get trapped to a sub-optimal solution. 
Therefore, the mode collapse issue is hard to resolve by employing a different metric unless (i) the objective function is modified such that the similarity between the fake and real samples can be enhanced at the population level, and (ii) a local-trap free optimization algorithm is employed for training. \textcolor{black}{Recently, there has been a growing trend in the literature to incorporate gradient flow into the training of generative models, as explored by \cite{Gao2019flow}. However, achieving this objective is generally considered a challenging task.}

The methods in the second category are to learn a mixture of generators under a probabilistic framework with a similar motivation to this work.  
A non-exhaustive list of such types of methods include  ensemble GAN \citep{Wang2016EnsemblesOG}, \textcolor{black}{ Mix+GAN \citep{Arora2017NND}}, AdaGAN \citep{Tolstikhin2017}, MAD-GAN \citep{Ghosh2018GAN}, MGAN \citep{hoang2018mgan}, Bayesian GAN \citep{saatciwilson2017}, and 
ProbGAN \citep{he2018probgan}. However, many of the methods are not defined in a proper probabilistic framework and, in consequence, the mode collapse issue cannot be overcome with a theoretical guarantee. In ensemble GAN, AdaGAN, MAD-GAN, \textcolor{black}{ Mix+GAN}, and MGAN, only a finite mixture of generators is learned and thus the mode collapse issue cannot be overcome in theory. Bayesian GAN aims to overcome this obstacle by simulating 
the discriminator and generator from their respective conditional 
posterior distributions; however, the two conditional posterior distributions are 
incompatible and can lead unpredictable behavior \citep{ArnoldP1989}. 
ProbGAN imposes an adaptive prior on the generator and updates the prior by successively multiplying the likelihood function at each iteration; consequently, the generator converges to a fixed point instead of a distribution.  

The remaining part of this paper is organized as follows. Section 2 describes the new formulation for the GAN based on randomized decision rules. 
Section 3 proposes a training method and proves its convergence to the Nash equilibrium. Section 4 illustrates the performance of the proposed method using synthetic and real data examples. Section 5 concludes the paper with a brief discussion.

\section{A New Formulation \textcolor{black}{for} GAN based on Randomized Decision Rules}
 
 \subsection{Pure Strategy Minimax Game}
 
 In the original work \cite{Goodfellow2014}, the GAN is trained by competing the discriminator and generator neural networks in a game. Let $\theta_d$ denote the parameters of the discriminator neural network, 
 and let $D_{\theta_d}(x)$ denote its output function which gives a score for discriminating whether or not the input sample $x$ is generated from the data distribution $p_{data}$. 
Let $G_{\theta_g}(z)$ denote the generator neural network with parameter $\theta_g$, whose input $z$ follows a given distribution $q(z)$, e.g., uniform or Gaussian, and whose output distribution is denoted by $p_{\theta_g}$. 
Define 
\begin{equation} \label{GANobj}
    \begin{split}
        \mathcal{J}_d(\theta_d;\theta_g)&=\mathbb{E}_{x\sim p_{data}}\phi_1(D_{\theta_d}(x))+\mathbb{E}_{x\sim p_{\theta_g}}\phi_2(D_{\theta_d}(x)),\\
        \mathcal{J}_g(\theta_g;\theta_d)&= - \mathbb{E}_{x\sim p_{data}}\phi_1(D_{\theta_d}(x))+
        \mathbb{E}_{x\sim p_{\theta_g}}\phi_3(D_{\theta_d}(x)),
    \end{split}
\end{equation}
where $\phi_1(D)=\log(D)$, 
$\phi_2(D)=\log(1-D)$, and $\phi_3(D)=-\log(1-D)$ or $\log(D)$ are as defined in \cite{Goodfellow2014}.
The general form of the game introduced by  \cite{Goodfellow2014} is given as follows: 
\begin{equation} \label{game1eq}
\small
(i) \ \ \max_{\theta_d}\mathcal{J}_d(\theta_d;\theta_g), \quad (ii) \ \  \max_{\theta_g}\mathcal{J}_g(\theta_g;\theta_d).
\end{equation} 
If $\phi_3=-\phi_2$, the objective of (\ref{game1eq}) represents a pure strategy minimax game, i.e., 
\begin{equation} \label{GANgame}
\small
\min_{\theta_g} \max_{\theta_d}\mathcal{J}_d(\theta_g,\theta_d),
\end{equation}
which is called minimax GAN. If $\phi_3(D)=\log(D)$, the \textcolor{black}{objective is said to be non-saturating},
which results in the same fixed point of the dynamics as the minimax GAN but addresses the issue of \textcolor{black}{vanishing gradient} suffered by the latter. 
Quite recently, \cite{zhou2019lipschitz} proposed to penalize $\mathcal{J}_d(\theta_d;\theta_g)$ by a quadratic function of the Lipschitz constant 
of $\theta_d$, which addresses the gradient
uninformativeness issue suffered by minimax GAN and improves its convergence.


 \subsection{Mixed Strategy Minimax Game}
 
 Let $\pi_g(\theta_g)$ denote a distribution of generators. Based on the randomized decision theory, we define a mixed strategy minimax game:   
 \begin{equation} \label{EBGANgame}
 \small
 \min_{\pi_g}\max_{\theta_d}  \mathbb{E}_{\pi_g} \mathcal{J}_d(\theta_d; \theta_g),
 \end{equation}
 where $\mathcal{J}_d(\theta_d; \theta_g)$ is as defined in (\ref{GANobj}), and 
 the expectation is taken with respect to $\pi_g(\theta_g)$. 
 That is, the game 
 is to iteratively search for an optimal discriminator $\theta_d$ by maximizing $\mathbb{E}_{\pi_g} \mathcal{J}_d(\theta_d; \theta_g)$ for a given generator distribution $\pi_g$ and an optimal generator distribution $\pi_g$ by minimizing $ \max_{\theta_d}  \mathbb{E}_{\pi_g} \mathcal{J}_d(\theta_d; \theta_g)$ for a given discriminator $\theta_d$. 
 In its Nash equilibrium, the discriminator is fixed and the generator is randomly drawn from the optimal generator distribution $\pi_g$, so the  equilibrium is a mixed strategy Nash equilibrium. 
  This is different from the pure strategy Nash equilibrium achieved by the minimax GAN, where both the discriminator and generator are fixed at equilibrium.
  
  From the viewpoint of statistical decision theory, (\ref{EBGANgame}) is a 
   minimax randomized decision problem, where $\pi_g$ can be viewed as a randomized decision rule and 
   $\mathbb{E}_{\pi_g} \mathcal{J}_d(\theta_d;\theta_g)$ can be viewed as a risk function. Compared to the deterministic decision formulation (\ref{GANgame}), such a randomized decision formulation naturally accounts for the uncertainty of the generator and thus helps to address the mode collapse issue. 
   \textcolor{black}{Note that a deterministic decision rule is a special case of a randomized decision rule where one decision or action has probability 1.}
   Further, \cite{YoungSmith2005} (p.11) pointed out that a minimax randomized decision rule might perform better than all other deterministic decision rules under certain situations.

   
    Let $p_{\pi_g}$ denote the distribution of the fake samples produced by the generators drawn from $\pi_g$, i.e., $p_{\pi_g}(x)=\int p_{\theta_g}(x) \pi_g(\theta_g) d\theta_g$.
  Lemma \ref{minimax} studies the basic property of the mixed strategy minimax game (\ref{EBGANgame}). \textcolor{black}{The proof of this lemma, along with the proofs of other theoretical results in this paper, is provided in the supplement.}
  
\begin{lemma}\label{minimax}
Suppose the discriminator and generator have enough capacity, $\phi_1(D)=\log(D)$, and $\phi_2(D)=\log(1-D)$. For the game (\ref{EBGANgame}), 
$\min_{\pi_g}\max_{\theta_d} \mathbb{E}_{\pi_g} \mathcal{J}_d(\theta_d; \theta_g)=-\log(4)$.
Further, if  $\tilde{\theta}_d=\arg\max_{\theta_d} \mathbb{E}_{\tilde{\pi}_g} \mathcal{J}_d(\theta_d; \theta_g)$ for some $\tilde{\pi}_g$, 
then $(\tilde{\theta}_d, \tilde{\pi}_g)$ is a Nash equilibrium point if and only if $\mathbb{E}_{\tilde{\pi}_g} \mathcal{J}_d(\tilde{\theta}_d; \theta_g)$ $=-\log(4)$; 
 at any Nash equilibrium point $(\tilde{\theta}_d, \tilde{\pi}_g)$,  $p_{\tilde{\pi}_g}=p_{data}$ holds and $D_{\tilde{\theta}_d}(x)=1/2$ for any $x\sim p_{data}$, where $p_{\tilde{\pi}_g}=\int p_{\theta_g} \tilde{\pi}_g(\theta_g)d \theta_g$ and \textcolor{black}{$x\sim p_{data}$ means $x$ is distributed according to $p_{data}$. }
\end{lemma}

 Lemma \ref{minimax} can be generalized to other choices of $\phi_1$ and $\phi_2$. In general, if $\phi_1$ and $\phi_2$ satisfy that (i) $\phi_1'>0$, $\phi_2'<0$, $\phi_1''\leq 0$, $\phi_2''\leq0$, where $\phi_i'$ and 
$\phi_i''$ denote the first and second derivatives of $\phi_i$ ($i=1,2$), respectively;  and 
(ii) there exists some value $a$ such that $\phi_1'(a)+\phi_2'(a)=0$, then the conclusion of the \textcolor{black}{lemma} still holds except that $D_{\tilde{\theta}_d} \equiv a$ in this case. 

\subsection{Mixed Strategy Nash Equilibrium} 
 
Let $q_g(\theta_g)$ denote the prior distribution of $\theta_g$, and let $N$ denote the training sample size. Define 
\begin{equation} \label{EBdist}
\small
\begin{split}
\pi(\theta_g|\theta_d,\mathcal{D}) & \propto \exp\{ \mathbb{J}_g(\theta_g; \theta_d)\} q_g(\theta_g),
\end{split}
\end{equation}
where
\begin{equation*} 
\small
 \mathbb{J}_g(\theta_g; \theta_d)
=N \mathcal{J}_g(\theta_g; \theta_d)=N( - \mathbb{E}_{x\sim p_{data}}\phi_1(D_{\theta_d}(x))+\mathbb{E}_{x\sim p_{\theta_g}} \phi_3(D_{\theta_d}(x)),
\end{equation*}
and $\phi_3$ is an appropriately defined function, e.g., $\phi_3(D)=-\log(1-D)$ or $\log(D)$ as in \cite{Goodfellow2014}. For the game (\ref{EBGANgame}), we propose to solve for $\theta_d$ by setting 
\begin{equation} \label{EBeq1}
\small
\tilde{\theta}_d=\arg\max_{\theta_d} \int \mathcal{J}_d(\theta_d;\theta_g) \pi(\theta_g|\theta_d,\mathcal{D}) d\theta_g,
\end{equation}
where $\mathcal{J}_d(\theta_d;\theta_g)$ is as defined in (\ref{GANobj}) and then, with a slight abuse of notation, setting 
\begin{equation} \label{EBeq2}
\small
\tilde{\pi}_g= \pi(\theta_g|\tilde{\theta}_d, \mathcal{D}).
\end{equation}
 
Theorem \ref{thm:00} shows that  $(\tilde{\theta}_d,\tilde{\pi}_g)$ defined in
(\ref{EBeq1})-(\ref{EBeq2}) is a \textcolor{black}{Nash equilibrium point for} the game (\ref{EBGANgame}) as $N\to \infty$. 
 
\begin{theorem} \label{thm:00}
Suppose that the discriminator and generator have enough capacity, $\phi_1(D)=\log(D)$, $\phi_2(D)=\log(1-D)$, $\phi_3=-\log(1-D)$, and the following conditions hold: (i)  $\dim(\theta_g)$, the dimension of the generator, grows with $N$ at a rate of $O(N^{\zeta})$ for some $0 \leq \zeta <1$; and (ii) the prior density function $q_g(\theta_g)$ is upper bounded on the parameter space $\Theta_g$ of the generator. Then $(\tilde{\theta}_d,\tilde{\pi}_g)$ defined in (\ref{EBeq1})-(\ref{EBeq2}) is a Nash equilibrium point \textcolor{black}{for} the game (\ref{EBGANgame}) as $N\to \infty$.
\end{theorem}

 Condition (ii) can be satisfied by many prior distributions, e.g., 
  the uninformative prior $q_g(\theta_g) \propto 1$ and the Gaussian prior. In 
  addition, we consider an extra type of prior, namely, KL-prior, in this paper. 
  The KL-prior is given by 
 \begin{equation} \label{KLprior}
 \small
 q_g(\theta_g) \propto \exp\{-\lambda D_{KL}(p_{data}|p_{\theta_g})\},
 \end{equation}
 where $\lambda$ is a pre-specified constant, and the KL-divergence $D_{KL}(p_{data}|p_{\theta_g})$ can be estimated by a $k$-nearest-neighbor density estimation method \citep{PrezCruz2008KullbackLeiblerDE,knnDiv2009Wang}  based on the real and fake samples.
 The motivation of this prior is to introduce to the proposed method a mechanism for enhancing the similarity between $p_{\theta_g}$ and $p_{data}$ at the density level. 
For the Gaussian prior, we generally suggest to set $\theta_g \sim \mathcal{N}(0, \sigma_N^2 I_{\dim(\theta_g)})$, where $\sigma_N^2 \geq 1/(2\pi)$ and increases with the training sample size $N$ \textcolor{black}{in such a way that the prior approaches uniformity  asymptotically as $N \to \infty$. }
\textcolor{black}{There are ways other than (\ref{EBdist})-(\ref{EBeq2}) to define $(\tilde{\theta}_d, \tilde{\pi}_g)$ and still have Theorem \ref{thm:00} be valid}. For example, one can define  $\pi(\theta_g|\theta_d, \mathcal{D}) \propto \exp\{ \mathbb{J}_g(\theta_g; \theta_d)/\tau\}$ $q_g(\theta_g)$ or $\pi(\theta_g|\theta_d, \mathcal{D}) \propto \exp\{ \mathbb{J}_g(\theta_g; \theta_d)/\tau\} (q_g(\theta_g))^{1/\tau}$ for some temperature $\tau>0$. That is, instead of   the exact conditional posterior $\pi(\theta_g|\tilde{\theta}_d,\mathcal{D})$, one can sample from its tempered version.  In the extreme case, one may employ the proposed method to find the Nash equilibrium point for the minimax GAN in a manner of simulated annealing \citep{KirkpatrickGV1983}. 
 
\begin{corollary} \label{corphi3}   
\textcolor{black}{The conclusion of Theorem \ref{thm:00} still holds if the function $\phi_3(D)=-\log(1-D)$ is replaced with $\phi_3(D)=\log(D)$.}
\end{corollary}
 
To make a more general formulation for the game (\ref{EBGANgame}), we can include a penalty term in 
$\mathcal{J}_d(\theta_d;\theta_g)$ such that 
\begin{equation} \label{penaltyeq2}
\small
 \mathcal{J}_d(\theta_d;\theta_g) =\mathbb{E}_{x\sim p_{data}}\phi_1(D_{\theta_d}(x))+\mathbb{E}_{x\sim p_{\theta_g}}\phi_2(D_{\theta_d}(x))-\lambda l(D_{\theta_d}),
 \end{equation} 
where $l(D_{\theta_d}) \geq 0$ denotes an appropriate penalty function 
on the discriminator. For example, one can set 
$l(D_{\theta_d})=\|D_{\theta_d}\|_{Lip}^{\alpha}$ for some $\alpha>1$, where 
 $\|D_{\theta_d}\|_{Lip}$ denotes the Lipschitz constant of the discriminator. 
As explained in \cite{zhou2019lipschitz}, including this penalty term enables the minimax GAN to overcome the gradient uninformativeness issue and improve its convergence.  
As implied by the proof of Lemma \ref{minimax},  where the mixture generator proposed in the paper 
can be represented as a single super generator, the arguments in \cite{zhou2019lipschitz} still apply 
and thus $\|D_{\tilde{\theta}_d}\|_{Lip}=0$ holds at the optimal discriminator $\tilde{\theta}_d=\arg\max_{\theta_d} \mathbb{E}_{\theta_g \sim \pi_g} \mathcal{J}_d(\theta_d;\theta_g)$. This further implies that the extra penalty term 
$-\lambda \|D_{\theta_d}\|_{Lip}^{\alpha}$ does not affect the definition of $\pi(\theta_g|\tilde{\theta}_d,\mathcal{D})$ and, therefore,  
Theorem \ref{thm:00} still holds with (\ref{penaltyeq2}).  

\section{Training Algorithm and Its Convergence}

This section proposes an algorithm for solving the integral optimization problem (\ref{EBeq1}) and studies its convergence to the Nash equilibrium. 

\subsection{The Training Algorithm} 
A straightforward calculation shows that 
\[
\small
\begin{split}
\nabla_{\theta_d} \int \mathcal{J}_d(\theta_d;\theta_g) \pi(\theta_g|\theta_d,\mathcal{D}) d\theta_g & = 
\mathbb{E}_{\pi_{g|d}} (\nabla_{\theta_d} \mathcal{J}_d(\theta_d;\theta_g)) \\
& +\mbox{Cov}_{\pi_{g|d}}(\mathcal{J}_d(\theta_d;\theta_g), \nabla_{\theta_d} \mathbb{J}_g(\theta_g; \theta_d)),
\end{split}
\]
where $\mathbb{E}_{\pi_{g|d}}(\cdot)$ and $\mbox{Cov}_{\pi_{g|d}}(\cdot)$ denote the mean and covariance operators with respect to $\pi(\theta_g|\theta_d,\mathcal{D})$, respectively. 
By Lemma \ref{minimax}, 
at any Nash equilibrium point we have $p_{\tilde{\pi}_g}=p_{data}$ and $D_{\tilde{\theta}_d} =1/2$. Then, following the arguments given in the proof of Theorem \ref{thm:00}, it is easy to show by Laplace approximation that at the Nash equilibrium point $\mbox{Cov}_{\pi_{g|d}}(\mathcal{J}_d(\tilde{\theta}_d;\theta_g), \nabla_{\theta_d} \mathbb{J}_g(\theta_g; \tilde{\theta}_d)) \to 0$  as $N\to \infty$. Therefore, when $N$ is sufficiently large, 
 the target equation $\nabla_{\theta_d} \int \mathcal{J}_d(\theta_d;\theta_g) \pi(\theta_g|\theta_d,\mathcal{D}) d\theta_g=0$ can be solved by solving the mean field equation  
\begin{equation} \label{expeq}
h(\theta_d)=\int H(\theta_d,\theta_g) \pi(\theta_g|\theta_d,\mathcal{D})=0,
\end{equation}
using a stochastic approximation algorithm, where $H(\theta_d,\theta_g)$ denotes an unbiased estimator of $\nabla_{\theta_d} \mathcal{J}_d(\theta_d;\theta_g)$. 
\textcolor{black}{The convergence of the solution to the Nash equilibrium can be assessed by examining the plots described in Section \ref{numsect}.}
 By the standard theory of stochastic approximation MCMC, see e.g., \cite{Benveniste1990, AndrieuMP2005,DengLiang2019,DongZLiang2022}, equation (\ref{expeq}) can be solved by iterating between the following two steps, where $\theta_d^{(t)}$ denotes the estimate of the discriminator obtained at iteration $t$, and $\theta_g^{(t)}$ denotes a generic sample of the generator simulated at iteration $t$: 
\begin{itemize}
\item[(i)] Simulate 
$\theta_g^{(t)}$ by a Markov transition kernel 
which leaves the conditional posterior $\pi(\theta_g|\theta_d^{(t-1)}, \mathcal{D})$  $\propto \exp\{ \mathbb{J}_g(\theta_g; \theta_d^{(t-1)})\} q_g(\theta_g)$ invariant.
\item[(ii)] Update the estimate of $\theta_d$ by setting $\theta_d^{(t)}=\theta_d^{(t-1)}+w_{t} H(\theta_d^{(t-1)},\theta_g^{(t)})$, where $w_{t}$ denotes the step size used at iteration $t$. 
\end{itemize} 
 Stochastic gradient MCMC algorithms, such as stochastic gradient Langevin dynamics (SGLD) \citep{Welling2011BayesianLV}, stochastic gradient Hamiltonian Monte Carlo (SGHMC) \citep{SGHMC2014} and 
momentum stochastic gradient Langevin dynamics (MSGLD) \citep{kim2020stochastic}, can be used in step (i).  Under appropriate conditions, we will show in Section \ref{theorysection}  that  $|\theta_d^{(t)}-\tilde{\theta}_d| \stackrel{p}{\to} 0$ and 
\textcolor{black}{$\theta_g^{(t)} \stackrel{d}{\to} \pi(\theta_g|\tilde{\theta}_d,\mathcal{D})$ as $t\to \infty$, 
where $\stackrel{p}{\to}$ and $\stackrel{d}{\to}$ denote convergences in 
probability and distribution, respectively.} That is, the proposed algorithm converges to the Nash equilibrium of the mixed strategy minimax game (\ref{EBGANgame}). 


The proposed algorithm can also be viewed as an empirical Bayes-like method \citep{Morris1983}. For the case $\phi_3=-\phi_2$, the posterior $\pi(\theta_g|\theta_d,\mathcal{D})$ can be expressed as 
\begin{equation} \label{EBeq3}
\small
\pi(\theta_g|\theta_d,\mathcal{D}) \propto 
\exp\Big\{ -\sum_{i=1}^N \phi_1(D_{\theta_d}(x_i)) -
 N* \mathbb{E}_{z \sim q} \phi_2(D_{\theta_d}(G_{\theta_g}(z))) \Big\}
 q_g(\theta_g),
\end{equation}
where $\theta_d$ can be viewed as a hyperparameter of the posterior; and the proposed algorithm is to determine $\theta_d$ by solving the equation
\begin{equation} \label{EBayeseq}
\small
 \frac{1}{N } 
 \sum_{i=1}^N \nabla_{\theta_d} \left [ \phi_1(D_{\theta_d}(x_i))+ \mathbb{E}_{\pi_g}\mathbb{E}_{z\sim q} \phi_2(D_{\theta_d}(G_{\theta_g}(z_i)))\right ] =0.
\end{equation}
In terms of the computational procedure, solving (\ref{EBayeseq}) is equivalent to maximizing the 
expected log-marginal posterior of $\theta_d$, which can be derived from 
(\ref{EBeq3}) by imposing on $\theta_d$ an improper prior $\pi(\theta_d) \propto 1$. 
To distinguish the proposed computational procedure from Bayesian GAN \citep{saatciwilson2017}, 
we call it an empirical Bayes-like GAN (or EBGAN in short). 

Algorithm \ref{alg:EBGAN} summarizes the proposed algorithm as a solver for (\ref{EBeq1}), where  
$k_g$ generators are simulated using MSGLD \citep{kim2020stochastic} 
at each iteration, and the gradients are 
estimated with a mini-batch data of size $n$ at each iteration. More precisely, we have 
\begin{equation} \label{gradeq}
\small
\begin{split}
\nabla_{\theta_g} \tilde{L}(\theta_g,\theta_d) & = 
\frac{N}{n} \sum_{i=1}^n \nabla_{\theta_g} 
\phi_3(D_{\theta_d}(G_{\theta_g}(z_i)))+\nabla_{\theta_g} 
\log q_g(\theta_g), \\
H(\theta_d,\theta_g^{(t)}) &=\frac{1}{n k_g} 
\sum_{j=1}^{k_g} \sum_{i=1}^n \nabla_{\theta_d} \left [ \phi_1(D_{\theta_d}(x_i^*))+ \phi_2(D_{\theta_d}(G_{\theta_g^{j,(t)}}(z_i)))\right ], \\
\end{split}
\end{equation}
where $\{x_i^*\}_{i=1}^n$ denotes a set of mini-batch data and $\{z_i\}_{i=1}^n$ denotes independent inputs for the generator. As illustrated by \cite{kim2020stochastic}, MSGLD tends to converge faster than SGLD, where the momentum bias term can help the sampler to escape from saddle points and accelerate its convergence in simulations on the energy landscape with  pathological curvatures.
 

\begin{algorithm}[htbp]
   \caption{Empirical Bayesian GAN}
   \label{alg:EBGAN}
\begin{algorithmic}
   \STATE {\bfseries Input:} Full data set $\mathcal{D}=\{x_i\}_{i=1}^N$,
   number of generators $k_g$, mini-batch size $n$, momentum smoothing factor $\alpha$, momentum biasing factor sequence $\{\rho_t\}_{t=1}^{\infty}$, learning rate sequence $\{\epsilon_t\}_{t=1}^{\infty}$, and 
   step size sequence $\{w_t\}_{t=1}^{\infty}$.
   \STATE {\bfseries Initialization:} $\theta_0$ from an  appropriate distribution,  set $m_0=0$; 
   \FOR{$t=1,2,\dots,$}
   \STATE (i) Sampling step:
   \FOR{$j=1,2,\dots,k_g$}
   \STATE Draw a mini-batch data $\{x_i^*\}_{i=1}^n$, and set 
   \STATE $\theta_{g}^{j,(t)}=\theta_g^{j,(t-1)}+\epsilon_t \left\{\nabla_{\theta_g}\tilde{L}(\theta_g^{j,(t-1)},\theta_d^{(t-1)})+ \rho_{t-1}  m^{j,(t-1)} \right\}+\mathcal{N}(0,2\tau\epsilon_t)$,
   \STATE $m^{j,(t)}=\alpha m^{j,(t-1)}+(1-\alpha)\nabla_{\theta_g}\tilde{L}(\theta_g^{j,(t-1)},\theta_d^{(t-1)})$.
   \ENDFOR
    \STATE (ii) Parameter estimating step:  $\theta_d^{(t)}=\theta_d^{(t-1)}+
     w_{t}H(\theta_d^{(t-1)},\theta_g^{(t)})$, where 
     $\theta_g^{(t)}=$ $(\theta_g^{1,(t)}, \ldots, \theta_g^{k_g,(t)})$.
   \ENDFOR
\end{algorithmic}
\end{algorithm}

Regarding hyperparameter settings, we have the following suggestions. In general, we set $w_t=c_1(t+c_2)^{-\zeta_1}$ for some constants $c_1>0$, $c_2 \geq 0$ and $\zeta_1 \in (0,1]$, 
which satisfies Assumption \ref{ass1}. In this paper, we set $\zeta_1=0.75$ in all computations.  
Both the learning rate sequence and the momentum biasing factor sequence are required to converge to 0 as $t \to \infty$, i.e.,  $\lim_{t\to \infty} \epsilon_t=0$ and $\lim_{t\to \infty} \rho_t=0$. For example, one might set $\epsilon_t=O(1/t^{\zeta_2})$ and $\rho_t=O(1/t^{\zeta_3})$ for some $\zeta_2, \zeta_3 \in (0,1)$. In the extreme case, one might set them to  small constants for certain problems, however, under this setting,  the convergence of $\theta_g^{(t)}$ to the target posterior distribution will hold  approximately even when $t\to \infty$. In this paper, we set $k_g=10$ and the momentum smoothing factor $\alpha=0.9$ as the default. 


\subsection{Convergence Analysis} \label{theorysection}

 Lemma \ref{thm1} establishes the convergence of the discriminator estimator, and Lemma \ref{thm2} shows how to construct the mixture generator desired for generating fake samples mimicking the real ones.
 For simplicity, we present the lemmas under the setting $k_g=1$. 

\begin{lemma}[Convergence of discriminator] \label{thm1} 
Suppose Assumptions \ref{ass1}-\ref{ass5} (given in the supplement) hold. 
If the learning rate $\epsilon_t$ is sufficiently small, then there exist a constant $\gamma$, an iteration number $t_0$ and an optimum $\tilde{\theta}_d=\arg\max_{\theta_d} $  $ \int \mathcal{J}_d(\theta_d; \theta_g) \pi(\theta_g|\theta_d,\mathcal{D}) d\theta_g$
such that for any $t \geq t_0$, 
\[
\small
\mathbb{E} \|\theta_d^{(t)}-\tilde{\theta}_d\|^2\leq \gamma w_t,
\]
where $t$ indexes iterations, $w_t$ is the step 
size satisfying Assumption \ref{ass1}, and an explicit formula of $\gamma$ is given in (\ref{rate_bound}).  
\end{lemma}
 
 As shown in (\ref{rate_bound}), the expression of $\gamma$ consists of two terms. The first term $\gamma_0$ depends only on the sequence $\{\omega_t\}$ and the stability constant of the mean field function  $h(\theta_d)$, while the second term characterizes the effects of the learning rate sequence $\{\epsilon_t\}$ and other constants (given in the assumptions) on the convergence of $\{\theta_d^{(t)}\}$.
 In particular,  $\{\epsilon_t\}$ \textcolor{black}{affects the convergence} of $\{\theta_d^{(t)}\}$ via the upper bound of $\mathbb{E} \|\theta_g^{(t)}\|^2$. See Lemma \ref{l2bound} for the definition of the upper bound.

\begin{lemma} [Ergodicity of generator]  \label{thm2}
Suppose Assumptions \ref{ass1}-\ref{ass7} (given in the supplement) hold. For a smooth test function $\psi(\theta_g)$ with $\|\psi(\theta_g)\| \leq C(1+\|\theta_g\|)$ \textcolor{black}{for some constant $C$}, define  
\begin{equation} \label{esteq}
\small
\hat\psi_T = \frac{\sum_{t=1}^T \epsilon_t \psi(\theta_g^{(t)})}{\sum_{t=1}^T \epsilon_t},
\end{equation}
where $T$ is the total number of iterations.  
Let $\bar\psi = \int \psi(\theta_g)\pi(\theta_g|\tilde{\theta}_d,\mathcal{D})d\theta_g$, 
$S_T=\sum_{t=1}^T \epsilon_t$, and 
$\Delta V_t= \nabla_{\theta_g} \tilde{L}(\theta_g^{(t)},\theta_d^{(t)})- \nabla_{\theta_g} L(\theta_g^{(t)},\theta_d^{(t)})$. 
\begin{itemize} 
\item[(i)] Suppose the following conditions are satisfied: 
the momentum biasing factor sequence $\{\rho_t: \ t=1,2,\ldots\}$ decays to 0,  
the learning rate sequence $\{\epsilon_t: \ t=1,2,\ldots\}$ decays to 0,    $\sum_{t=1}^{\infty} \epsilon_t=\infty$, and $\lim_{T\to \infty} \frac{\sum_{t=1}^T \epsilon_t^2}{\sum_{t=1}^T \epsilon_t}=0$. Then there exists a constant $C$ such that 
\[
\small
\mathbb{E}\|\hat\psi_T-\bar\psi_T\|^2  \leq C\left( \sum_{t=1}^T \frac{\epsilon_t^2}{S_T^2} \mathbb{E} \|\Delta V_t\|^2 + \frac{1}{S_T} + \frac{ (\sum_{t=1}^T \epsilon_t^2)^2 }{S_T^2} \right).
\]
\item[(ii)] Suppose a constant learning rate of $\epsilon$ and a constant momentum biasing factor of $\rho$ are used. Then there exists a constant $C$ such that
\[
\small
\mathbb{E}\|\hat{\psi}_T- \bar\psi\|^2 \leq C \left( \frac{\sum_{t=1}^T \mathbb{E} \|\Delta V_t\|^2}{T^2}+ \frac{1}{T \epsilon}
+ \epsilon^2 +\rho^2\right).
\]
\end{itemize}
\end{lemma}

The estimator (\ref{esteq}) provides us a convenient way to construct $p_{\tilde{\pi}_g}$; that is, as $T \to \infty$, the corresponding mixture generator can contain all the generators simulated by 
Algorithm \ref{alg:EBGAN} in a run. We note that, by Theorem 1 of \cite{SongLiang2020},  the estimator (\ref{esteq}) can be simplified to the simple path average $\hat{\psi}_T^{\prime}=\frac{1}{T} \sum_{t=1}^T \psi(\theta_g^{(t)})$ provided that $\epsilon_t \prec \frac{1}{t}$ holds, where $a_t \prec b_t$ means $\frac{a_t}{b_t} \to 0$ as $t\to \infty$. In practice, we can use only the generators simulated after the algorithm has converged or those simulated at the last iteration. For the latter, we may require $k_g$ \textcolor{black}{to be} reasonably large. 


 \begin{remark} \label{rem2022a}
While the mixture generator produced by Algorithm \ref{alg:EBGAN} can overcome the mode collapse issue, a single generator might not, especially when an uninformative or Gaussian prior is used.
  Suppose that the uninformative prior $q_g(\theta_g) \propto 1$ is used, $\phi_3(D)=-\log(1-D)$,  and a discriminator $\tilde{\theta}_d$ with $D_{\tilde{\theta}_d}(x)=1/2$ for $x \in p_{data}$ has been obtained. With such a discriminator, there are many $\vartheta_g$'s maximizing 
 $\mathbb{J}_g(\vartheta_g;\tilde{\theta}_d)$ as long as 
 $p_{\vartheta_g} \subset p_{data}$, because the GAN evaluates the fake samples at the individual level. Here we use the notation $p_{\vartheta_g} \subset p_{data}$ to denote that the fake samples generated from $p_{\vartheta_g}$ resemble only a subset of the real samples.  At such a point $(\tilde{\theta}_d, \vartheta_g)$, we have 
 $\mathcal{J}_d(\tilde{\theta}_d;\vartheta_g)=-\log 4$ and  
 $-\mathbb{J}_g(\vartheta_g;\tilde{\theta}_d)=-N\log 4$. The latter means that 
 the generator has attained its minimum energy, although $p_{\vartheta_g} \subset p_{data}$ is still sub-optimal; in other words, 
 such a generator is trapped to a sub-optimal solution.
 However, if Algorithm \ref{alg:EBGAN} is run for sufficiently long time and the generators from different iterations are used for estimation, we can still have $\frac{1}{T k_g} \sum_{t=1}^T \sum_{i=1}^{k_g}  \int p_{\vartheta_{g,i}^{(t)}} \pi(\vartheta_{g,i}^{(t)}|\tilde{\theta}_d^{(t)}, \mathcal{D}) d\vartheta_{g,i}^{(t)} \approx p_{data}$ by assembling many sub-optimal generators (provided the learning rate  $\epsilon_t \prec 1/t$), where $\vartheta_{g,i}^{(t)}$ denotes the $i$th generator at iteration $t$ and $\tilde{\theta}_d^{(t)}$ denotes the discriminator at iteration $t$. That is, using mixture generator is a valid way for overcoming the mode collapse issue.  For the case of $\phi_3(D)=\log(D)$ and the case of the Gaussian prior, 
 this is similar. The KL-prior provides a stronger force to drive 
 $\frac{1}{k_g} \sum_{i=1}^{k_g}  \int p_{\vartheta_{g,i}^{(t)}} \pi(\vartheta_{g,i}^{(t)}|\tilde{\theta}_d^{(t)}, \mathcal{D}) d\vartheta_{g,i}^{(t)}$ 
 to $p_{data}$ as $t \to \infty$, while the choice of $k_g$ is not crucial.  
 \end{remark}

\vspace{-0.2in} 
\section{Numerical Studies} 
\label{numsect}

We illustrate the performance of the EBGAN using various examples. 
Due to the space limit, some of the examples are presented in the supplement. \footnote{The code to reproduce the results of the experiments can be found at \url{https://github.com/sehwankimstat/EBGAN}}
 
 \subsection{A Gaussian Example} \label{GaussianEx}

Consider a 2-D Gaussian example, where the real samples are generated in the following procedure \citep{saatciwilson2017}: 
(i) generate the cluster mean: $\mu \sim \mathcal{N}(0,I_2)$, where $I_2$ denotes a 2-dimensional identity matrix; (ii) \textcolor{black}{generate a mapping matrix 
$M \in \mathbb{R}^{2\times 2}$ with each element 
independently drawn from $\mathcal{N}(0,1)$;} (iii) generate 10,000 observations: $x_{i} \sim (\mathcal{N}(0,I_2) + \mu) \times M^T$,  for $i=1,2,\ldots, 10,000$. The code used for data generation is  available at \url{https://github.com/andrewgordonwilson/bayesgan/blob/master/bgan_util.py}.
Both the discriminator and generators used for this example are fully connected neural networks with ReLU activation. The discriminator has a structure of 
 $2-1000-1$, and the generator has a structure of  $10-1000-2$.
 
The original GAN \citep{Goodfellow2014} was first applied to this example with the parameter settings given in the supplement.  
Figure \ref{GaussianKL}(a) shows the empirical means of $D_{\theta_d^{(t)}}(x)$ and $D_{\theta_d^{(t)}}(\tilde{x})$ along with iterations, 
where $x$ represents a real sample and 
$\tilde{x}$ represents a fake sample simulated by the generator.
For the given choices of $\phi_1$ and $\phi_2$, 
as implied by Lemma \ref{minimax}, we should have $\mathbb{E}(D_{\theta_d^{(t)}}(x))=
\mathbb{E}(D_{\theta_d^{(t)}}(\tilde{x}))=0.5$ at the Nash equilibrium. 
As shown by Figure \ref{GaussianKL}(a), the GAN did reach the 0.5-0.5 convergence. 
However, as shown by Figure \ref{GaussianKL}(b), the generator still suffers from the mode collapse issue at this solution, where the fake samples resemble only a subset of the real samples. 
As mentioned previously,  this is due to the reasons: 
{\it The GAN evaluates the fake samples at the individual level, lacking a mechanism for enhancing the diversity of fake samples, and tends to get trapped at a sub-optimal solution for which $p_{\theta_g} \subset p_{data}$ holds while the ideal objective value $-\log 4$ can still be attained.} 
 
The mode collapse issue can be tackled by EBGAN, for which we consider both the KL-prior and Gaussian prior. 
 
 \vspace{-0.1in}
\subsubsection{KL-prior}

\begin{figure}[htbp]
\begin{center}
\begin{tabular}{cc}
(a) &(b) \\
\includegraphics[height=1.25in,width=2.0in]{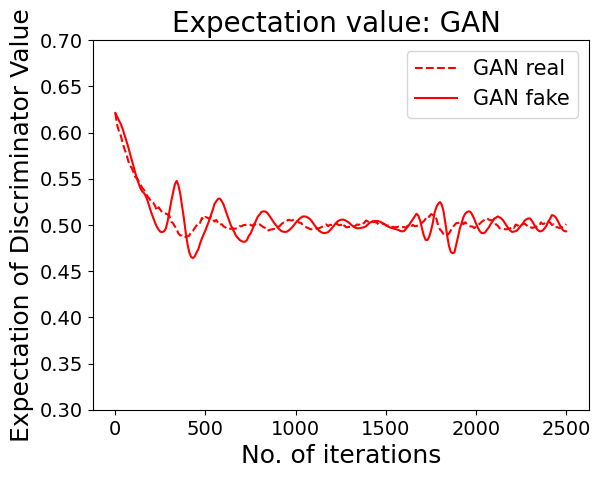} & 
\includegraphics[height=1.25in,width=2.0in]{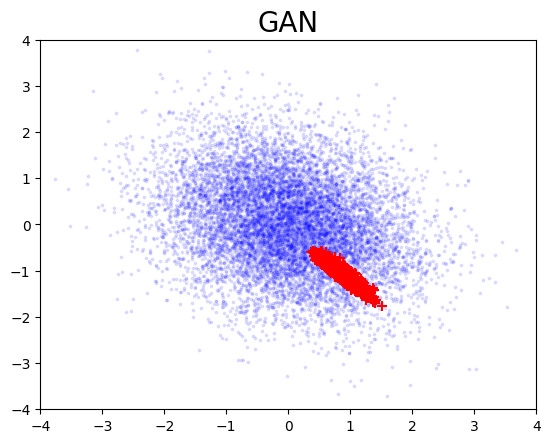} \\ 
 (c) & (d) \\
 \includegraphics[height=1.25in,width=2.0in]{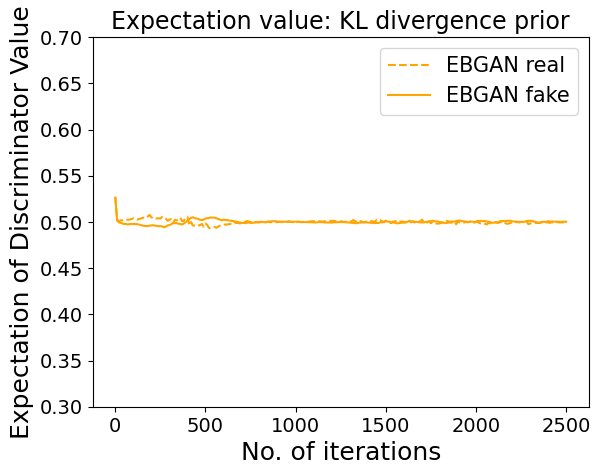} & 
\includegraphics[height=1.25in,width=2.0in]{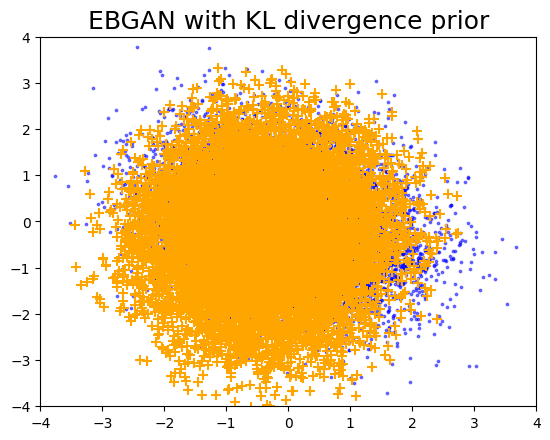} \\
\end{tabular}
\end{center}
\caption{\label{GaussianKL} Illustration of the mode collapse issue: 
(a) empirical means of $D_{\theta_d^{(t)}}(x_i)$ and $D_{\theta_d^{(t)}}(\tilde{x}_i)$ produced by GAN; (b) coverage plot of the real (dots) and fake samples (`+') generated by GAN; (c) empirical means of $D_{\theta_d^{(t)}}(x_i)$ and $D_{\theta_d^{(t)}}(\tilde{x}_i)$ produced by EBGAN;
(d) coverage plot of the real  (dots) and fake samples (`+') generated by a {\it single} generator of EBGAN. }
\end{figure}

The KL-prior is given in (\ref{KLprior}), which enhances 
the similarity between $p_{\theta_g}$ and $p_{data}$ at the density level. 
For this example, we set $\lambda=100$, set $k=1$ for $k$-nearest-neighbor density estimation (see \cite{PrezCruz2008KullbackLeiblerDE} for the estimator), and used the auto-differentiation method to evaluate the gradient $\nabla_{\theta_g} \log q_g(\theta_g)$. 
Figure \ref{GaussianKL}(c)\&(d) summarize the results of EBGAN for this example with
$\phi_3(D)=\log(D)$ and $k_g=10$. The settings for the other parameters can be found in the supplement. For EBGAN, 
Figure \ref{GaussianKL}(c) shows that it converges to the Nash equilibrium very fast, and
Figure \ref{GaussianKL}(d) shows that the fake samples simulated by a {\it single} generator match the real samples almost perfectly. 
 
In summary, this example shows that EBGAN can overcome the mode collapse issue by employing a KL-prior that enhances the similarity between $p_{\theta_g}$ and $p_{data}$ at the density level.




\subsubsection{Gaussian prior}

\begin{figure}[htbp]
\begin{center}
\begin{tabular}{c}
 (a) Evolution of coverage by GAN \\
\includegraphics[width=0.95\textwidth]{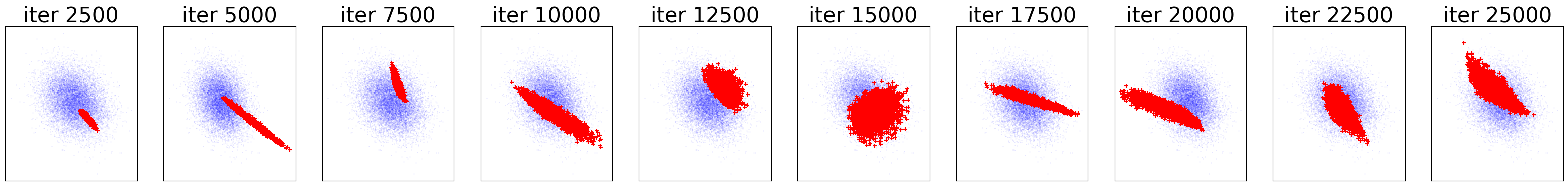} \\
 (b) Evolution of coverage by one generator from EBGAN\\
\includegraphics[width=0.95\textwidth]{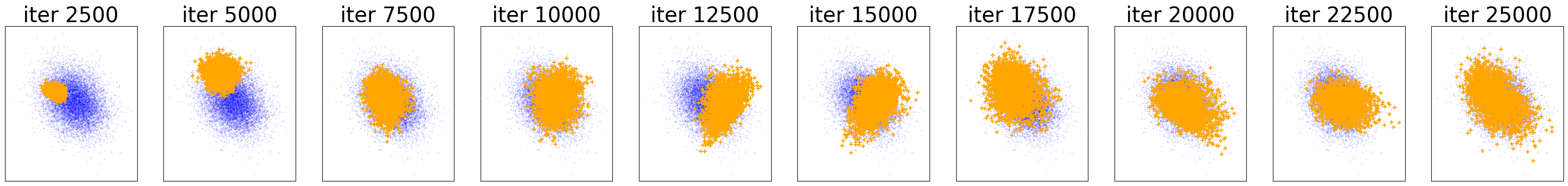} \\
(c) Evolution of coverage by integration of 10 generators from EBGAN\\
\includegraphics[width=0.95\textwidth]{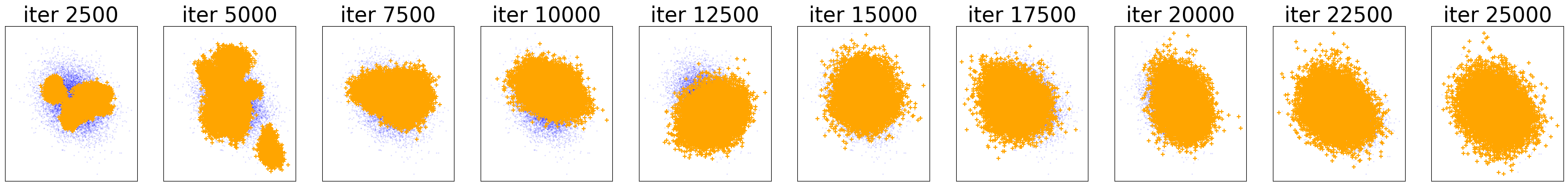} \\
\end{tabular}
\end{center}
\caption{\label{Evolutionplot} Coverage plots produced by (a) GAN, (b) a single generator of EBGAN, and (c) all 10 generators of EBGAN, where the generators for the plots from left to right are collected at iterations 2500, 5000, $\ldots$, 25000, respectively. The dot points (in blue) represent real samples; and `+' points represent fake samples produced by GAN (in red) or EBGAN (in yellow).  }
\end{figure}

We have also tried the simple Gaussian prior  $\theta_g\sim N(0,I_d)$ for this example. Compared to the KL-divergence prior, the Gaussian prior lacks the ability to enhance the similarity between $p_{\theta_g}$ and $p_{data}$, but it is much cheaper in computation. For this example, we have run EBGAN with $\phi_3(D)=\log(D)$ and $k_g=10$. The settings for other parameters can be found in the supplement.  To examine the performance of EBGAN with this cheap prior, we made a long run of 30,000 iterations. For comparison, the GAN was also applied to this example with $\phi_3(D)=\log(D)$. 
Figure S1 (in the supplement) shows the empirical means of $D_{\theta_d^{(t)}}(x)$ and $D_{\theta_d^{(t)}}(\tilde{x})$ produced by the two methods along with iterations, which indicates that both methods can reach the 0.5-0.5 convergence very fast.   
Figure \ref{Evolutionplot} shows the evolution of the coverage plot of fake samples. 
Figure \ref{Evolutionplot}(a) indicates that the GAN has not reached the Nash equilibrium even with 25,000 iterations. In contrast, Figure \ref{Evolutionplot}(b) shows that even with the cheap Gaussian prior, the EBGAN can approximately reach the Nash equilibrium with 25,000 iterations, although it also suffers from the mode collapse issue in the early stage of the run. 
Figure \ref{Evolutionplot}(c) shows that for the EBGAN, the mode collapse issue can be easily overcome by integrating multiple generators. 


Finally, we note that the convergence of the GAN and EBGAN should be checked in two types of plots, namely, empirical convergence plot of $\mathbb{E} D_{\theta_d^{(t)}}(x)$ and $\mathbb{E} D_{\theta_d^{(t)}}(\tilde{x})$, and coverage plot of the fake and real samples. The former measures 
how well an individual fake sample fits into the population of real samples, while the latter measures the diversity of fake samples, i.e., whether a wide range of fake samples is generated. 

\vspace{-0.2in} 
\subsection{A Mixture Gaussian Example} \label{multicomponentEx}


To further illustrate the performance of the EBGAN, we consider a more complex example which was taken from \cite{saatciwilson2017}. The dataset was generated from a 10-component mixture Gaussian distribution in the following procedure:
 (i) generate 10 cluster means: $\mu^{(j)} \sim \mathcal{N}(0,25 I_2)$, $j=1,2,\ldots, 10$, where $I_2$ denotes a 2-dimensional identity matrix; 
 (ii) \textcolor{black}{generate 10 mapping matrices: 
 $M^{(j)} \in \mathbb{R}^{100 \times 2}$ for $j=1,2,\ldots,10$, with each element of the matrices independently drawn from $\mathcal{N}(0,25)$.}
(iii) generate 1000 observations of $x^{(j)}$ for each $(\mu^{(j)}, M^{(j)})$: $x_{i}^{(j)} \sim (\mathcal{N}(0,I_2)*0.5 + \mu^{(j)}) \times  (M^{(j)})^T$,  for $j=1,2,\ldots,10$, $i=1,2,\ldots, 1000$.
 
For this example, the EBGAN was run with  the prior $q_g=\mathcal{N}(0,I)$, $k_g=10$,  and $\phi_3(D)=-\log(1-D)$ and $\log(D)$. The discriminator has a structure of 
 $100-1000-1$ and the generator has a structure of  $10-1000-100$, which are the same as those used in  \cite{saatciwilson2017}. The results with  $\phi_3(D)=-\log(1-D)$ are presented below and those with 
 $\phi_3(D)=\log(D)$ are presented in the supplement.

\begin{figure}[htbp]
\begin{center}
\begin{tabular}{ccc}
(a) & (b) & (c) \\
        \includegraphics[width=2.0in,height=1.25in]{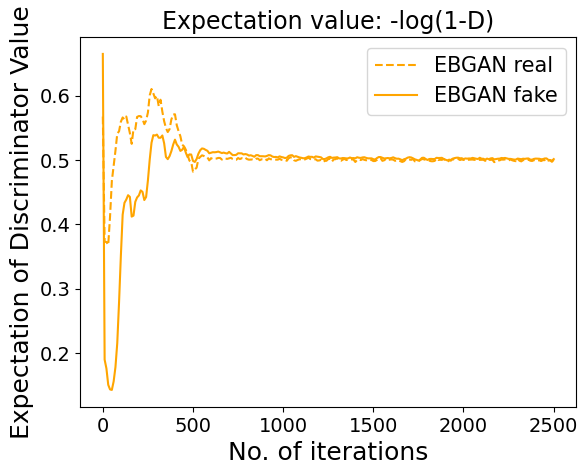} & 
        \includegraphics[width=2.0in,height=1.25in]{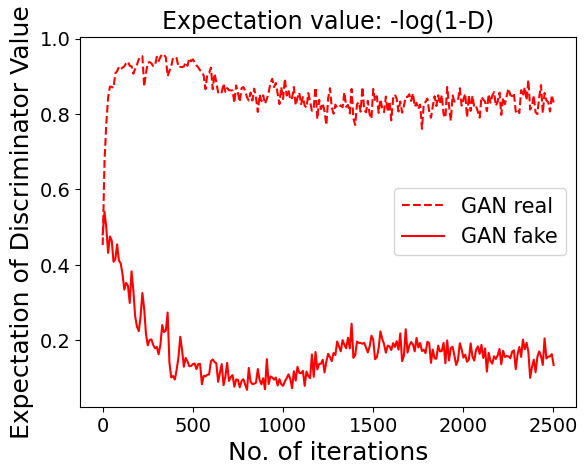} &
         \includegraphics[width=2.0in,height=1.25in]{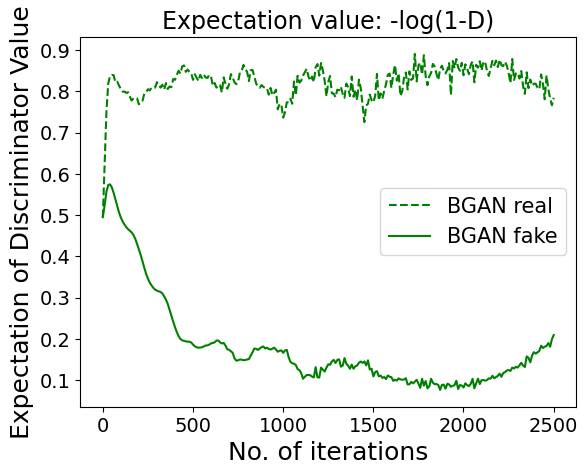} \\ 
(d) & (e) & (f) \\
        \includegraphics[width=2.0in,height=1.25in]{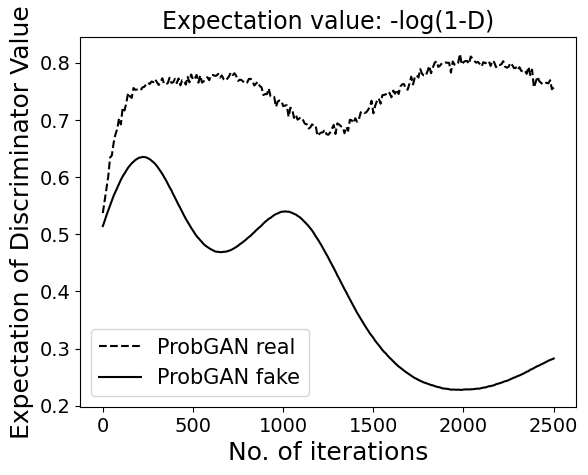} &
       \includegraphics[width=2.0in,height=1.25in]{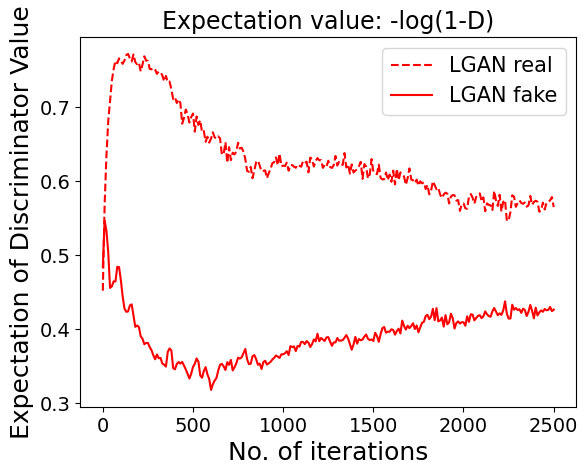} & 
         \includegraphics[width=2.0in,height=1.25in]{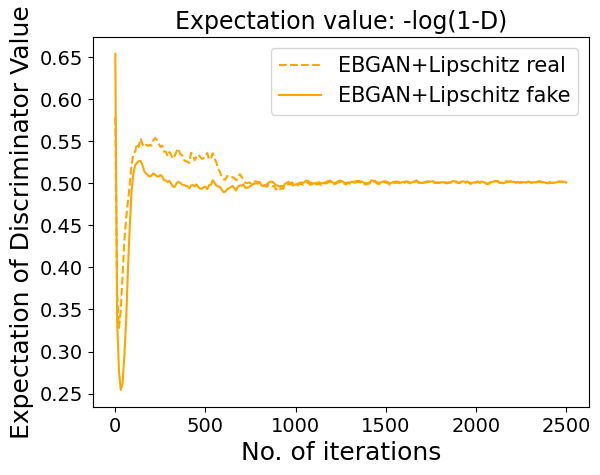} \\
\end{tabular}
\end{center}
\caption{ \label{minmaxfig1} Nash equilibrium convergence  plots with $\phi_3(D)=-\log(1-D)$, which compare the empirical means of $D_{\theta_d^{(t)}}(x_i)$ and $D_{\theta_d^{(t)}}(\tilde{x}_i)$ produced by 
 different methods along with iterations: (a) EBGAN,
  (b) minimax GAN, (c) Bayesian GAN, (d) ProbGAN, (e) Lipschitz-GAN, (f) EBGAN with a Lipschitz penalty.  } 
\end{figure}

For comparison, the minimax GAN \citep{Goodfellow2014}, Bayesian GAN \citep{saatciwilson2017}, ProbGAN \citep{he2018probgan}, and Lipschitz GAN \citep{zhou2019lipschitz} were applied 
to this example with the parameter settings given in the supplement. For all the four methods, we employed the same settings of $\phi_1$, $\phi_2$ and $\phi_3$ and the same discriminator and generator as the EBGAN. 
For a thorough comparison, we have tried to train the EBGAN with a Lipschitz penalty. 

Figure \ref{minmaxfig1} examines the convergence of $\mathbb{E}(D_{\theta_d^{(t)}}(x))$ and  $\mathbb{E}(D_{\theta_d^{(t)}}(\tilde{x}))$. 
It indicates that except for the EBGAN, 
none of the four methods, minimax GAN, Bayesian GAN, ProbGAN and Lipschitz-GAN, 
has reached the 0.5-0.5 convergence. Compared to Figure \ref{minmaxfig1}(a), Figure \ref{minmaxfig1}(f) shows that the Lipschitz penalty improves the convergence of the EBGAN slightly.
  
\begin{figure}[htbp]
\begin{center}
\begin{tabular}{ccc}
(a) & (b) & (c) \\
        \includegraphics[width=2.0in,height=1.25in]{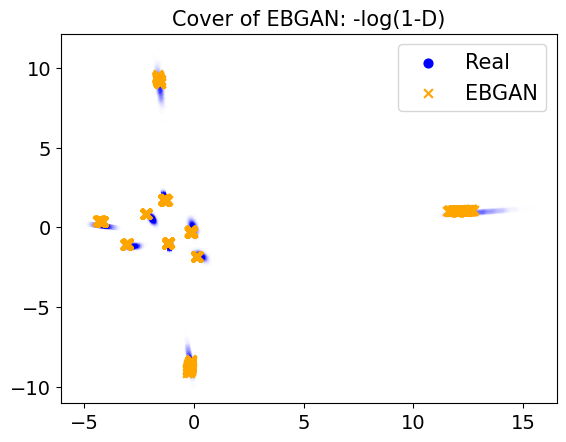} & 
            \includegraphics[width=2.0in,height=1.25in]{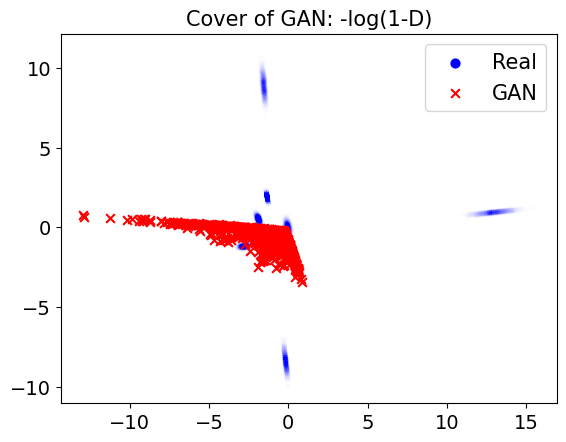} &
        \includegraphics[width=2.0in,height=1.25in]{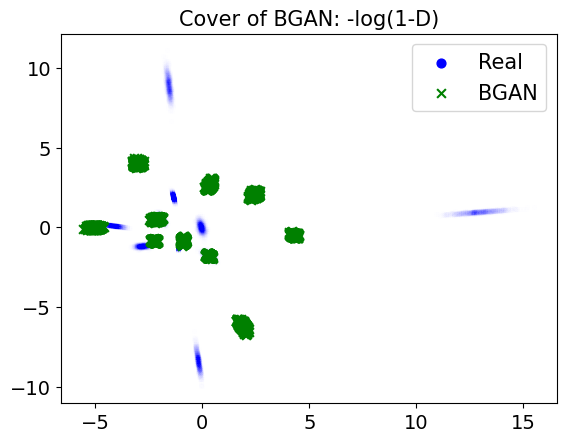} \\
(d) & (e) & (f) \\
        \includegraphics[width=2.0in,height=1.25in]{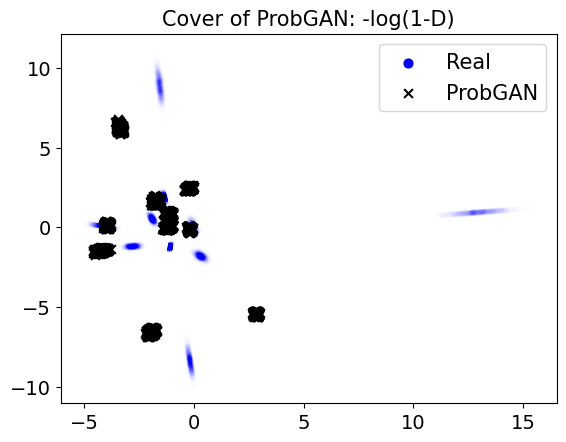} & 
        \includegraphics[width=2.0in,height=1.25in]{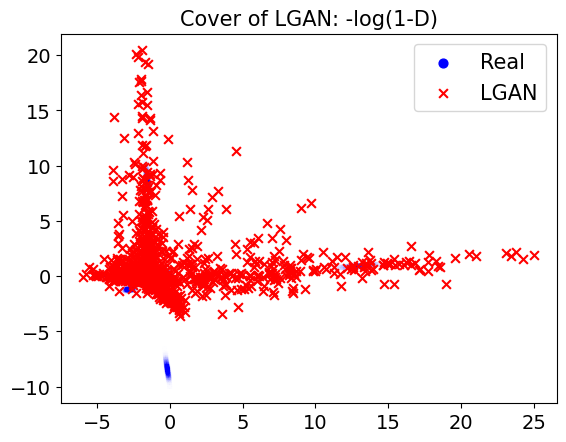} &
          \includegraphics[width=2.0in,height=1.25in]{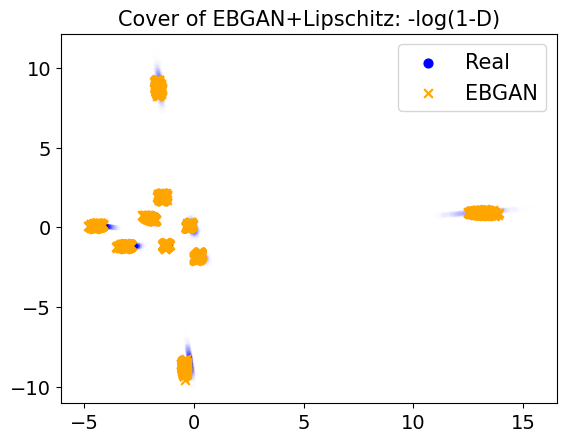} \\
    \end{tabular}
    \end{center}
    \caption{\label{minmaxfig2} Component recovery plots with $\phi_3(D)=-\log(1-D)$:  (a) EBGAN with $\lambda=0$, (b) minimax GAN, (c) BGAN, (d) ProbGAN, (e) Lipschitz GAN, and (f) EBGAN with a Lipschitz penalty. } 
\end{figure}

Other than the convergence plots of $\mathbb{E}(D_{\theta_d^{(t)}}(x))$ and 
$\mathbb{E}(D_{\theta_d^{(t)}}(\tilde{x}))$,
we checked whether the fake samples recover all 10 components of the mixture distribution, 
where the principal component analysis (PCA) was used for high-dimensional data visualization. For EBGAN,
we used only the generators obtained at the last iteration: we simulated 1000 fake samples from each of $k_g=10$ generators.
As shown in Figure \ref{minmaxfig2}, EBGAN recovered all 10 components in both cases with or without the penalty term, while the other four methods failed to do so. 
\textcolor{black}{The minimax GAN and Lipschitz GAN, both of which work with a single generator, failed for this example.} The BGAN and ProbGAN worked better than minimax GAN, but still missed a few components. 
In the supplement, we compared the performance of different methods with $\phi_3=\log(D)$. The results are very similar to Figure \ref{minmaxfig1} and Figure \ref{minmaxfig2}. 


For the EBGAN, we have also tried to use generators simulated at multiple iterations, e.g., those in the last 2000 iterations. We found that the component recovery plot can be further improved.
 For simplicity, we used only the generators obtained at the last iteration. If the EBGAN is run with a larger value of $k_g$, the overlapping area can also be further improved.

In summary, EBGAN performs very well for this mixture example: the fake samples generated by it exhibit both good quality and diversity.
The comparison with existing methods indicates that integrating multiple generators is essential for overcoming the mode collapse issue, particularly when the objective function lacks a mechanism to enhance the similarity between $p_{\theta_g}$ and $p_{data}$ at the density level.
 

 \subsection{Image Generation} \label{imagesect}
 
 Fashion-MNIST is a dataset of 60,000 training images and 10,000 test images. Each image is of size  $28\times28$ and has a label from 10 classes: T-shirt, Trouser, $\ldots$, Ankle boot. The full description for the dataset can be found at \url{https://github.com/zalandoresearch/fashion-mnist}. 
 
 For this example, we compared EBGAN with GAN, BGAN and ProbGAN with parameter settings given in the supplement. 
 The results are summarized in Figure \ref{Fganfig} and Table \ref{FashionMNISTtab}. 
 Figure \ref{Fganfig} shows that for this example, the EBGAN can approximately achieve the 0.5-0.5 convergence for $\mathbb{E}(D_{\theta_d^{(t)}}(x_i))$ and $\mathbb{E}(D_{\theta_d^{(t)}}(\tilde{x}_i))$; that is, the EBGAN can produce high quality images which are almost indistinguishable from real ones. 
 However, none of the existing three methods can achieve such good convergence. 

 \begin{figure}[h!]
\begin{center}
\begin{tabular}{ccccc}
(a) & (b) & (c) & (d) & (e) \\
       \includegraphics[width=0.17\textwidth,height=0.90 in]{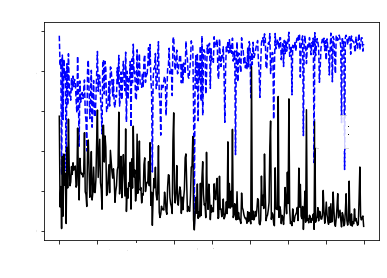} & 
        \includegraphics[width=0.17\textwidth]{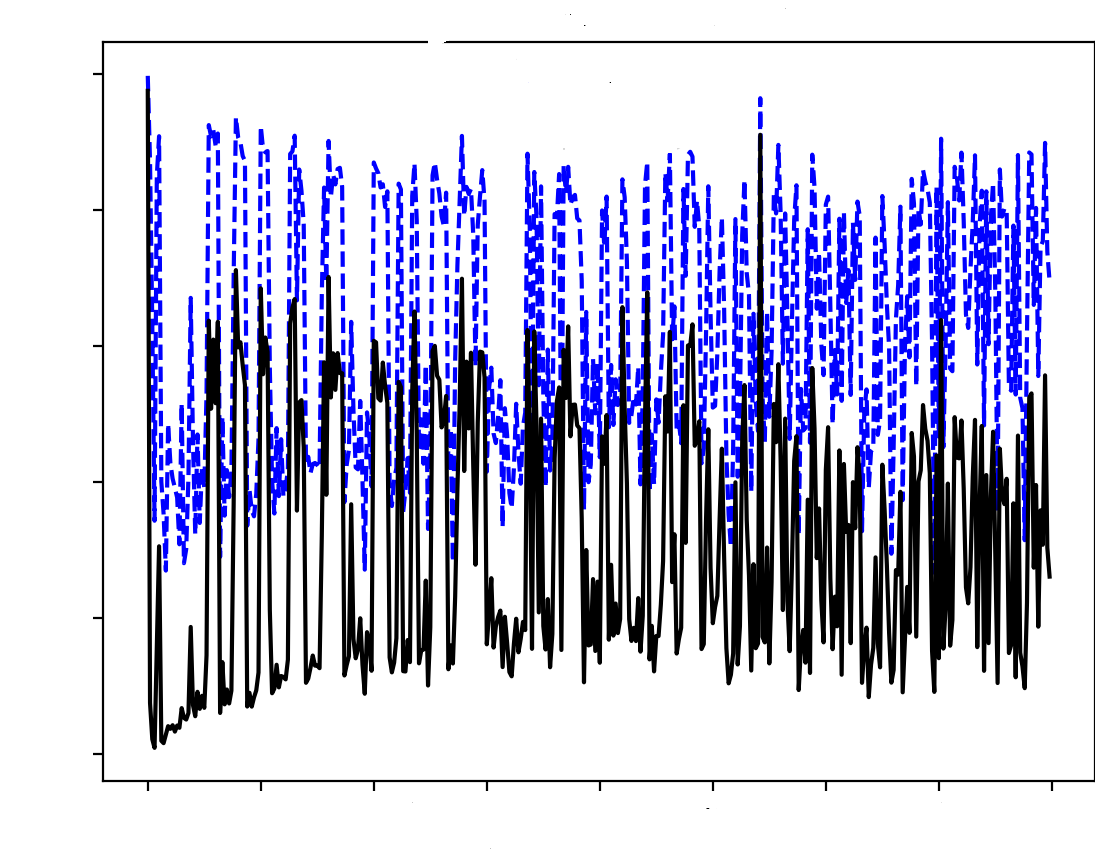} & \includegraphics[width=0.17\textwidth]{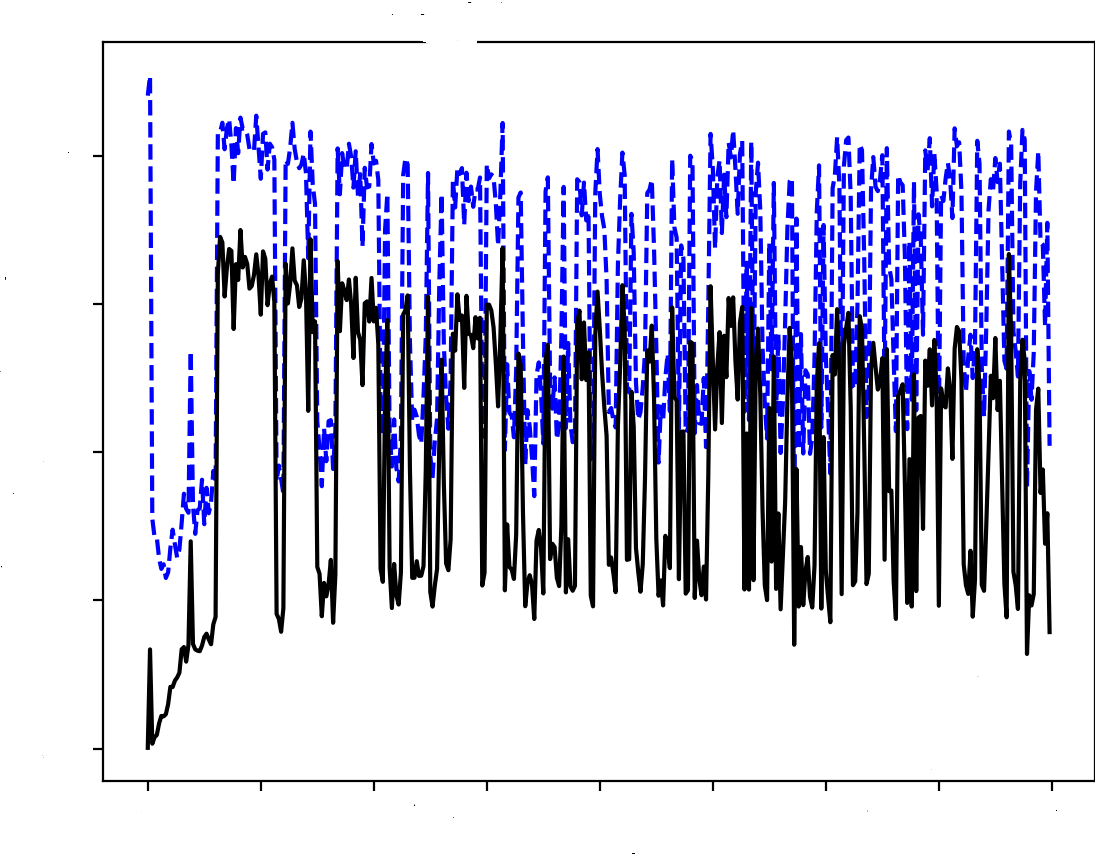} &
         \includegraphics[width=0.17\textwidth]{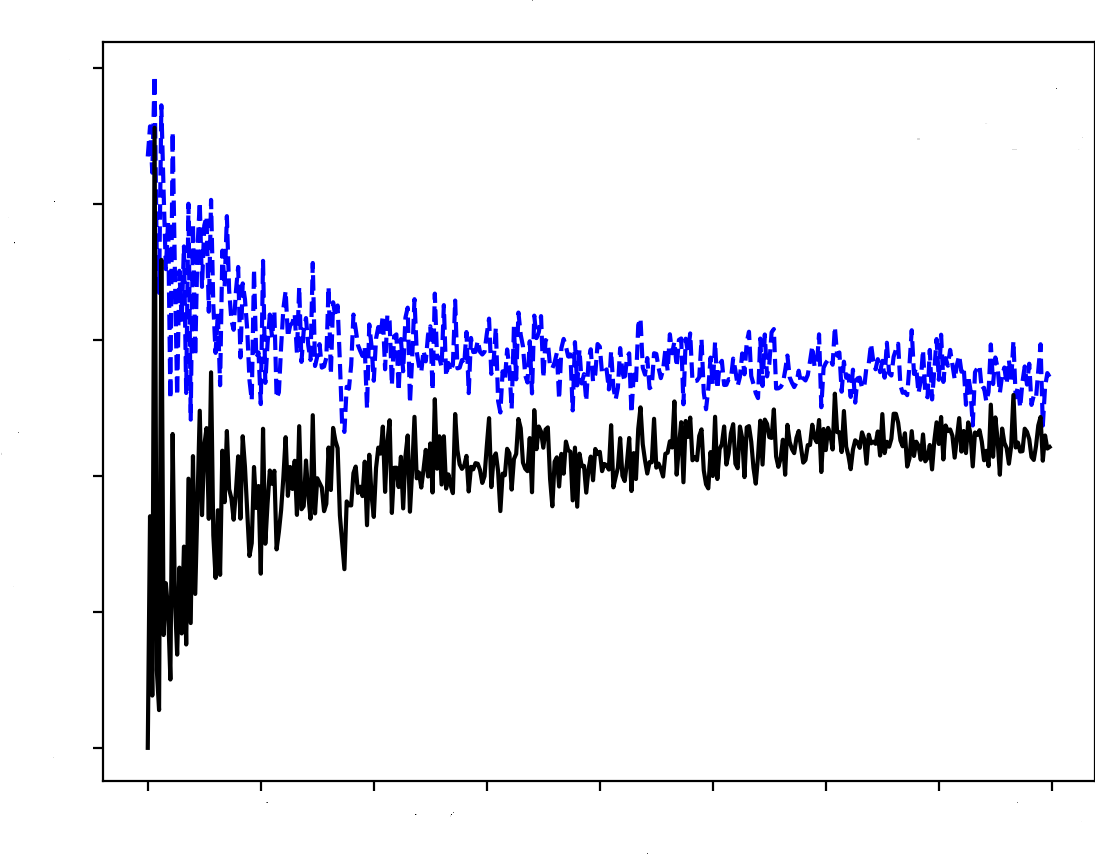}  &
        \includegraphics[width=0.17\textwidth]{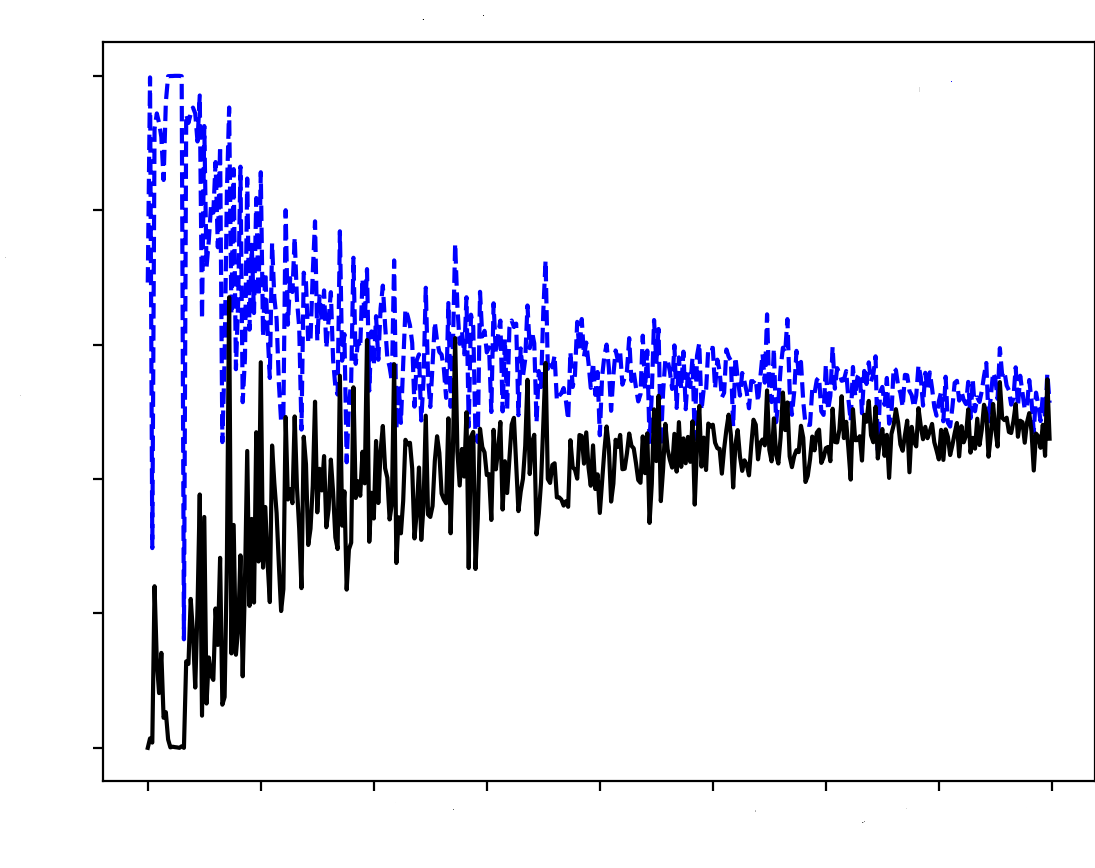} \\ 
            \includegraphics[width=0.17\textwidth]{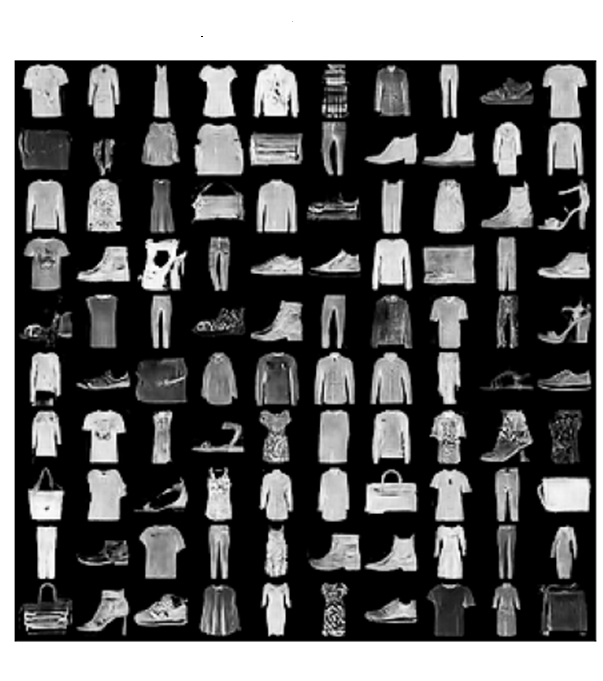}  & 
            \includegraphics[width=0.17\textwidth]{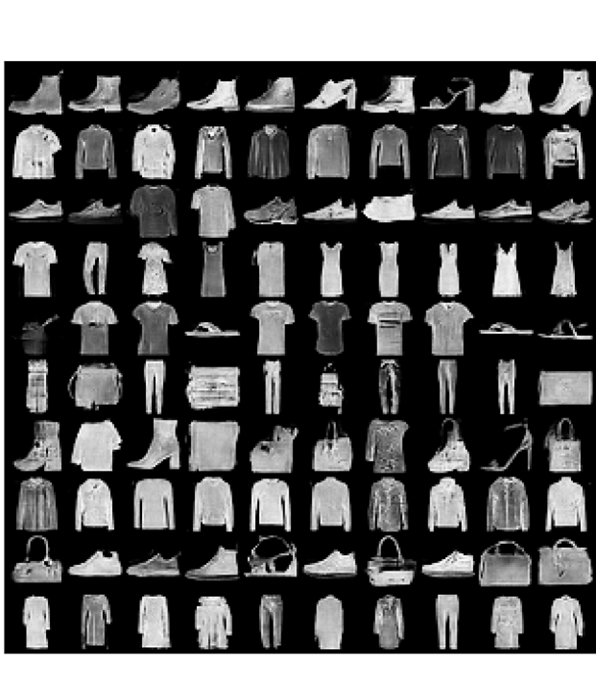} &  \includegraphics[width=0.17\textwidth]{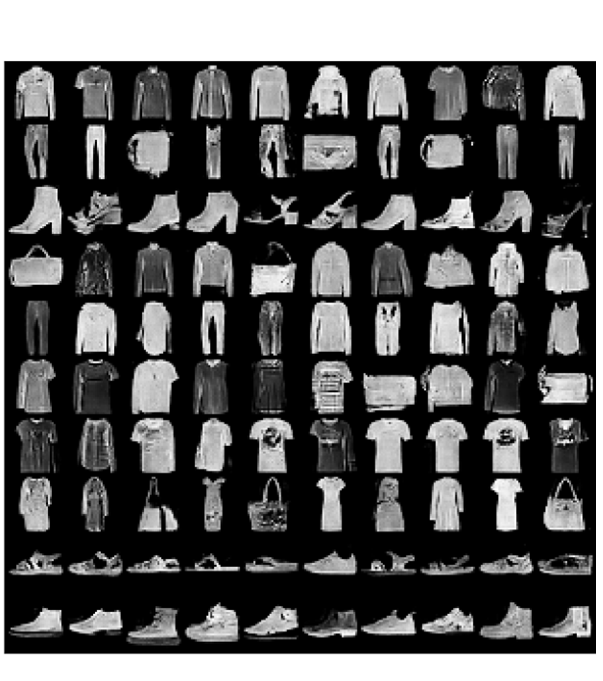} &
            \includegraphics[width=0.17\textwidth]{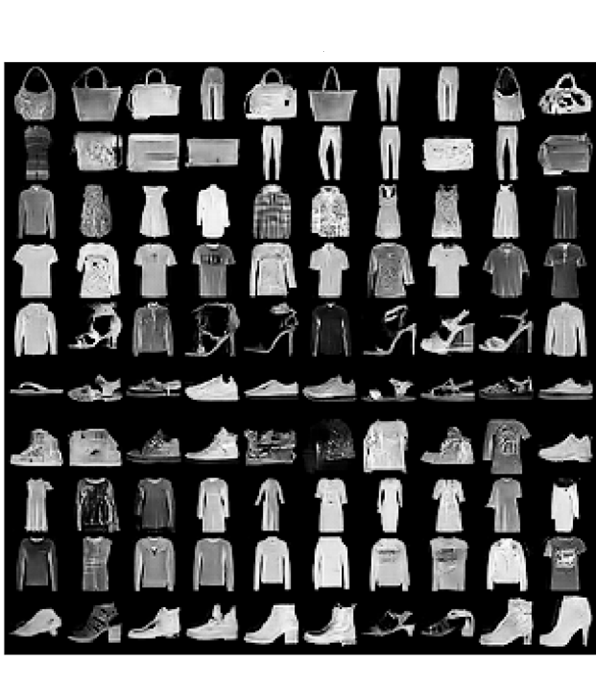} & 
            \includegraphics[width=0.17\textwidth]{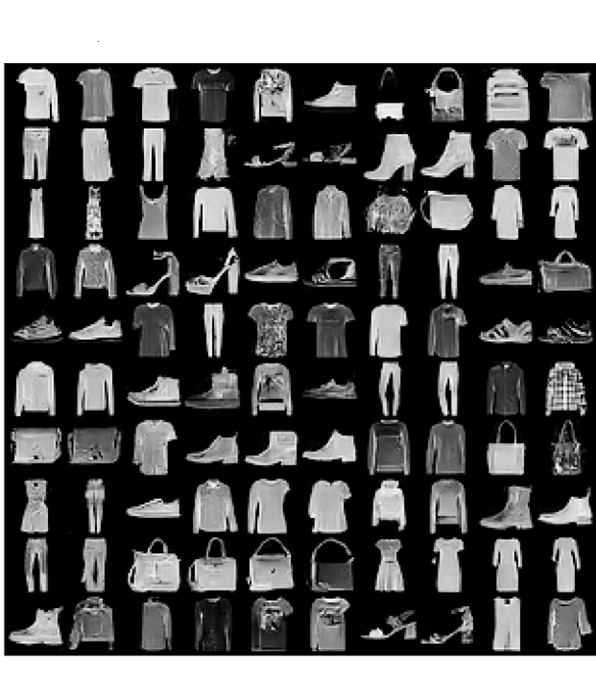} 
\end{tabular}
\end{center}
\caption{Convergence plots and images produced by (a) GAN, (b) Bayesian GAN, 
(c) ProbGAN,  (d) EBGAN with a KL-divergence prior,
and (e) EBGAN with a Gaussian prior. 
For each of the convergence plots, $x$-axis represents iterations, ranging from 0 to 40,000; and $y$-axis represents the empirical mean of discriminator values, ranging from 0 to 1.0, where dotted and solid lines are for real and fake samples, respectively.
} 
\vspace{-0.2in}
\label{Fganfig}
\end{figure}

\begin{table}[htbp]
  \caption{Average IS, 1-WD, and MMD values produced by different methods for Fashion MNIST, where the averages and standard deviations (given in the parentheses) were calculated based on 5 independent runs.}
  \label{FashionMNISTtab}
  \centering
\begin{adjustbox}{width=1.0\textwidth}
\begin{tabular}{cccccc} \toprule
  Method   & GAN & Bayesian GAN & ProbGAN & EBGAN(KL) & EBGAN(Gaussian)\\
\midrule
 IS    & 7.525 (0.011) & 7.450 (0.044)  & 7.384 (0.056)  & 7.606 (0.035) & 7.712 (0.024)  \\ 
 1-WD  &  6.360 (0.012)  & 6.363 (0.015)  & 6.367 (0.024) &   6.356 (0.015)   &6.287 (0.025)\\ 
 MMD  & 0.276 (0.002)  & 0.277 (0.002)& 0.276 (0.002) &  0.257 (0.001)   & 0.275 (0.003)     \\ 
  \bottomrule
\end{tabular}
\end{adjustbox}
\end{table}

We further assess the quality of images generated by different methods using 
three metrics: inception scores (IS) \citep{Salimans2016ImprovedTF},  \textcolor{black}{first moment Wasserstein distance (1-WD), 
 and maximum mean discrepancy (MMD). 
Refer to Section \ref{metricsection} of the supplement for their definitions and calculation procedures.} The results are summarized in Table \ref{FashionMNISTtab}, which indicates the superiority of the EBGAN 
in image generation.

\subsection{Nonparametric Clustering} \label{clustersect}

Clustering has been extensively studied in unsupervised learning with classical methods such as expectation-maximization (EM) \citep{Dempster1977}  and K-means. 
Although its main focus is to group data into different classes, it would be even more beneficial if clustering was done along with dimension reduction, as it enhances the interpretability of clusters. This simultaneous goal of clustering and dimension reduction can be achieved through  generative clustering methods such as Cluster GAN \citep{Mukher19clustergan} and GAN-EM \citep{Zhao2019GANEM}.


Since the cluster structure is generally not retained in the latent space of the GAN, Cluster GAN modifies the structure of the GAN to include an encoder network, which enforces precise recovery of the latent vector so that it can be used for clustering.  Let the encoder be parameterized by $\theta_e$.
Cluster GAN works with the following two objective functions: 
\begin{equation} \label{ClusterGANobj}
\small
    \begin{split}
        \mathcal{J}_{d}(\theta_d;\theta_g)&=\mathbb{E}_{x\sim p_{data}}\phi_1(D_{\theta_d}(x))+\mathbb{E}_{x\sim p_{\theta_g}}\phi_2(D_{\theta_d}(x)),\\
        \mathcal{J}_{g,e}(\theta_g,\theta_e;\theta_d)&=
        -\mathbb{E}_{x\sim p_{data}}\phi_1(D_{\theta_d}(x))+
        \mathbb{E}_{x\sim p_{\theta_g}}\phi_3(D_{\theta_d}(x)) \\
     &\quad  -\beta_n\mathbb{E}_{x\sim p_{\theta_g}}\|z_n-\mathcal{E}_{\theta_e}^{(1)}(x)\|^2 
       -\beta_c\mathbb{E}_{x\sim p_{\theta_g}}\mathcal{H}(z_c,\mathcal{E}_{\theta_e}^{(2)}(x)),
    \end{split}
\end{equation} 
where $z=(z_n,z_c)$ is used as the latent vector to generate data, $z_n$ denotes a random noise vector, and $z_c$ denotes an one-hot vector representing the index of clusters; $\mathcal{H}(\cdot,\cdot)$ is the cross-entropy loss; and $\beta_n\geq 0$ and $\beta_c\geq 0$ are regularization parameters. 
The choice of $\beta_n$ and $\beta_c$ should balance the two ends: large values of $\beta_n$ and $\beta_c$ will delay the convergence of the generator, while small values of $\beta_n$ and $\beta_c$ will delay the convergence of the encoder.  In general, we set $\beta_n=o(1)$ and $\beta_c=o(1)$. 
In this setup, the encoder is the inverse of the  generator so that it recovers from data to the low-dimensional latent vector by $\mathcal{E}_{\theta_e}(x)=(\mathcal{E}_{\theta_e}^{(1)}(x),\mathcal{E}_{\theta_e}^{(2)}(x)):\mathbb{R}^{d\times d}\rightarrow z=(\widehat{z}_n,\widehat{z}_c)$, where $\mathcal{E}_{\theta_e}^{(2)}(x)$ can be used for clustering. 

Cluster EBGAN 
extends the structure of Cluster GAN by allowing multiple generators to be trained simultaneously. Similar to (\ref{EBeq1})-(\ref{EBeq2}), cluster EBGAN works by solving the following integral optimization problem  
\begin{equation} \label{clusterEB1} 
 \tilde{\theta}_d  =  \arg\max_{\theta_d}
\int \mathcal{J}_d(\theta_d;\theta_g) \pi(\theta_g,\theta_e|\theta_d,\mathcal{D}) d\theta_g d\theta_e, 
\end{equation}
and then simulate $(\theta_g, \theta_e)$ from the distribution
\begin{equation} \label{clusterEB2}
\small
\pi(\theta_g,\theta_e|\tilde{\theta}_d,\mathcal{D}) \propto \exp\{ \mathbb{J}_{g,e}(\theta_g,\theta_e; \tilde{\theta}_d)\} p_{g,e}(\theta_g,\theta_e),
\end{equation}
where $\mathbb{J}_{g,e}(\theta_g,\theta_e; \theta_d)=
N \mathcal{J}_{g,e}(\theta_g, \theta_e;\theta_d)$ and 
$p_{g,e}(\theta_g,\theta_e)$ denotes the prior density function of $(\theta_g,\theta_e)$. 
Let $\pi_c(\theta_g|\tilde{\theta}_d,\mathcal{D})=\int \pi(\theta_g,\theta_e|\tilde{\theta}_d,\mathcal{D})d\theta_e$ be the marginal conditional density function of $\theta_g$. Then, by Theorem \ref{thm:00} and Corollary \ref{corphi3},  $(\tilde{\theta}_d,\pi_c(\theta_g|\tilde{\theta}_d,\mathcal{D}))$ is
an asymptotic solution to the game (\ref{EBGANgame}) as $N\to \infty$. 
Note that for  $\pi_c(\theta_g|\tilde{\theta}_d,\mathcal{D})$, we can simply treat 
$q_g(\theta_g) \propto \int \exp\{-N \beta_n\mathbb{E}_{x\sim p_{\theta_g}}\|z_n-\mathcal{E}_{\theta_e}^{(1)}(x)\|^2 
      -N \beta_c\mathbb{E}_{x\sim p_{\theta_g}}\mathcal{H}(z_c,\mathcal{E}_{\theta_e}^{(2)}(x))\} p_{g,e}(\theta_g,\theta_e) d\theta_e$ as the prior of $\theta_g$.
      Therefore, Theorem \ref{thm:00} and Corollary \ref{corphi3} still apply. 
 
 The equations (\ref{clusterEB1})-(\ref{clusterEB2}) provide a general formulation for extended applications of the EBGAN. In particular, the embedded decoder (latent variable $\to$ fake data) and encoder (fake data $\to$ latent variable) enable the EBGAN to be used in many nonparametric unsupervised statistical tasks. Other than clustering, it can also be used for tasks such as  dimension reduction and image compression.


Classical clustering methods can be roughly grouped into three categories, namely, partitional clustering, hierarchical clustering, and density-based clustering. The K-means clustering, agglomerative clustering, and density-based spatial clustering of applications with noise (DBSCAN) are well known representatives of the three categories, respectively. In what follows, Cluster-EBGAN is compared with the representative clustering methods as well as Cluster GAN on four different datasets. 
For MNIST, we used a deep convolutional GAN (DCGAN) with conv-deconv layers, batch normalization and leaky relu activations. For other datasets, simple feed forward neural networks were used. 
The results are summarized in Table \ref{purity-table}, which indicates the superiority of the Cluster-EBGAN in nonparametric clustering. Refer to the supplement for the details
of the experiments. 

\begin{table}[htbp]
 \caption{Comparison of Cluster EBGAN and other methods on different datasets, where  average purity, 
  adjusted rand index (ARI) and their standard errors (in parentheses) were computed based on five independent runs.}
 \vspace{2mm}
  \label{purity-table}
  \centering
  \begin{adjustbox}{width=\textwidth}
  \begin{tabular}{ccccccc} \toprule
  Data & Metric & K-means& Agglomerative & DBSCAN &Cluster-GAN & Cluster-EBGAN \\ \midrule
         &  Purity &  0.8933 & 0.8933 & 0.8867 & 0.8973(0.041)  & {\bf 0.9333(0.023)} \\ 
\raisebox{1.5ex}{Iris}        & ARI & 0.7302 & 0.7312 & 0.5206&0.5694(0.169) & {\bf 0.8294(0.050)}\\\midrule
   &  Purity & 0.8905 & 0.8714 & - & 0.7686(0.049) & {\bf 0.9105(0.005)}\\ 
   
   \raisebox{1.5ex}{Seeds}        & ARI & 0.7049 & 0.6752 & - &0.4875(0.090) &  {\bf 0.7550(0.011)} \\\midrule
   &  Purity & 0.5776 & 0.7787 & - &0.7217(0.02) & {\bf 0.8826(0.02)} \\ 
   
\raisebox{1.5ex}{MNIST}        & ARI & 0.3607 & 0.5965& - & 0.5634(0.02) & {\bf 0.7780(0.03) }\\ \bottomrule
    \end{tabular}
\end{adjustbox}
\end{table}

 \vspace{-0.3in}
 \section{Conclusion}
 

This paper has identified the reasons why the GAN suffers from the mode collapse issue and proposes a new formulation to address this problem. Additionally, an empirical Bayes-like method is proposed for training the GAN under the new formulation. The proposed new formulation is general, allowing for easy reformulation and training of various GAN variants such as Lipschitz GAN and cluster GAN using the proposed empirical Bayes-like method. 
 
The proposed empirical Bayes-like method can be extended in various ways. 
For example, the generator can be simulated using other stochastic gradient MCMC algorithms such as SGHMC \citep{SGHMC2014} and preconditioned SGLD \citep{pSGLD}; and the discriminator can be trained using an advanced SGD algorithm such as Adam \citep{KingmaB2015},
AdaMax \citep{KingmaB2015} and Adadelta \citep{AdaDelta2012}.  
Moreover, the proposed method can be easily extended to learn sparse generators by imposing an appropriate prior distribution on the generator. Refer to \cite{SunSLiang2021sparseDNN,SunSLiang2022PRL} for prior settings for consistent sparse deep learning. 
 

In summary, this paper has presented a new formulation for the GANs as randomized decision problems, and proposed an effective method to solve them. From the perspective of statistical decision theory, further investigation into the application 
of the proposed method to other classes of risk functions would be of great interest. 
We anticipate that the proposed method will find wide applications in the field of statistical decision science.


\par
\section*{Acknowledgements}

This research is supported in part by the NSF grants DMS-1811812 (Song) and DMS-2015498 (Liang) and the 
NIH grant R01-GM126089 (Liang). The authors thank the reviewers for their insightful and helpful comments.

\par

\newpage

\appendix 

\begin{center}
\noindent {\bf \Large Supplementary Material}
\end{center}

 This supplementary material is organized as follows.
 Section S1 gives the proofs for the theoretical results of the paper. 
 Section S2 defines the metrics, including the inception score, Wasserstein distance, 
 and maximum mean discrepancy, used for quantifying the performance of image generation for different methods. 
 Section S3 presents more numerical examples, including those on image generation, conditional independence tests, and nonparametric clustering.
 Section S4 presents parameter settings used in the numerical experiments.
 

\setcounter{section}{0}
\setcounter{equation}{0}
\def\theequation{S\arabic{section}.\arabic{equation}}
\def\thesection{S\arabic{section}}

\setcounter{table}{0}
\renewcommand{\thetable}{S\arabic{table}}
\setcounter{figure}{0}
\renewcommand{\thefigure}{S\arabic{figure}}
\setcounter{lemma}{0}
\renewcommand{\thelemma}{S\arabic{lemma}}
\setcounter{theorem}{0}
\renewcommand{\thetheorem}{S\arabic{theorem}}
\setcounter{remark}{0}
\renewcommand{\theremark}{S\arabic{remark}}



\section{Theoretical Proofs} 

\subsection{Proof of Lemma \ref{minimax}} 

\begin{proof}
\begin{equation} \label{minimaxproofeq1}
\small
\begin{split}
\mathbb{E}_{\pi_g} \mathcal{J}_d(\theta_d; \theta_g) &= 
\int \phi_1(D_{\theta_d}(x)) p_{data}(x) dx + \int \int \phi_2(D_{\theta_d}(G_{\theta_g}(z))) q(z) \pi_g(\theta_g) dz d\theta_g 
\\
&=\int \phi_1(D_{\theta_d}(x)) p_{data}(x) dx + \int \int \phi_2(D_{\theta_d}(x)) p_{\theta_g}(x) \pi_g(\theta_g) d\theta_g dx 
\\
&= \int \left[\phi_1(D_{\theta_d}(x)) p_{data}(x) + \phi_2(D_{\theta_d}(x)) p_{\pi_g}(x)\right] dx,
\end{split}
\end{equation}
where the mixture generator formed by $\pi_g$ can be viewed as a single super generator 
$\theta_g^*$ such that $p_{\theta_g^*}(x)=p_{\pi_g}(x)$. 
Then, by the proof of Theorem 1 of \cite{Goodfellow2014}, we have 
 $\min_{\pi_g}\max_{\theta_d} \mathbb{E}_{\pi_g} \mathcal{J}_d(\theta_d; \theta_g)=-\log(4)$.
 It is easy to verify that at the Nash equilibrium point, $\mathbb{E}_{\tilde{\pi}_g} \mathcal{J}_d(\tilde{\theta}_d; \theta_g)=-\log(4)$. 

By the proof of Theorem 1 of \cite{Goodfellow2014}, if $$\tilde{\theta}_d=\arg\max_{\theta_d} \mathbb{E}_{\tilde{\pi}_g} \mathcal{J}_d(\theta_d; \theta_g)$$ holds,  then 
$\mathbb{E}_{\tilde{\pi}_g} \mathcal{J}_d(\tilde{\theta}_d; \theta_g)=-\log(4)$ 
implies the Jensen–Shannon divergence $JSD(p_{data}|p_{\tilde{\pi}_g})=0$ and thus 
$p_{\tilde{\pi}_g}=p_{data}$. Further, by Proposition 1 of \cite{Goodfellow2014}, 
we have $D_{\tilde{\theta}_d}(x)=1/2$ when $p_{\tilde{\pi}_g}=p_{data}$ holds. 
\end{proof}

\subsection{Proof of Theorem \ref{thm:00}}

\begin{proof} The proof consists of two steps. 
   First, we would prove that 
  \begin{equation} \label{Julyeq4}
  \int \mathcal{J}_d(\tilde{\theta}_d;\theta_g) \pi(\theta_g|\tilde{\theta}_d,\mathcal{D}) d\theta_g = -\log 4, \quad \mbox{as $N \to \infty$}.
\end{equation}   
  For the game (\ref{EBGANgame}), it is easy to see that 
 \begin{equation} \label{Julyeq5}
 \small
 \begin{split}
 &  \min_{\pi_g} \max_{\theta_d} \mathbb{E}_{\theta_g\sim \pi_g} \mathcal{J}_d(\theta_g;\theta_g)
  \leq \max_{\theta_d} \int \mathcal{J}_d(\theta_d;\theta_g) \pi(\theta_g|\theta_d,\mathcal{D}) d\theta_g = \int \mathcal{J}_d(\tilde{\theta}_d;\theta_g) \pi(\theta_g|\tilde{\theta}_d,\mathcal{D}) d\theta_g\\
   & = -\log 4 +\frac{1}{N} \int \{N (\mathcal{J}_d(\tilde{\theta}_d; \theta_g)+\log 4) -\log q_g(\theta_g)+\log Z(\tilde{\theta}_d)\} \pi(\theta_g|\tilde{\theta}_d,\mathcal{D}) d\theta_g \\ 
  & \quad  + \frac{1}{N} \int  \{\log q_g(\theta_g)\} \pi(\theta_g|\tilde{\theta}_d,\mathcal{D}) d\theta_g -\frac{1}{N} \log Z(\tilde{\theta}_d) \\
  &=-\log 4 +(I)+(II)+(III),\\
  \end{split}
  \end{equation}
  where $Z(\tilde{\theta}_d)$ is the normalizing constant of $\pi(\theta_g|\tilde{\theta}_d,\mathcal{D})$. 
  
 As implied by (\ref{minimaxproofeq1}), $\max_{\theta_d}  \int \mathcal{J}_d(\theta_d;\theta_g) \pi_g(\theta_g) d\theta_g$ is equivalent to $\max_{\theta_d}  \mathcal{J}_d(\theta_d;\theta_g)$ for a fixed generator $\theta_g$ that $p_{\theta_g}(x)=p_{\pi_g}(x)$ holds. 
 Therefore, by Theorem 1 of \cite{Goodfellow2014}, we have $\mathcal{J}_d(\tilde{\theta}_d; \theta_g) \geq -\log 4$ for any $\theta_g \in \Theta_g$.  
  That is, $N (\mathcal{J}_d(\tilde{\theta}_d; \theta_g)+\log 4) -\log q_g(\theta_g)$ can be treated as the energy of the posterior $\pi(\theta_g|\tilde{\theta}_d,\mathcal{D})$, and  then 
  \[
  (I)=-\frac{1}{N} \int \{\log \pi(\theta_g|\tilde{\theta}_d,\mathcal{D})\}  \pi(\theta_g|\tilde{\theta}_d,\mathcal{D})d \theta_g.
  \]
  By the Kullback-Leibler divergence $D_{KL}(\pi(\theta_g|\tilde{\theta}_d,\mathcal{D})|q_g)\geq 0$, 
  \[
  (II) \leq \frac{1}{N} \int \{\log \pi(\theta_g|\tilde{\theta}_d,\mathcal{D})\}  \pi(\theta_g|\tilde{\theta}_d,\mathcal{D})d \theta_g.
  \]
  As justified in Remark \ref{rem1}, $|\log Z(\tilde{\theta}_d)|$ is of the order $O(dim(\theta_g) \log {N})$ and thus $(III) \to 0$ as $N\to \infty$. Summarizing these terms, we have 
  \begin{equation} \label{Julyeq3b}
  \begin{split}
   \int \mathcal{J}_d(\tilde{\theta}_d;\theta_g) \pi(\theta_g|\tilde{\theta}_d,\mathcal{D}) d\theta_g 
   \stackrel{N\to \infty}{\leq} -\log 4.
   \end{split}
  \end{equation}
 By (\ref{Julyeq5}) and Lemma \ref{minimax}, we have  $$\int \mathcal{J}_d(\tilde{\theta}_d;\theta_g) \pi(\theta_g|\tilde{\theta}_d,\mathcal{D}) d\theta_g \geq  \min_{\pi_g} \max_{\theta_d} \mathbb{E}_{\theta_g\sim \pi_g} \mathcal{J}_d(\theta_g;\theta_g)=-\log 4.$$
 Combining it with (\ref{Julyeq3b}), we can conclude equation (\ref{Julyeq4}).  

Next, to apply Lemma \ref{minimax} to claim that $(\tilde{\theta}_d, p_{\tilde{\pi}_g})$ is a Nash equilibrium point, we still need to prove that $\tilde{\theta}_d$ is also the maximizer of $\int \mathcal{J}_d(\theta_d;\theta_g) \pi(\theta_g|\tilde{\theta}_d,\mathcal{D}) d\theta_g$.
We do this by proof of contradiction. 
Suppose $$\|\tilde{\theta}_d -\arg\max_{\theta_d} \int \mathcal{J}_d(\theta_d;\theta_g) \pi(\theta_g|\tilde{\theta}_d,\mathcal{D}) d\theta_g\|>\delta_0$$ 
for some $\delta_0>0$.
 Then, by Proposition 1 of \cite{Goodfellow2014}, there exist a function 
 $\epsilon(x)$ and a constant $\epsilon_0>0$ such that 
\[
D_{\tilde{\theta}_d}(x)=\frac{p_{data}(x)+\epsilon(x)}{p_{data}(x)+p_{\tilde{\pi}_g}(x)},
\]
and $|\epsilon(x)|>\epsilon_0$ on some non-zero measure set of $\mathcal{X}$, 
where $\mathcal{X}$ denotes the domain of $x$ and $-p_{data}(x) \leq \epsilon(x) \leq p_{\tilde{\pi}}(x)$ for ensuring $0\leq D_{\tilde{\theta}_d}(x) \leq 1$. 
Following the proof of Theorem 1 of \cite{Goodfellow2014}, we have 

\begin{equation} \label{Augeq1}
\small
\begin{split}
&\int \mathcal{J}_d(\tilde{\theta}_d;\theta_g)  \pi(\theta_g|\tilde{\theta}_d,\mathcal{D}) d\theta_g
 = \mathbb{E}_{x \sim p_{data}} \log \frac{p_{data}(x)+\epsilon(x)}{p_{data}(x)+p_{\tilde{\pi}_g}(x)} +  
\mathbb{E}_{x\sim p_{\tilde{\pi}_g}}  \log \frac{p_{\tilde{\pi}_g}(x)-\epsilon(x)}{p_{data}(x)+p_{\tilde{\pi}_g}(x)} \\ 
 & = -\log4 + 2 JSD(p_{data}|p_{\tilde{\pi}_g}) + \mathbb{E}_{x \sim p_{data}} \log(1+\frac{\epsilon(x)}{p_{data}(x)}) +  
\mathbb{E}_{x\sim p_{\tilde{\pi}_g}} \log(1-\frac{\epsilon(x)}{p_{\tilde{\pi}_g}(x)}). \\
\end{split}
\end{equation}

If $p_{\tilde{\pi}_g}=p_{data}$, then  $JSD(p_{data}|p_{\tilde{\pi}_g})=0$,
 $\mathbb{E}_{x \sim p_{data}} \log(1+\frac{\epsilon(x)}{p_{data}(x)}) +  
\mathbb{E}_{x\sim p_{\tilde{\pi}_g}} \log(1-\frac{\epsilon(x)}{p_{\tilde{\pi}_g}(x)}) <0$ by Jensen's inequality,  
and thus 
\[
\int \mathcal{J}_d(\tilde{\theta}_d;\theta_g) \pi(\theta_g|\tilde{\theta}_d,\mathcal{D}) d\theta_g < -\log 4.
\]
In what follows we show that this is in contradiction to (\ref{Julyeq4}) by showing that the $\tilde{\theta}_d$ corresponding to $p_{\tilde{\pi}_g}=p_{data}$ is a solution to the problem 
$\max_{\theta_d}\int \mathcal{J}_d(\theta_d;\theta_g) \pi(\theta_g|\theta_d,\mathcal{D}) d\theta_g$.

Suppose that $N$ is sufficiently large and $p_{\pi_g}=p_{data}$ holds, then we have: (i) $D_{\tilde{\theta}_d'}=1/2$ by (\ref{minimaxproofeq1}) and Proposition 1 of \cite{Goodfellow2014}, where $\tilde{\theta}_d'=\arg\max \mathbb{E}_{\pi_g} \mathcal{J}_d(\theta_d; \theta_g)$ with $p_{\pi_g}=p_{data}$; 
(ii) in the space of $p_{\theta_g}$ the posterior $\pi(\theta_g|\tilde{\theta}_d', \mathcal{D})$ has the mode at $p_{\theta_g}=p_{data}$ as $N \to \infty$ following from the arguments that $\mathcal{J}_g(\theta_g; \tilde{\theta}_d')$ is concave with respect to $p_{\theta_g}$ as shown in Proposition 2 of \cite{Goodfellow2014}, and that $\mathcal{J}_g(\theta_g; \tilde{\theta}_d')$ attains its maximum at $p_{\theta_g}=p_{data}$ by Theorem 1 of \cite{Goodfellow2014};  and (iii) $\mathcal{J}_d(\tilde{\theta}_d';\theta_g)=-\log 4$ at  the posterior mode $p_{\theta_g}=p_{data}$. 
Then, by Laplace approximation \citep{KassTK1990}, we have 
$\int \mathcal{J}_d(\tilde{\theta}_d';\theta_g) \pi(\theta_g|\tilde{\theta}_d',\mathcal{D}) d\theta_g \to -\log 4$  and $p_{\tilde{\pi}_g'}=\int p_{\theta_g} \pi(\theta_g|\tilde{\theta}_d',\mathcal{D}) d\theta_g =p_{data}$ as $N\to \infty$. 
That is, the $\tilde{\theta}_d$ corresponding to $p_{\tilde{\pi}_g}=p_{data}$ (changing the notations 
$\tilde{\theta}_d'$ to $\tilde{\theta}_d$ and $p_{\tilde{\pi}_g'}$ to $p_{\tilde{\pi}_g}$) is indeed a maximizer of $\int \mathcal{J}_d(\theta_d;\theta_g) \pi(\theta_g|\theta_d,\mathcal{D}) d\theta_g$ as $N \to \infty$. 
 Note that $\pi(\theta_g|\tilde{\theta}_d, \mathcal{D})$ may contain multiple equal modes in the space of $\theta_g$ due to the nonidentifiability of the neural network model, which does not affect the validity of the above arguments.
 Therefore, by the contradiction, we can conclude that 
$\tilde{\theta}_d = \arg\max_{\theta_d} \int \mathcal{J}_d(\theta_d;\theta_g) \pi(\theta_g|\tilde{\theta}_d,\mathcal{D}) d\theta_g$ by the arbitrariness of $\delta_0$.
 
  
The proof can then be concluded by Lemma \ref{minimax} with the results of the above two steps.
\end{proof}

\begin{remark} \label{rem1}
The order of $|\log Z(\tilde{\theta}_d)|$ given in the proof of Theorem \ref{thm:00} can be justified based on Laplace approximation \citep{KassTK1990}, and the justification can be extended to any fixed value of $\theta_d$. 
Let $c=\min_{\theta_g\in \Theta_g} \mathcal{J}_d(\theta_d; \theta_g)$ for any fixed value of $\theta_d$.  Applying the Laplace approximation 
to  the integral $\int \exp\{-N(\mathcal{J}_d(\theta_d; \theta_g)-c)\} q_g(\theta_g)d \theta_g$, we have 
\begin{equation} \label{Zeq1}
Z(\theta_d) =(2\pi)^{dim(\theta_g)/2} [det(N \bH_e)]^{-1/2} \exp\{-N (\mathcal{J}_d(\theta_d;\hat{\theta}_g)-c)\} q_g(\hat{\theta}_g)\Big(1+O(\frac{1}{N})\Big),
\end{equation}
where $\hat{\theta}_g=\arg\max_{\theta_g\in \Theta_g} \{-(\mathcal{J}_d(\theta_d,\theta_g)-c)+\frac{1}{N}\log q_g(\theta_g)\}$, $\bH_e$ is the Hessian of $\mathcal{J}_d(\theta_d;\theta_g)-c-\frac{1}{N}\log q_g(\theta_g)$ evaluated at $\hat{\theta}_g$, and 
$det(\cdot)$ denotes the determinant operator. By the convexity of  $\mathcal{J}_d(\theta_d,\theta_g)$ (with respect to $p_{\theta_g}$ as shown in Proposition 2 of \cite{Goodfellow2014}) and the boundedness of the prior density function by Assumption (ii) of Theorem \ref{thm:00}, it is easy to see that $N(\mathcal{J}_d(\theta_d,\hat{\theta}_g)-c)-\log q_g(\hat{\theta}_g)$ is finite and thus $(\mathcal{J}_d(\theta_d;\hat{\theta}_g)-c)-\frac{1}{N} \log q_g(\hat{\theta}_g) \to 0$ as $N\to \infty$.
If all the eigenvalues of $\bH_e$ are bounded by some positive constants, then 
$-\frac{1}{N}\log Z(\theta_d)=O(dim(\theta_g)\log N/N)=o(1)$.  
Finally, we note that the analytical assumptions for Laplace's method \citep{KassTK1990} can be verified based on the convexity of $\mathcal{J}_d(\theta_d,\theta_g)$ and some mild assumptions on the derivatives of $\mathcal{J}_d(\theta_d,\theta_g)-c-\frac{1}{N} \log q_g(\theta_g)$ at $\hat{\theta}_g$; and that the posterior may contain multiple equal modes in the space of $\theta_g$ due to the nonidentifiability of the neural network model, which does not affect the validity of the above approximation.
\end{remark}

\subsection{Proof of Corollary \ref{corphi3}} 

\begin{proof}
Extension of Theorem \ref{thm:00} to the case $\phi_3(D)=\log(D)$ can be justified as follows. 
Let $$\pi'(\theta_g|\tilde{\theta}_d,\mathcal{D})=\exp\{N(-\mathbb{E}_{x\sim p_{data}} \phi_1(D_{\tilde{\theta}_d}(x))+ \mathbb{E}_{x\sim p_{\theta_g}} \phi_3(D_{\tilde{\theta}_d}(x))\} q_g(\theta_g)/Z'(\tilde{\theta}_d)$$
for $\phi_3(D)=\log(D)$, and let 
$$\pi(\theta_g|\tilde{\theta}_d,\mathcal{D})= \exp\{N ( -\mathbb{E}_{x\sim p_{data}} \phi_1(D_{\tilde{\theta}_d}(x))+\mathbb{E}_{x\sim p_{\theta_g}} \phi_3(D_{\tilde{\theta}_d}(x))-c)\} q_g(\theta_g)/Z(\tilde{\theta}_d)$$ 
for $\phi_3(D)=-\log(1-D)$, where $c=-\log4$, and 
 $Z'(\tilde{\theta}_d)$ and $Z(\tilde{\theta}_d)$ denote their respective normalizing constants.
  Then 
\[
\small
\begin{split}
& \int \mathcal{J}_d(\tilde{\theta}_d;\theta_g)  \pi'(\theta_g|\tilde{\theta}_d,\mathcal{D}) =c+ \frac{1}{N}
\int \left[ N(\mathcal{J}_d(\tilde{\theta}_d;\theta_g)-c)-\log q_g(\theta_g)+\log Z(\tilde{\theta}_d)\right] \pi'(\theta_g|\tilde{\theta}_d,\mathcal{D}) d \theta_g \\
 & \quad \quad + \frac{1}{N} \int \log q_g(\theta_g) \pi_g'(\theta_g|\tilde{\theta}_d,\mathcal{D})d\theta_g -\frac{1}{N} \log Z(\tilde{\theta}_d) \\
& \leq c+\frac{1}{N} \int \left[-\log\pi(\theta_g|\tilde{\theta}_d,\mathcal{D}) + 
\log \pi'(\theta_g|\tilde{\theta}_d,\mathcal{D}) \right] \pi'(\theta_g|\tilde{\theta}_d,\mathcal{D}) d\theta_g-\frac{1}{N} \log Z(\tilde{\theta}_d) \\
& = c+(I) +(II), \\ 
\end{split}
\]
where the inequality follows from that the Kullback-Leibler divergence $D_{KL}(\pi_g'|q_g)\geq 0$. 

By Remark \ref{rem1}, we have $(II) \to 0$ as $N\to \infty$. 
The term (I) is the Kullback-Leibler divergence between $\pi'(\theta_g|\tilde{\theta}_d,\mathcal{D})$ and 
$\pi(\theta_g|\tilde{\theta}_d,\mathcal{D})$. 
By the upper bound of the Kullback-Leibler divergence \citep{DRAGOMIR2000}, we have 
\[
\small
\begin{split}
(I)  & \leq \frac{1}{N} \int \frac{\pi'(\theta_g|\tilde{\theta}_d,\mathcal{D})}{\pi(\theta_g|\tilde{\theta}_d,\mathcal{D})} \pi'(\theta_g|\tilde{\theta}_d,\mathcal{D}) d\theta_g- \frac{1}{N}  \\
 & = \frac{1}{N} \times \frac{Z(\tilde{\theta}_d)}{Z'(\tilde{\theta}_d)}  \times
 \int \prod_{x_i \sim p_{\theta_g}, i=1,2,\ldots,N} [4 D_{\tilde{\theta}_d}(x_i)(1-D_{\tilde{\theta}_d}(x_i))]  \pi'(\theta_g|\tilde{\theta}_d,\mathcal{D}) d\theta_g
 - \frac{1}{N} \\
 &=\frac{1}{N}\times (I_1)\times (I_2)-\frac{1}{N}.
 \end{split}
 \]
 
Since $4 D_{\tilde{\theta}_d}(x_i)(1-D_{\tilde{\theta}_d}(x_i)) \leq 1$ for each $x_i$, 
we have $(I_2) \leq 1$. 
Next, we consider the term $(I_1)$. For both choices of $\phi_3$, as implied by (\ref{minimaxproofeq1}) where the mixture generator proposed in the paper is represented as a single super generator, the arguments in \cite{Goodfellow2014} on the non-saturating case can be applied here, and thus $\pi(\theta_g|\tilde{\theta}_d,\mathcal{D})$ and $\pi'(\theta_g|\tilde{\theta}_d,\mathcal{D})$ have the same maximum {\it a posteriori} (MAP) estimate $\hat{\theta}_g$ as $N\to \infty$. Further, 
by Lemma \ref{minimax}, we have $D_{\tilde{\theta}_d}(x)=1/2$ for any $x\in p_{data}$. 
Then it is easy to see that $\log \pi(\theta_g|\tilde{\theta}_d,\mathcal{D})$ and $\log \pi'(\theta_g|\tilde{\theta}_d,\mathcal{D})$
have exactly the same first and second gradients at $(\tilde{\theta}_d,\hat{\theta}_g)$, which implies that they have the same Hessian matrix. Therefore, by (\ref{Zeq1}), $(I_1)=Z(\tilde{\theta}_d)/Z'(\tilde{\theta}_d)\to 1$ as $N\to \infty$.   
Summarizing $(I_1)$ and $(I_2)$, we have $(I)\to 0$ as $N\to \infty$. 
Summarizing all the above arguments, we have $\int \mathcal{J}_d(\tilde{\theta}_d;\theta_g)  \pi_g'(\theta_g|\tilde{\theta}_d,\mathcal{D}) \to -\log 4$ as $N\to \infty$.   

 The proof for $\tilde{\theta}_d =\arg\max_{\theta_d} \int \mathcal{J}_d(\theta_d;\theta_g) \pi'(\theta_g|\tilde{\theta}_d,\mathcal{D}) d\theta_g$ is similar to step 2 of the proof of 
 Theorem \ref{thm:00}. The corollary can then be concluded. 
\end{proof}

\subsection{Adaptive Stochastic Gradient MCMC}

Consider to solve the mean field equation:   
\begin{equation} \label{rooteq}
    h(\theta)=\int_{\mX} H(\theta,\beta)\pi(\beta|\theta)d\beta=0,
\end{equation}
where $\beta \in \mX$ can be viewed a latent variable. 
Following \cite{DengLiang2019}, we propose the following adaptive stochastic gradient MCMC algorithm for solving the equation (\ref{rooteq}):

\begin{algorithm}[htbp]
\caption{An adaptive stochastic gradient MCMC algorithm} 
\label{SAmcmcalg}
\begin{enumerate}
    \item $\beta_{k+1}=\beta_k+\epsilon_{k+1} (\nabla_{\beta}\tilde{L}(\beta_k,\theta_k)+\rho_{k} m_k)+\sqrt{2\epsilon\tau}\mathcal{N}(0,I)$,
    
    \item $m_{k+1}=\alpha m_k+(1-\alpha) \nabla_{\beta}\tilde{L}(\beta_k,\theta_k)$,
    
    \item $\theta_{k+1}=\theta_k+w_{k+1}H(\theta_k,\beta_{k+1})$,
\end{enumerate}
\end{algorithm}

In this algorithm, MSGLD \citep{kim2020stochastic} is used in drawing samples of $\beta$, $\nabla_{\beta}\tilde{L}(\beta_k,\theta_k)$ denotes 
an unbiased estimator of $\nabla_{\beta} \log \pi(\beta|\theta_k)$ obtained with the sample $\beta_k$, $\epsilon_{k+1}$ is called the learning rate used at iteration $k+1$, $\tau$ is the temperature, $w_{k+1}$ is the step size used at iteration $k+1$, $\alpha$ is the momentum smoothing factor, 
and $\rho_k$ is the momentum biasing factor.  
The algorithm is said ``adaptive'', as the parameter $\theta$ changes along with iterations.

\paragraph*{\it Notations}  
Algorithm \ref{SAmcmcalg} has the following notational correspondence with the EBGAN: $(\beta,\theta)$ in Algorithm \ref{SAmcmcalg} corresponds to 
 $(\theta_g,\theta_d)$ in the EBGAN; equation (\ref{rooteq}) corresponds to  
\[
h(\theta_d)= \int  H(\theta_d,\theta_g)  
  \pi(\theta_g|\theta_d,\mathcal{D}) d\theta_g=0,
 \]
where $H(\theta_d; \theta_g)$ is as defined in (\ref{gradeq}),   
and $\pi(\theta_g|\theta_d,\mathcal{D}) \propto  \exp(\mathbb{J}_g(\theta_g;\theta_d)) q_g(\theta_g)$;  
$L(\beta,\theta)$ corresponds to $\log \pi(\theta_g|\theta_d,\mathcal{D})$ (up to an additive constant),   and the stochastic gradient $\nabla_{\beta} \tilde{L}(\beta,\theta)$ in Algorithm \ref{SAmcmcalg} corresponds to $\nabla_{\theta_g} \tilde{L}(\theta_g,\theta_d)$ defined in (\ref{gradeq}).

\subsection{Convergence of the discriminator} 

To establish convergence of $\{\theta_k\}$ for Algorithm \ref{SAmcmcalg}, we make the following assumptions. 

\begin{assumption} \label{ass1} (Conditions on stability and $\{\omega_k\}_{k \in \mathbb{N}}$)
 There exist a constant $\delta$ and a stationary point $\theta^*$ such that $\langle \theta-\theta^*, h(\theta)\rangle\leq -\delta \|\theta-\theta^*\|^2$ for any $\theta\in \Theta$.
 The step sizes $\{w_k\}_{k\in \mathbb{N}}$ form a positive decreasing sequence such that 
\begin{equation} \label{ass1eq}
    w_k\to 0,\quad\sum_{k=1}^{\infty}w_k=+\infty, \quad\liminf_{k\to \infty} 2\delta\frac{w_k}{w_{k+1}}+\frac{w_{k+1}-w_k}{w_{k+1}^2}>0.
\end{equation}
\end{assumption}

Similar to \cite{Benveniste1990} (p.244), we can show that the following choice of $\{w_k\}$ satisfying (\ref{ass1eq}):
\begin{equation}
    w_k=c_1/(c_2+k)^{\zeta_1},
\end{equation}
for some constants $c_1>0$, $c_2 \geq 0$ and $\zeta_1 \in (0,1]$, provided that $c_1$ has been chosen large enough such that $2 \delta c_1> 1$ holds.

\begin{assumption} \label{ass2} (Smoothness and Dissipativity) $L(\beta,\theta)$ is \textit{M-smooth} on $\theta$ and $\beta$, and $(m,b)$-dissipative on $\beta$. In other words, for any $\beta,\beta_1,\beta_2\in \mathcal{X}$ and $\theta_1,\theta_2 \in\Theta$, the following inequalities hold:
\begin{align}
    \|\nabla_{\beta}L(\beta_1,\theta_1)- \nabla_{\beta}L(\beta_2,\theta_2) \| &\leq M \|\beta_1-\beta_2 \|  + M\| \theta_1-\theta_2 \|, \\
    \langle \nabla_{\beta}L(\beta,\theta),\beta \rangle &\leq b-m\|\beta\|^2.
\end{align}
\end{assumption}
 
Let $\beta^*$ be a maximizer  such that $\nabla_{\beta}L(\beta^*,\theta^*)=0$, where $\theta^*$ is the stationary point defined in Assumption \ref{ass1}. By the dissipativity in Assumption \ref{ass2}, we have $\|\beta^*\|^2\leq \frac{b}{m}$. Therefore, 
\begin{equation*}
\begin{split}
    \|\nabla_{\beta}L(\beta,\theta)\|&\leq \| \nabla_{\beta}L(\beta^*,\theta^*)\|+M\|\beta^*-\beta\|+M\|\theta-\theta^*\| \\
    & \leq M\|\theta\|+M\|\beta\|+\bar{B},
\end{split}    
\end{equation*}
where $\bar{B}=M(\sqrt{\frac{b}{m}}+\|\theta^*\|)$. This further implies 
\begin{equation} \label{prop1}
\|L_{\beta}(\beta,\theta)\|^2\leq 3M^2\|\beta\|^2+3M^2\|\theta\|^2+3\bar{B}^2.
\end{equation}

\begin{assumption} \label{ass3}  (Noisy gradient) Let $\xi_k=\nabla_{\beta}\tilde{L}(\beta_k,\theta_k)- \nabla_{\beta} L(\beta_k,\theta_k)$ denote the white noise contained in the stochastic gradient. The  white noises  
$\xi_1, \xi_2,\ldots$ are mutually independent 
 and satisfy the conditions: 
\begin{equation}
    \begin{split}
    &E(\xi_k|\mathcal{F}_k)=0,   \\ 
    &E\|\xi_k\|^2\leq M^2E\|\beta\|^2+M^2E\|\theta\|^2+B^2,
    \end{split}
\end{equation}
where $\mathcal{F}_k=\sigma\{\theta_1,\beta_1,\theta_2,\beta_2,\ldots\}$ denotes a $\sigma$-filter. 
\end{assumption}

The smoothness, dissipativity and noisy gradient conditions are regular for studying the convergence of stochastic gradient MCMC algorithms. 
Similar conditions have been used in many existing works such as \cite{raginsky2017non}, \cite{DengLiang2019}, and \cite{gao2021global}.  

\begin{assumption} \label{ass4} (Boundedness)
Assume that the trajectory of $\theta$ belongs to a compact set $\Theta$, i.e. $\{\theta_k\}_{k=1}^{\infty}\subset \Theta$ and $\|\theta_k\|\leq M$ for some constant $M$.  
\end{assumption}

This assumption is more or less a technical condition. Otherwise, we can show that the Markov transition kernel used in Algorithm \ref{SAmcmcalg} satisfies the drift condition and, therefore, the varying truncation technique (see e.g. \cite{ChenZhu1986,AndrieuMP2005}) can be employed in the algorithm for ensuring that $\{\theta_k: k=1,2,\ldots\}$ is almost surely contained in a compact space.  
 
\begin{lemma} (Uniform $L_2$ bound) \label{l2bound} Suppose Assumptions \ref{ass1}-\ref{ass4} hold. Given a sufficiently small learning rate $\epsilon$, we have 
\[
\begin{split}
& \sup_t E\|\beta_t\|^2\leq G_{\beta}, \\
& \sup_t E\langle \beta_{t}, m_{t} \rangle\leq G_m,  
\end{split}
\]
for some constants $G_{\beta}$ and $G_m$.
\end{lemma}
\begin{proof}
We prove this lemma by mathematical induction under the weakest condition that both $\epsilon_t$ and $\rho_t$ are set to constants. Assume that $E\|\beta_t\|^2\leq G_{\beta}$ and $E\langle \beta_{t}, m_{t} \rangle\leq G_{m}$ for all $t=1,\dots,k$. By Algorithm \ref{SAmcmcalg}, we have 
\begin{equation}\label{eq-1}
    \begin{split}
    & E\|\beta_{k+1}\|^2 =E\|\beta_k+\epsilon[\nabla_{\beta}\tilde{L}(\beta_k,\theta_k)+ \rho m_k]\|^2+2\tau\epsilon d \\
      &=E\|\beta_k+\epsilon[\nabla_{\beta}L(\beta_k,\theta_k)+ \rho m_k] \|^2+\epsilon^2 E\|\xi_k\|^2+2\tau\epsilon d \quad (\mbox{by Assumption \ref{ass3}})\\
        &=E\|\beta_k\|^2+2\epsilon E\langle \beta_k, \nabla_{\beta}L(\beta_k,\theta_k)\rangle
        +2 \rho \epsilon E\langle \beta_k, m_k\rangle+\epsilon^2E\|\nabla_{\beta}{L}(\beta_k,\theta_k)+ \rho m_k\|^2\\
        &\quad +\epsilon^2 (M^2E\|\beta_k\|^2+M^2E\|\theta_k\|^2+B^2)+2\tau\epsilon d,
    \end{split}
\end{equation}
where $d$ is the dimension of $\beta$. Further, 
we can show that $m_k=(1-\alpha)\nabla_\beta \tilde L(\beta_{k-1},\theta_{k-1})+\alpha(\alpha-1)\nabla_\beta \tilde L(\beta_{k-2},\theta_{k-2})+\alpha^2(\alpha-1)\nabla_\beta \tilde L(\beta_{k-3},\theta_{k-3})+\cdots$. 
By Assumption \ref{ass2}-\ref{ass3} and equation (\ref{prop1}), for any $i\geq1$, we have $E\|\nabla_\beta \tilde L(\beta_{k-i},\theta_{k-i})\|^2
\leq E\|\nabla_\beta L(\beta_{k-i},\theta_{k-i})\|^2+E\|\xi_{k-i}\|^2\leq 4M^2E\|\beta_{k-i}\|^2+4M^2E\|\theta\|^2+3\bar{B}^2+B^2\leq 4M^2G_{\beta}+4M^4+3\bar{B}^2+B^2$. Therefore,
\begin{equation} \label{momeq}
\small
    \begin{split}
        &E\|m_k\|^2=\sum_{i=1}^k [(1-\alpha)\alpha^{i-1}]^2E\|\nabla_\beta \tilde L(\beta_{k-i},\theta_{k-i})\|^2\\
        &+2\sum_{1\leq i,j\leq k} [(1-\alpha)\alpha^{i-1}][(1-\alpha)\alpha^{j-1}]\sqrt{E\|\nabla_\beta \tilde L(\beta_{k-i},\theta_{k-i})\|^2}\sqrt{ E\|\nabla_\beta \tilde L(\beta_{k-j},\theta_{k-j})\|^2}\\
        &\leq 4M^2G_{\beta}+4M^4+3\bar{B}^2+B^2.
    \end{split}
\end{equation}
Combined with (\ref{eq-1}), this further implies 
\begin{equation}\label{eq-2}
\small
    \begin{split}
        &E\|\beta_{k+1}\|^2\leq E\|\beta_k\|^2+2\epsilon E(b-m\|\beta_k\|^2)
        +2 \rho \epsilon G_{m}\\
        &\quad+2\epsilon^2(3M^2E\|\beta_k\|^2+3M^4+3\bar{B}^2)+2\epsilon^2 \rho^2(4M^2G_{\beta}+4M^4+3\bar{B}^2+B^2)\\
        &\quad +\epsilon^2 (M^2E\|\beta_k\|^2+M^2E\|\theta_k\|^2+B^2)+2\tau\epsilon d\\
        &=(1-2\epsilon m+7M^2\epsilon^2)E\|\beta_k\|^2+2\epsilon b+2 \rho \epsilon G_{m}+2\tau\epsilon d+2\epsilon^2(3M^4+3\bar B^2)\\
        &\quad+2\epsilon^2 \rho^2(4M^2G_{\beta}+4M^4+3\bar{B}^2+B^2)+\epsilon^2(M^4+B^2).
    \end{split}
\end{equation}
On the other hand,
\begin{equation}\label{eq-3}
\small
\begin{split}
    &E\langle \beta_{k+1}, m_{k+1}\rangle=E\langle \beta_{k}+\epsilon[\nabla_{\beta}\tilde{L}(\beta_k,\theta_k)+\rho m_k], \alpha m_{k}+(1-\alpha)\nabla_{\beta}\tilde{L}(\beta_k,\theta_k)\rangle\\
    \leq& \alpha E\langle \beta_{k},m_k\rangle+E\langle \beta_{k},(1-\alpha)\nabla_{\beta}{L}(\beta_k,\theta_k)\rangle+\epsilon(1+\rho) \max\{E\|\nabla_\beta\tilde{L}(\beta_k,\theta_k)\|^2,E\|m_k\|^2\}\\
    \leq& \alpha G_{m}+(1-\alpha) b +\epsilon(1+\rho)( 4M^2G_{\beta}+4M^4+3\bar{B}^2+B^2).
\end{split}
\end{equation}

To induce mathematical induction, following from (\ref{eq-2}) and (\ref{eq-3}), it is sufficient to show 
\[
\small
\begin{split}
G_{\beta}\leq& \frac{1}{2\epsilon m-7M^2\epsilon^2-8\epsilon^2 \rho^2 M^2}\bigg\{2\epsilon b+2\rho\epsilon G_{m}+2\tau\epsilon d\\
&+2\epsilon^2(3M^4+3\bar B^2)+2\epsilon^2 \rho^2(4M^4+3\bar{B}^2+B^2)+\epsilon^2(M^4+B^2)\bigg\},\\
G_{m}\leq&\frac{1}{1-\alpha}\bigg\{(1-\alpha) b +\epsilon(1+\rho)( 4M^2G_{\beta}+4M^4+3\bar{B}^2+B^2)\bigg\}.
\end{split}
\]
When $\epsilon$ is sufficiently small, it is not difficult to see that the above inequalities holds for some $G_{\beta}$ and $G_{m}$.
\end{proof}


\begin{assumption} \label{ass4b} (Lipschitz condition of $H(\theta,\beta)$) $H(\theta,\beta)$ is Lipschitz continuous on $\beta$; i.e., there exists a constant $M$ such that 
\[
 \|H(\theta,\beta_1)- H(\theta,\beta_2) \| \leq M \|\beta_1-\beta_2 \|.
\]
\end{assumption}


By Assumption \ref{ass4b},  
$\|H(\theta_k,\beta_{k+1}) \|^2 \leq 2M \|\beta_{k+1} \|^2+ 2\|H(\theta_k,0)\|^2$.
Since $\theta_{k}$ belongs to a compact set and $H(\theta,0)$ is a continuous function, 
there exists a constant $B$ such that 
\begin{equation} \label{Hbound}
\|H(\theta_k,\beta_{k+1}) \|^2 \leq 2M^2 \|\beta_{k+1} \|^2+ 2B^2. 
\end{equation}

\begin{assumption} \label{ass5} (Solution of Poisson equation)
For any $\theta\in\Theta$, $\beta \in \mX$, and a function $V(\beta)=1+\|\beta\|$, 
there exists a function $\mu_{\theta}(\beta)$ that solves the Poisson equation
 $\mu_{\theta}(\beta)-\mathcal{T}_{\theta}\mu_{\theta}(\beta)=H(\theta,\beta)-h(\theta)$ such that 
\begin{equation} \label{poissoneq}
    H(\theta_k,\beta_{k+1})=h(\theta_k)+\mu_{\theta_k}(\beta_{k+1})-\mathcal{T}_{\theta_k}\mu_{\theta_k}(\beta_{k+1}), \quad k=1,2,\ldots,
\end{equation}
where $\mathcal{T}_{\theta}$ is the probability transition kernel and 
$\mathcal{T}_{\theta}\mu_{\theta}(\beta)=\int \mu_{\theta}(\beta') \mathcal{T}_{\theta}(\beta,d\beta')$.
 Moreover, for all $\theta, \theta'\in \Theta$ and $\beta \in \mX$, we have 
$\|\mu_{\theta}(\beta)-\mu_{\theta'}(\beta)\|   \leq \varsigma_1 \|\theta-\theta'\| V(\beta)$ and $\|\mu_{\theta}(\beta) \| \leq \varsigma_2 V(\beta)$ for some constants $\varsigma_1>0$ and $\varsigma_2>0$. 
\end{assumption}

This assumption has often been used in the study for the convergence of the SGLD algorithm, see e.g. \cite{Whye2016ConsistencyAF} and \cite{DengLiang2019}. Alternatively, as mentioned above, we can show that the Markov transition kernel used in Algorithm \ref{SAmcmcalg} satisfies the drift condition and thus Assumption \ref{ass5} can be verified as in \cite{AndrieuMP2005}.

\paragraph*{Proof of Lemma \ref{thm1}} 
\begin{proof}
Our proof follows the proof of Theorem 1 in \cite{DengLiang2019}. However, Algorithm \ref{SAmcmcalg} employs MSGLD for updating $\beta$, while \cite{DengLiang2019} employs SGLD. We replace Lemma 1 of \cite{DengLiang2019} by Lemma \ref{l2bound} to accommodate this difference. In addition, Proposition 3 and Proposition 4 in \cite{DengLiang2019} are replaced by equation (\ref{prop1}) and equation (\ref{Hbound}) respectively. 
 
Further, based on the proof of \cite{DengLiang2019}, we can derive an explicit formula for $\gamma$:
\begin{equation} \label{rate_bound}
\gamma=\gamma_0+ 12\sqrt{3} M \left((2M^2+\varsigma_2^2) G_{\beta}+2B^2+\varsigma_2^2  \right)^{\frac{1}{2}},
\end{equation}
where $\gamma_0$ together with $t_0$ can be derived from Lemma 3 of \cite{DengLiang2019} and they depend on $\delta$ and $\{\omega_t\}$ only. 
 The second term of $\gamma$ is  
 obtained by applying the Cauchy-Schwarz inequality to bound the expectation   
$E \langle \theta_t-\theta^*, \mathcal{T}_{\theta_{t-1}}\mu_{\theta_{t-1}}(\beta_t) \rangle$, where  $E\|\theta_t-\theta^*\|^2$ can be bounded by $2M^2$ by Assumption \ref{ass4} 
and $E \|\mathcal{T}_{\theta_{t-1}}\mu_{\theta_{t-1}}(\beta_t)\|^2$ can be bounded according to (\ref{poissoneq}) and the upper bound of $H(\theta,\beta)$ given in (\ref{Hbound}).   
\end{proof}

\subsection{Convergence of the Generator}

To establish the weak convergence of $\beta_t$ in Algorithm \ref{SAmcmcalg}, we need more assumptions. Let the fluctuation between $\psi$ and $\bar{\psi}$: 
\begin{equation} \label{poissoneqf}
    \mathcal{L}f(\theta)=\psi(\theta)-\bar \psi,
\end{equation}
where $f(\theta)$ is the solution to the Poisson equation, and $\mathcal{L}$ is the infinitesimal generator of the Langevin diffusion 

\begin{assumption} \label{ass7} 
Given a sufficiently smooth function $f(\theta)$ as defined in (\ref{poissoneqf}) and a function $\mathcal{V}(\theta)$ such that the derivatives satisfy the inequality $\|D^j f\|\lesssim \mathcal{V}^{p_j}(\theta)$ for some constant
 $p_j>0$, where $j\in\{0,1,2,3\}$. In addition, $\mathcal{V}^p$ has a bounded expectation, i.e., $\sup_{k} E[\mathcal{V}^p(\theta_k)]<\infty$;
 and $\mathcal{V}^p$ is smooth, i.e. $\sup_{s\in (0, 1)} \mathcal{V}^p(s\theta+(1-s)\vartheta)\lesssim \mathcal{V}^p(\theta)+\mathcal{V}^p(\vartheta)$ for all $\theta,\vartheta\in\Theta$ and $p\leq 2\max_j\{p_j\}$.
 \end{assumption} 
 

\paragraph*{Proof of Lemma \ref{thm2}}
\begin{proof}
The update of $\beta$ can be rewritten as 
\[\beta_{k+1}=\beta_k+\epsilon_{k+1} (\nabla_{\beta}{L}(\beta_k,\tilde{\theta}_d)+\Delta \tilde{V}_k)+\sqrt{2\epsilon\tau}\mathcal{N}(0,I),
\]
where $\Delta \tilde{V}_k=\nabla_{\beta}{L}(\beta_k,\theta_k)-\nabla_{\beta}{L}(\beta_k,\tilde{\theta}_d)+\xi_k+\rho_k m_k$ can be viewed as the estimation error of 
$\nabla_{\beta} \tilde{L}(\beta_k,\theta_k)$
for the ``true'' gradient $\nabla_{\beta}{L}(\beta_k,\tilde{\theta}_d)$. For the terms in $\Delta \tilde{V}_k$, by Lemma \ref{thm1} and Assumption \ref{ass2}, we have 
\[
\mathbb{E}\|\nabla_{\beta}{L}(\beta_k,\theta_k)-\nabla_{\beta}{L}(\beta_k,\tilde{\theta}_d)\|
\leq M \mathbb{E} \|\theta_k-\tilde{\theta}_d\| \leq M \sqrt{\gamma \omega_k} \to 0;
\]
by Assumption 3 and Lemma \ref{l2bound}, $\mathbb{E} \|\xi_k\|^2\leq M^2 \mathbb{E} \|\beta\|^2+M^2 \mathbb{E} \|\theta\|^2+B^2$ is upper bounded; 
and as implied by (\ref{momeq}), there exists a constant $C$ such that $\mathbb{E} \|\rho_k m_k\| \leq C \rho_k$.    
Then parts (i) and (ii) can be concluded by applying Theorem 5 and Theorem 3 of \cite{chen2015convergence}, respectively, where the proofs only need to be slightly modified to accommodate the convergence of $\theta_k\to \tilde{\theta}_d$ (as shown in Lemma \ref{thm1}) and the momentum biasing factor $\rho_k$.   
\end{proof}


\section{Evaluation Metrics for Generative Adversarial Networks} \label{metricsection}

The inception scores (IS) \citep{Salimans2016ImprovedTF},  Wasserstein distance (WD), 
and maximum mean discrepancy (MMD) 
are metrics that are often used for assessing the quality of images generated by a generative image model. See \cite{Xu2018ESGAN} for an empirical evaluation on them. 

Let $p_{g}(x)$ be a probability distribution of the images generated by the model, and let $p_{dis}(y|x)$ be the probability that image $x$ has label $y$ according to a pretrained discriminator.
The IS of $p_{gen}$ relative to $p_{dis}$ is given by 
\[
IS(p_{g})=\exp\left\{ \mathbb{E}_{x\sim p_{g}} D_{KL}( p_{dis}(y|x) | \int p_{dis}(y|x) p_{g}(x)dx \right\},
\]
which takes values in the interval $[1,m]$ with $m$ being the total number of possible labels. A higher IS value is preferred as it means $p_{g}$ is a sharp and distinct collection of images. 
 To calculate IS, we employed transfer learning to obtain a pretrained discriminator, 
 which involves retraining the pretrained ResNet50, the baseline model, on Fashion MNIST 
 data by tuning the weights on the first and last hidden layers.

The first moment Wasserstein distance, denoted by 1-WD in the paper, for the two distributions $p_g$ and $p_{data}$ is defined as 
\[
WD(p_g, p_{data})=\inf_{\gamma \in \Gamma(p_g,p_{data})} \mathbb{E}_{x_g\sim p_g, x_r \sim p_{data}} \|x_g-x_r\|,
\]
where $\Gamma(p_g,p_{data})$ denotes the set of all joint distributions with the respective marginals $p_g$ and $p_{data}$.  The 1-WD also refers to the earth mover's distance.
Let $\{x_{g,i}: i=1,2,\ldots,n\}$ denote $n$ samples drawn from $p_g$, and let $\{x_{r,i}: i=1,2,\ldots,n\}$ denote $n$ samples drawn from $p_{data}$. 
With the samples, the 1-WD can be calculated by solving the optimal transport problem: 
\[
\begin{split}
WD(p_g, p_{data}) & =\min_{w \in \mathbb{R}^{n \times n}} \sum_{i=1}^n \sum_{j=1}^n w_{ij} 
\|x_{g,i}-x_{r,j}\|, \\
s.t. &  \ \ \sum_{j=1}^n w_{ij}=p_g(x_{g,i}), \forall i; \quad  \sum_{i=1}^n w_{ij}=p_{data}(x_{r,j}), \forall j. \\
\end{split}
\]
To calculate Wasserstein distance, we used the code provided at \url{https://github.com/xuqiantong/GAN-Metrics/}. 

To address the computational complexity of 1-WD, which is of $O(n^3)$, we partitioned the samples drawn at each run to 1000 groups, each group being of size 100, and calculated 1-WD for each group and then average the distance over the groups. 
The distance values from each run were further averaged over five independent runs and 
reported in Table 1 of the main text.

The MMD measures the dissimilarity between the two distributions $p_g$ and $p_{data}$ for some fixed kernel function $\kappa(\cdot,\cdot)$, and it is defined as 
\[
MMD^2(p_g,p_{data})=\mathbb{E}_{x_g,x_g'\sim p_g; x_r,x_r'\sim p_{data}} [ \kappa(x_g,x_g')-2 \kappa(x_g,x_r)+\kappa(x_r,x_r')].
\]
A lower MMD value means that $p_g$ is closer to $p_{data}$. 
In this paper, we calculated MMD values using 
the code provided at {https://www.onurtunali.com/ml/2019/03/08/maximum-mean-discrepancy-in-machine-learning.html} with the ``rbf" kernel option.  
We calculated the MMD values with the same sample grouping method as used in calculation of  1-WD.


\section{More Numerical Examples}

\subsection{A Gaussian Example: Additional Results}
\begin{figure}[htbp]
\begin{center}
\begin{tabular}{cc}
 (a) & (b) \\
\includegraphics[width=2.5in]{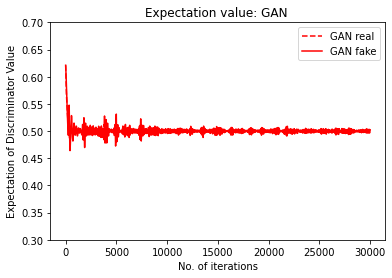} &
\includegraphics[width=2.5in]{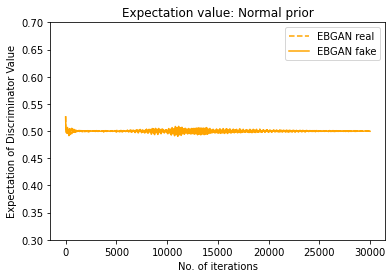} \\
\end{tabular}
\end{center}
\caption{\label{Gaussian_convergence} 
Empirical means of $D_{\theta_d^{(t)}}(x_i)$ and $D_{\theta_d^{(t)}}(\tilde{x}_i)$ 
produced by (a) GAN  and (b) EBGAN  with a Gaussian prior along with iterations.  }
\end{figure}

Figure \ref{Gaussian_convergence} shows the empirical means of $D_{\theta_d^{(t)}}(x)$ and $D_{\theta_d^{(t)}}(\tilde{x})$ produced by the two methods along with iterations, which indicates that both methods can reach the 0.5-0.5 convergence very fast.   
 
\subsection{A Mixture Gaussian Example: Additional Results}
  
\begin{figure}[htbp]
\begin{center}
\begin{tabular}{ccc}
 (a) & (b) & (c) \\
\includegraphics[width=2.0in]{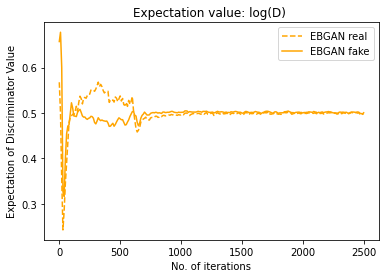} &
\includegraphics[width=2.0in]{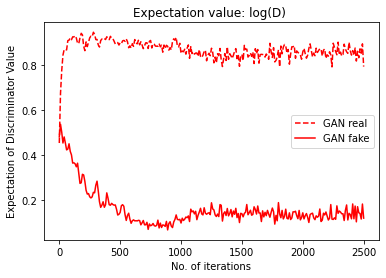} & 
\includegraphics[width=2.0in]{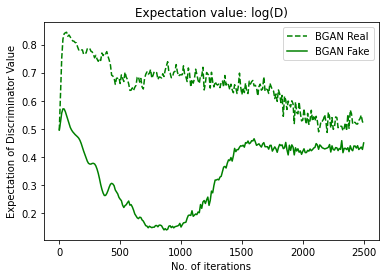} \\
(d) & (e) & (f) \\
\includegraphics[width=2.0in]{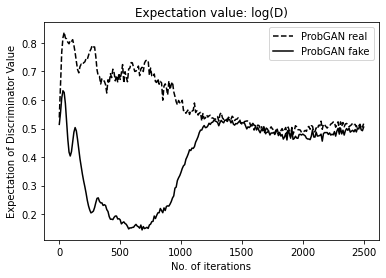} & 
\includegraphics[width=2.0in]{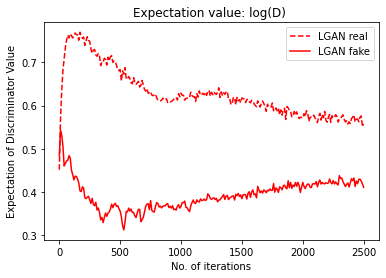} &
\includegraphics[width=2.0in]{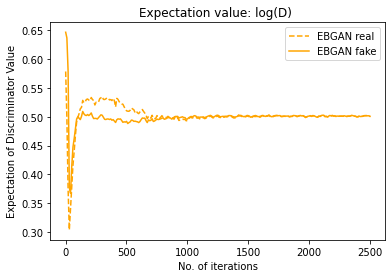}\\
\end{tabular}
\end{center}
\caption{\label{simFig3} Nash equilibrium convergence 
 plots with $\phi_3(D)=\log(D)$, which compare the empirical means of $D_{\theta_d^{(t)}}(x_i)$ and  $D_{\theta_d^{(t)}}(\tilde{x}_i)$ produced by 
  different methods along with iterations: (a) EBGAN with $\lambda=0$,
  (b) non-saturating GAN, (c) BGAN, (d) ProbGAN,
  (e) Lipschitz GAN, and (f) EBGAN with a Lipschitz penalty. }
\end{figure}

 For this example, we have tried the choice $\phi_3(D)=\log(D)$ of non-saturating GAN. 
 Under this non-saturating setting, the game is no longer of the minimax style. However, it helps to overcome the gradient vanishing issue suffered by the minimax GAN. 
Figure \ref{simFig3} shows the empirical means  $\mathbb{E}(D_{\theta_d^{(t)}}(x_i))$  and  $\mathbb{E}(D_{\theta_d^{(t)}}(\tilde{x}_i))$ produced by different methods along with iterations. 
The non-saturating GAN and Lipschitz GAN still 
failed to converge to the Nash equilibrium, 
but BGAN and ProbGAN nearly converged 
after about 2000 iterations. 
In contrast, EBGAN still worked very well: It can converge to the Nash equilibrium in either case, with or without a  Lipschitz penalty. 

\begin{figure}[htbp]
\begin{center}
\begin{tabular}{ccc}
(a) & (b) & (c) \\
\includegraphics[width=2.0in]{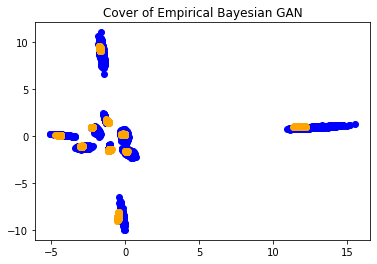} &
 \includegraphics[width=2.0in]{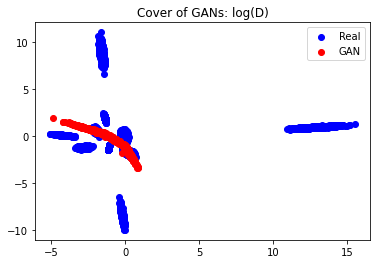} &
  \includegraphics[width=2.0in]{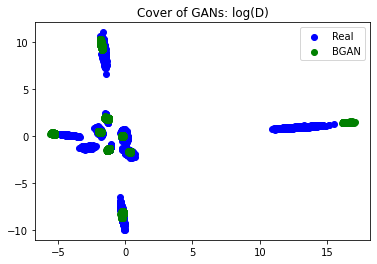}  \\
(e) & (e) & (f) \\ 
   \includegraphics[width=2.0in]{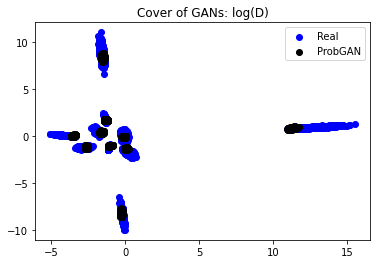}  & 
   \includegraphics[width=2.0in]{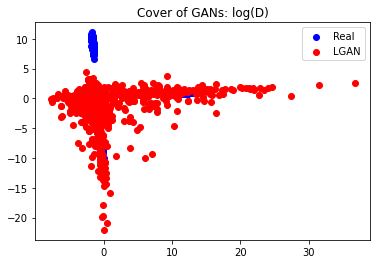} &
   \includegraphics[width=2.0in]{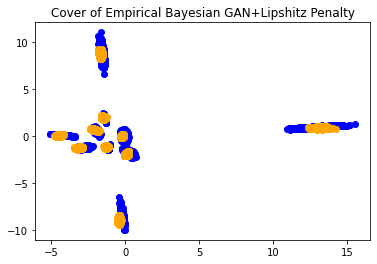}  \\
\end{tabular}
\end{center}
 \caption{\label{simFig4} Component recovery plots produced by different methods with $\phi_3(D)=\log(D)$:  (a) EBGAN with $\lambda=0$, (b) non-saturating GAN, (c) BGAN, (d) ProbGAN, (e) Lipschitz GAN, and (f) EBGAN with a Lipschitz penalty. }
\end{figure}

Figure \ref{simFig4} shows the plots of component recovery from the fake data. It indicates that EBGAN has recovered all 10 components of the real data in either case, with or without a Lipschitz penalty. 
Both ProbGAN and BGAN worked much better 
with this non-saturating choice than with the minimax choice of $\phi_3$:
The ProbGAN has even recovered all 10 components, 
although the coverage area is smaller than that by EBGAN; and BGAN just had one component missed in 
recovery. The non-saturating GAN and Lipschitz GAN still failed for this example, which is perhaps due to the 
the model collapse issue. Using a single generator is hard to generate data following a multi-modal distribution. 

\subsection{Image Generation: HMNIST} 

We compared GAN and EBGAN on another real data problem, the HAM10000 (“Human Against Machine with 10000 training images”) dataset, which is also known as HMNIST and available at { https://www.kaggle.com/kmader/skin-cancer-mnist-ham10000}. The dataset consists of a total of 10,015 dermatoscopic images of skin lesions classified to seven types of skin cancer. Unlike other benchmark computer vision datasets, HMNIST has imbalanced group sizes. The largest group size is 6705, while the smallest one is 115, which makes it hard for conventional GAN training algorithms.

\begin{figure}[htbp]
\begin{center}
\begin{tabular}{cc}
(a) & (b) \\
\includegraphics[width=0.45\textwidth]{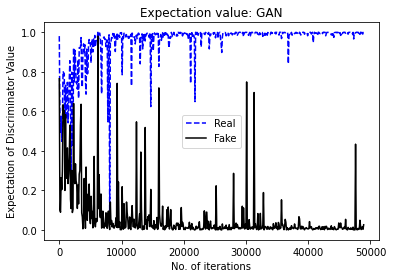}  &  \includegraphics[width=0.45\textwidth]{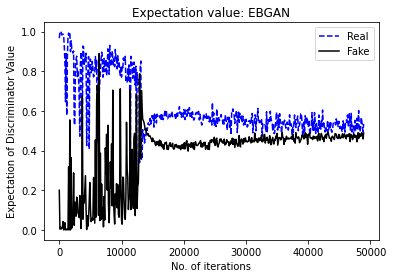} \\ 
        (c) & (d) \\
            \includegraphics[width=0.45\textwidth]{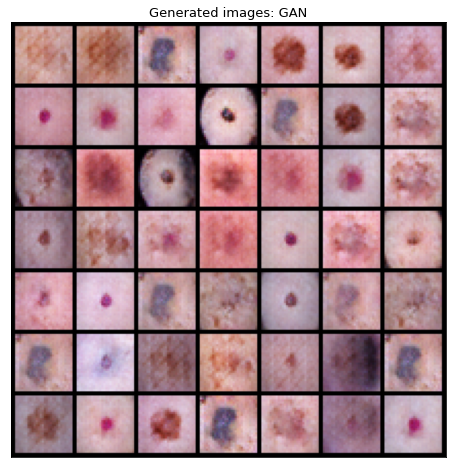}  & 
            \includegraphics[width=0.45\textwidth]{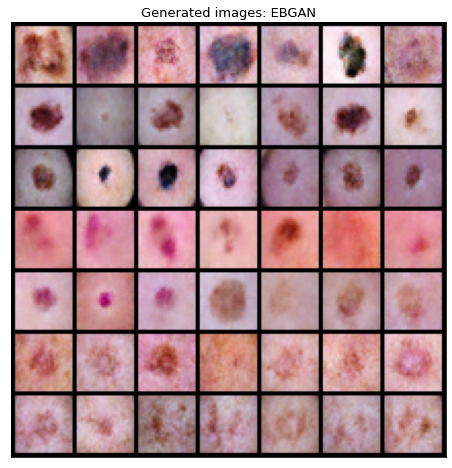} \\
            (e) & (f) \\
            \includegraphics[width=0.45\textwidth]{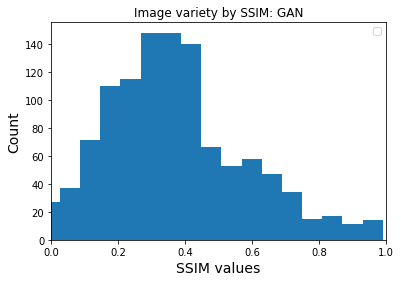}  & 
            \includegraphics[width=0.45\textwidth]{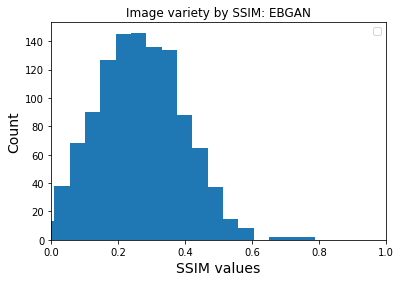} \\
\end{tabular}
\end{center}
\caption{Results for the HMNIST example: (a) convergence plot of GAN; (b) convergence plot of EBGAN;  (c) images generated by GAN; (d) images generated by EBGAN; 
(e) histograms of SSIMs for the images shown in plot (c) ; and (f) histograms of SSIMs for the images shown in plot (d). } 
\label{HMNISTfig2}
\end{figure}

Our results for the example are shown in Figure \ref{HMNISTfig2}, which indicates again that the EBGAN outperforms the GAN. In particular, the GAN is far from the 0.5-0.5 convergence, while EBGAN can achieve it. 
 In terms of images generated by the two methods, it is clear that GAN suffers from a mode collapse issue; many images generated by it have a similar pattern even, e.g., those 
 shown in the cells (1,4), (2,1), (4,1), (4,5), (5,2), (5,5) and (6,2) 
 share a similar pattern.
 In contrast, the images generated by EBGAN show a clear clustering structure; each row corresponds to one different pattern.

Since the clusters in the dataset are imbalanced, the IS score does not work well for measuring the quality of the generated images. To tackle this issue, we calculated the 
structural similarity index measure (SSIM) \cite{Wang2004ImageQA} for each pair of the images shown in Figure \ref{HMNISTfig2}(c) and those shown in Figure \ref{HMNISTfig2}(d), respectively. SSIM is a metric that measures the similarity between two images; it takes a value of 1 if two images are identical.  Figure \ref{HMNISTfig2}(e) \& (f) shows the histograms of SSIMs for the images shown in Figure \ref{HMNISTfig2}(c) \& (d), respectively. The comparison shows clearly that the images generated by EBGAN have a larger diversity than those by GAN.     


\subsection{Conditional Independence Tests} \label{CITsect}

Conditional independence is a fundamental concept in graphical modeling \citep{Lauritzen1996} and causal inference \citep{Pearl2009} for multivariate data. Conditional independence tests have long been studied in statistics, which are to test the hypotheses
\[
H_0:\  X \indep Y|Z \quad versus \quad H_1: \ X \centernot\indep Y |Z,
\]
where $X\in \mathbb{R}^{d_x}$, $Y\in \mathbb{R}^{d_y}$ and $Z\in \mathbb{R}^{d_z}$. For the case that the variables are discrete and the dimensions are low, the Pearson $\chi^2$-test and the likelihood ratio tests are often used. For the case that the variables are Gaussian and linearly dependent, one often conducts the test using the partial correlation coefficient or its equivalent measure, see e.g., \cite{PC1993} and \cite{LiangSQ2015}. However, in real-life situations, the normality and linear dependence assumptions are often not satisfied and thus nonparametric conditional independence tests are required. An abundance of such type of tests have been developed in the literature, e.g., permutation-based tests \citep{Doran2014,Berrett2019}, kernel-based tests  \citep{KernelCItest2012,Strobl2019}, classification or regression-based tests \citep{Sen2017,Zhang2017FeaturetoFeatureRF}, and knockoff tests \citep{knockoffs2018}.
Refer to \cite{ChunF2019} for an overview.

As pointed out in \cite{ChunF2019}, the existing nonparametric conditional independence tests often suffer from the curse of dimensionality in the confounding vector $Z$; that is, the tests may be ineffective when the sample size is small, since the accumulation of spurious correlations from a large number of variables in $Z$ makes it difficult to discriminate between the hypotheses.
As a remedy to this issue, \cite{bellot2019CItest} proposed a generative conditional independent test (GCIT) based on GAN. The method belongs to the class of nonparametric conditional independence tests and it consists of three steps:
(i) simulating samples  $\tilde{X}_1, \ldots, \tilde{X}_M \sim q_{H_0}(X)$ under the 
null hypothesis $H_0$ via GAN, 
where $q_{H_0}(X)$ denotes 
the distribution of $X$ under $H_0$;  
(ii) defining an appropriate test statistic  $\varrho(\cdot)$ which captures the $X$-$Y$ dependency in each of the samples $\{(\tilde{X}_1,Y,Z), (\tilde{X}_2,Y,Z), \ldots, (\tilde{X}_M,Y,Z)\}$;  
and (iii) calculating the $p$-value 
 \begin{equation}
   \widehat{p}=  \frac{\sum_{m=1}^M \bf{1}\left\{\varrho(\tilde{X}_m,Y,Z )>\varrho(X,Y,Z) \right\}}{M},
 \end{equation}
which can be made arbitrarily close to the true probability
\[
\mathbb{E}_{\tilde{X} \sim q_{H_0}(X)}
\bf{1}\left\{\varrho(\tilde{X},Y,Z)\geq  \varrho(X, Y, Z)\right\}
\]
by sampling a large number of samples $\tilde{X}$ 
from $q_{H_0}(X)$. \cite{bellot2019CItest} proved that this test is valid and showed empirically that it is robust with respect to the dimension of the confounding vector $Z$.  It is obvious that the power of the GCIT depends on how well the samples $\{\tilde{X}_1, \tilde{X}_2, \ldots, \tilde{X}_m\}$ approximate the distribution 
$q_{H_0}(X)$.

\subsubsection{Simulation Studies}  \label{GCITexam}

To show that EBGAN improves the testing power of GCIT, we consider a simulation example taken from \cite{bellot2019CItest} for testing the  hypotheses: 
\begin{equation*}
    \begin{split}
        H_0: & \quad X=f_1(A_x Z+\epsilon_x), \quad Y=f_2(A_y Z+\epsilon_y),\\
        H_1: & \quad X=f_1(A_x Z+\epsilon_x), \quad  Y= f_3(\alpha A_{xy}X+A_{y}Z+\epsilon_y), 
    \end{split}
\end{equation*}
where the matrix dimensions of $A_{(\cdot)}$ are such that $X$ and $Y$ are univariate. The entries of 
$A_{(\cdot)}$ as well as the parameter $\alpha$ are randomly drawn from  Unif$[0, 1]$, and the noise variables $\epsilon_{(\cdot)}$ are Gaussian with mean 0 and variance 0.025. Three specific cases are considered in the simulation: 
 
\begin{itemize}
\item Case 1. Multivariate Gaussian: $f_1$, $f_2$ and $f_3$ are identity functions,
$Z \sim \mathcal{N}(0, I_{d_z})$, which result in multivariate Gaussian data and linear dependence under $H_1$.
 
\item Case 2. Arbitrary relationship:
  $f_1$, $f_2$ and $f_3$ are randomly sampled from  $\{\tanh(x), \exp(-x)$, $x^2\}$, $Z \sim N (0, I_{d_z})$, which 
 results in more complex distributions and variable dependencies. It resembles the complexities we can expect in real applications.
 
 \item Case 3. Arbitrary relationships with a mixture $Z$ distribution:
 \begin{equation*}
    \begin{split}
        H_0: & \quad X=\{ f_{1,a}(A_x Z_a+\epsilon_x), f_{1,b}(A_x Z_b+\epsilon_x)\}, \quad
        Y= f_{2}(A_y Z+\epsilon_y), \\
        H_1:  & \quad X=\{ f_{1,a}(A_x Z_a+\epsilon_x), f_{1,b}(A_x Z_b+\epsilon_x)\}, \quad
        Y= f_3(\alpha A_{xy}X+A_{y}Z+\epsilon_y), 
    \end{split}
\end{equation*} 
where $Z_a\sim \mathcal{N}(1_d,I_d)$, $Z_b\sim \mathcal{N}(-1_d,I_d)$, $Z_a,Z_b\in \mathbb{R}^{\frac{n}{2}\times d}$, 
$Z=(Z_a^T,Z_b^T)^T$, $f_{1,a}$, $f_{1,b}$, $f_2$ and $f_3$ are randomly sampled from $\{\tanh(x),\exp(-x), x^2\}$ and $f_{1,a}\neq f_{1,b}$. 

\end{itemize} 
For each case, we simulated 100 datasets under 
$H_1$, where each dataset consisted of $150$ samples. Both GAN and EBGAN were 
applied to this example with the 
randomized dependence coefficient \citep{Lopez2013rdc} used as the test statistic $\varrho(\cdot)$. 
Here GAN was trained as in \cite{bellot2019CItest} with the code available at \url{https://github.com/alexisbellot/GCIT}. 
In GAN, the objective function of the generator was regularized by a mutual information which  
 encourages to generate samples $\tilde{X}$ as independent as possible from the observed variables $X$ and thus enhances the power of the test. 
 The EBGAN was trained in a plain manner without the mutual information term included in $\mathcal{J}(\theta_d;\theta_g)$.  
Detailed settings of the experiments were given in the supplement.  
In addition, two kernel-based methods, 
KCIT \citep{KernelCItest2012} and RCoT \citep{Strobl2019}, were applied to this example for comparison.  

 
\begin{figure}[htbp]
\begin{center}
\begin{tabular}{ccc}
(a) \\
\includegraphics[width=2.75in]{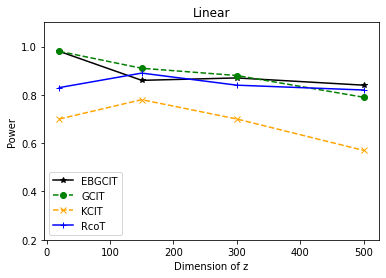} \\
(b) \\
 \includegraphics[width=2.75in]{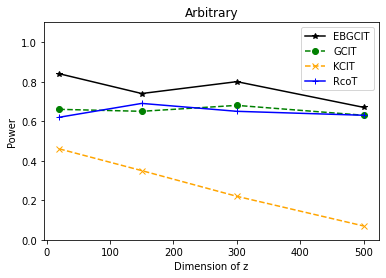} \\ 
(c) \\
 \includegraphics[width=2.75in]{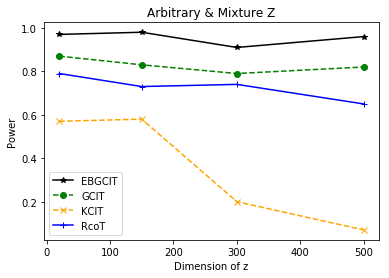} \\
\end{tabular}
\end{center}
 \caption{\label{FigCItest} Generative conditional independence tests with GAN (denoted by GCIT) and EBGAN (denoted by EBGCIT): 
 (a) Power curve for Case 1; (b) Power curve for Case 2 ; (c) Power curve for Case 3}
\end{figure}

Figure \ref{FigCItest} summarizes the results of the experiments. For case 1, EBGAN, GAN and RCoT are almost the same, and they all outperform KCIT. For case 2 and case 3, EBGAN outperforms the other three methods significantly.  
Note that, by \cite{bellot2019CItest}, GAN represents the state-of-the-art method for high-dimensional nonparametric conditional independence tests. For similar examples, 
\cite{bellot2019CItest} showed that GAN significantly outperformed the existing statistical tests, including the kernel-based tests \citep{KernelCItest2012, Strobl2019},
knockoff-based test \citep{knockoffs2018},
and classification-based test \citep{Sen2017}.

\subsubsection{Identifications of Drug Sensitive Mutations}

As a real data example, we applied EBGAN to identification of genetic mutations that affect response of cancer cells to an anti-cancer drug. This helps cancer clinics, as the  treatment for a cancer patient can strongly depend on the mutation of his/her genome \citep{Garnett2012} in precision medicine.  We used a sub-dataset of 
Cancer Cell Line Encyclopedia (CCLE), which relates the drug response of PLX4720 with 466 genetic mutations. The dataset consists of 474 cell lines.
Detailed settings of the experiment were given in the supplement.

Table \ref{gen_mut} shows the mutations identified by EBGAN at a significance level of 0.05, where the dependency of the drug response on the first 12 mutations has been validated by the existing literature at PubMed.  Since 
PLX4720 was designed as a BRAF inhibitor, the low p-values of BRAF.MC, BRAF.V600E and BRAF confirm the validity of the proposed test. EBGAN also identified MYC as a drug sensitive mutation, but which was not detected via GAN in \cite{bellot2019CItest}.
Our finding is validated by 
\cite{Singleton2017}, which reported that 
BRAF mutant cell lines with intrinsic resistance to BRAF rapidly upregulate MYC upon treatment of PLX4720. CRKL is another mutation identified by EBGAN but not by GAN, and this finding can be validated by the experimental results reported in \cite{Tripathi2020}.


\begin{table}[htbp]
  \caption{Genetic experiment results: Each cell gives the p-value indicating the dependency between a mutation and drug response, where the superscript $^-$ indicates that the dependency of drug response on the mutation has not yet been validated in the literature.}
  \label{gen_mut}
  \vspace{2mm}
  \centering
    \begin{adjustbox}{width=1.0\textwidth}
  \begin{tabular}{cccccccc} \toprule
 BRAF.MC & IRAK1 & BRAF.V600E & BRAF & HIP1  & SRPK3 & MAP2K4 & FGR  \\  
 0.001 & 0.002 &0.003 &0.003 &0.004 &0.012& 0.014& 0.014  \\ \midrule
  PRKD1 & CRKL & MPL  & MYC &  MTCP1$^-$ & ADCK2$^-$ & RAD51L1$^-$  &  \\  
  0.015& 0.016& 0.027& 0.037 & 0.011 & 0.037 & 0.044  & \\
  \bottomrule
    \end{tabular}
    \end{adjustbox}
\end{table}

\subsection{Nonparametric Clustering} \label{clustersect2}
 
 This section gives details for different datasets we tried. 
 
\subsubsection{Two-Circle Problem}
 The most notorious example for classical  clustering methods is the two-circle problem. The dataset is generated as follows: 
\begin{equation} \label{2circle}
    \begin{split}
    Z_i=(z_{1i},z_{2i}&),  \quad \rm{where} \quad z_{1i}, z_{2i}\overset{\textit{iid}}{\sim} Unif[-1,1], \quad i=1,\dots,1000; \\
        \mbox{Inner Circle}&: 0.25*\left(\frac{z_{1i}}{\sqrt{z_{1i}^2+z_{2i}^2 }},\frac{z_{2i}}{\sqrt{z_{1i}^2+z_{2i}^2}}\right)+\epsilon, \quad i=1,\dots,500;\\
        \mbox{Outer Circle}&:  \left(\frac{z_{1i}}{\sqrt{z_{1i}^2+z_{2i}^2 }},\frac{z_{2i}}{\sqrt{z_{1i}^2+z_{2i}^2}}\right)+\epsilon,  \quad i=501,\dots,1000, 
    \end{split}
\end{equation}
where $\epsilon\sim \mathcal{N}(0,0.05^2I_2)$. 
 For this example, the K-means and agglomerative clustering methods are known to fail to detect the inner circle unless the data are appropriately transformed; DBSCAN is able to detect the inner circle, but it is hard to apply to other  high-dimensional problems due to its density estimation-based nature. 
  
\begin{figure}[htbp]
    \centering
    \includegraphics[width=0.8\textwidth]{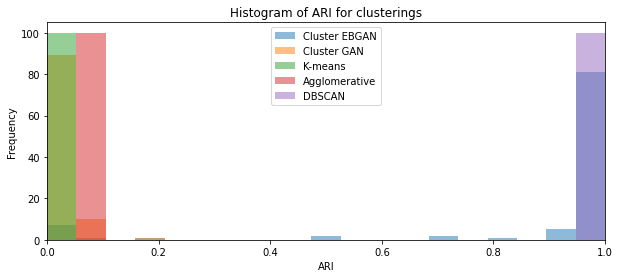}
    \caption{Histogram of ARI produced by different methods: Cluster EBGAN, Cluster GAN, K-means, Agglomerative, and DBSCAN. }
    \label{fig:2-circle-ARI}
    \vspace{-0.1in}
\end{figure}
\begin{figure}[htbp]
\centering
\begin{tabular}{cccc}
(a) & (b) & (c) & (d) \\
 \includegraphics[width=1.0in,height=1.0in]{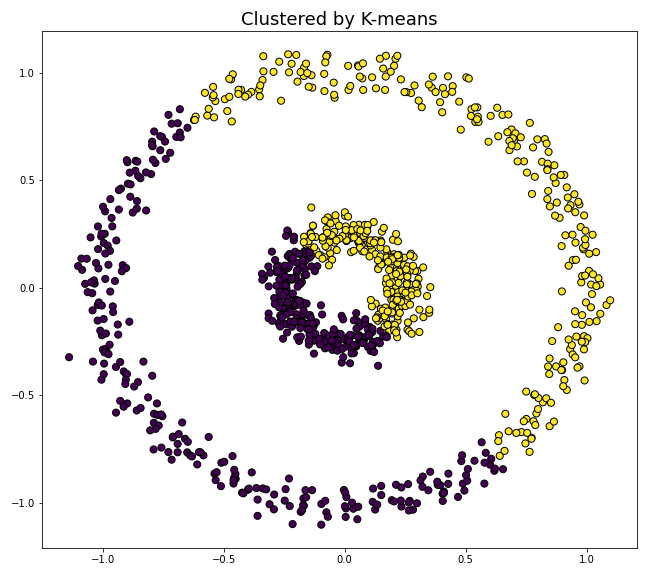} &
 \includegraphics[width=1.0in,height=1.0in]{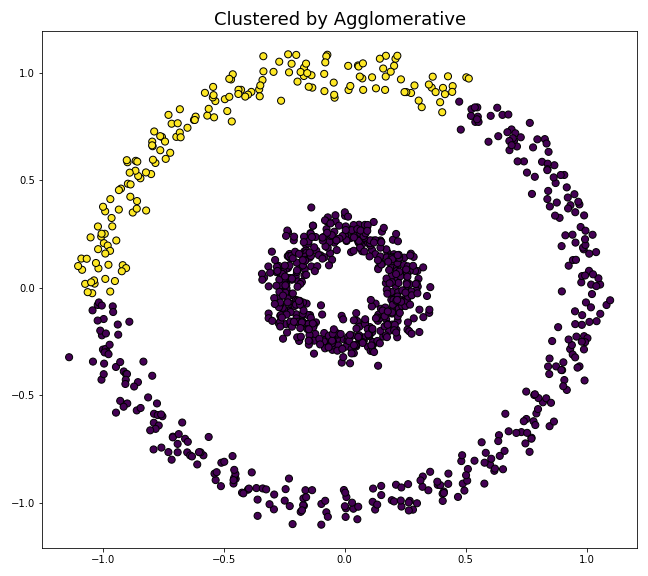} &
 \includegraphics[width=1.0in,height=1.0in]{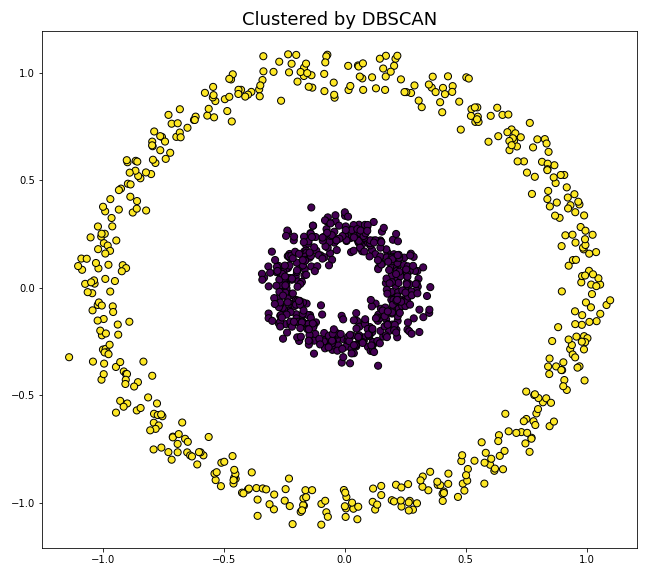} &
 \includegraphics[width=1.0in,height=1.0in]{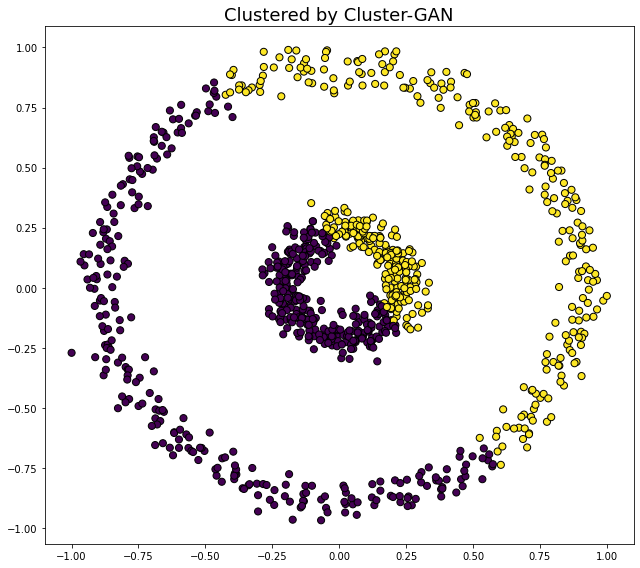} 
\\
 (e) & (f) & (g) & (h) \\
 \includegraphics[width=1.0in,height=1.0in]{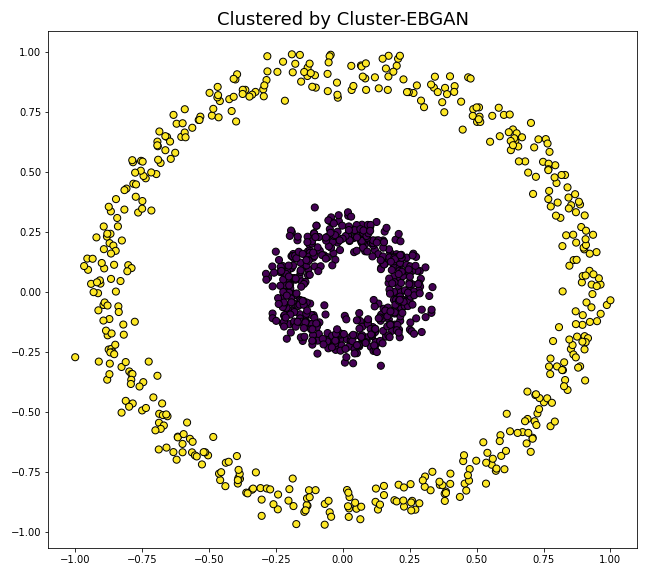} &
 \includegraphics[width=1.0in,height=1.0in]{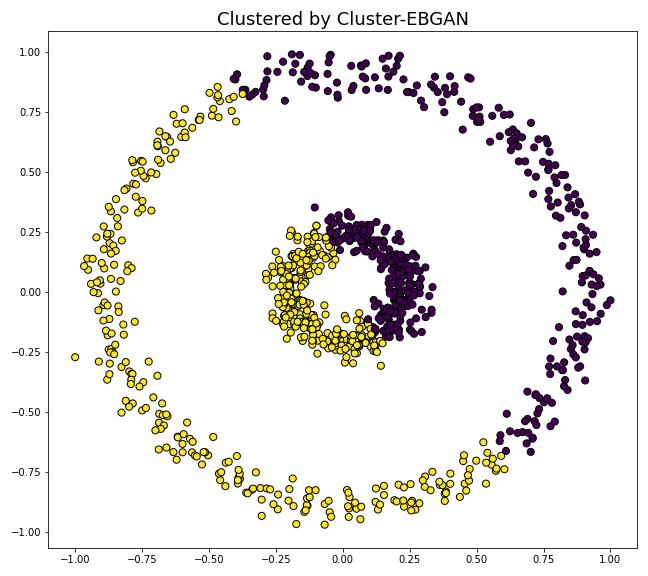} &
 \includegraphics[width=1.0in,height=1.0in]{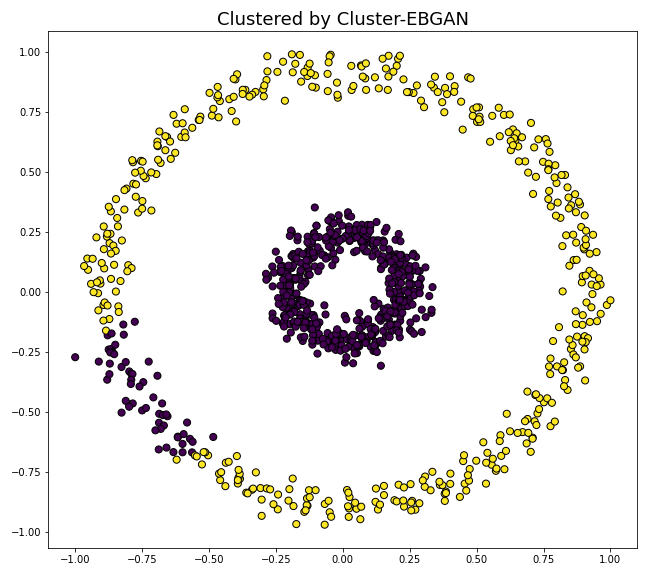} &
 \includegraphics[width=1.0in,height=1.0in]{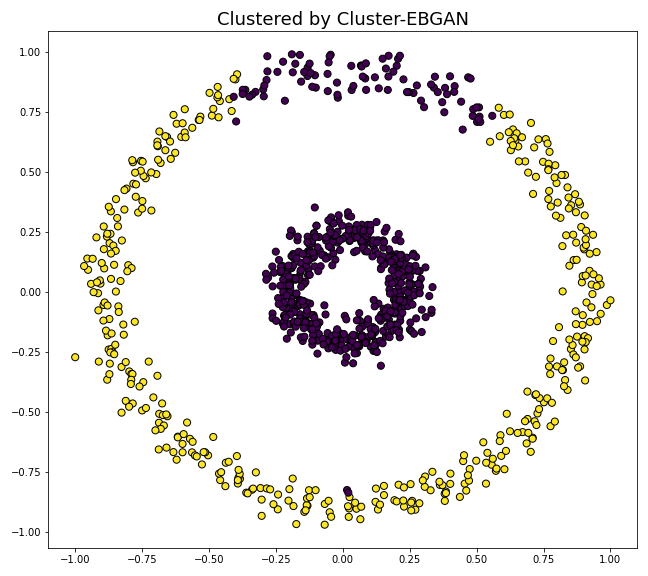}\\
 
\end{tabular} \caption{\label{clustergangen2circle}  (a) K-means clustering, (b) Agglomerative clustering, (c)  DBSCAN, (d) Cluster-GAN, (e)-(h) Cluster-EBGAN in different runs. }
 \end{figure}

For a simulated dataset, each of the methods, including K-means, agglomerative, DBSCAN, Cluster GAN and Cluster EBGAN, was run for 100 times with different initializations. 
Figure \ref{fig:2-circle-ARI} shows the histogram of the adjust Rand index (ARI) \citep{ARI1971} values obtained in those runs. It indicates that  K-means, agglomerative and DBSCAN produced the same clustering results in different runs, while Cluster GAN and Cluster EBGAN produced different ones in different runs. In particular, the ARI values resulted from Cluster GAN are around 0, whereas those from Cluster EBGAN are around 1.0. Figure \ref{clustergangen2circle} shows some clustering results produced by these methods.   

Figure \ref{fig:2-circle-ARI} and Figure \ref{clustergangen2circle} indicate that DBSCAN can constantly detect the inner circle; Cluster EBGAN can detect the inner circle in nearly 80\% of the runs; while K-means, agglomerative, and Cluster GAN failed to detect the inner circle. The comparison with Cluster GAN indicates that Cluster EBGAN has made a significant improvement in GAN training.

\paragraph{Iris} This is a classical clustering example.
 It contains the data for 50 flowers from each of three species - Setosa, Versicolor and Virginica. The dataset is available at 
\url{https://archive.ics.uci.edu/ml/datasets/iris}, 
which gives the measurements of the variables sepal length and width and petal length and width for each of the flowers.  
Table \ref{purity-table} summarizes the performance of different methods on the dataset. Other than ARI, 
the cluster purity is also calculated as a measure of the quality of clusters.  
Suppose that the data consists of $K$ clusters and each cluster consists of $n_k$ observations denoted by $\{x_k^j\}_{j=1}^{n_k}$. If the data were grouped into $M$ clusters, then the cluster purity is defined by 
 \begin{equation*}
\frac{\sum_{k=1}^{K}\max\{ \sum_{j=1}^{n_k}1(\mathcal{E}_2(x_k^j)=l):  l=1,2,\ldots,M\}}{n_1+\dots+n_K},
 \end{equation*} 
 which measures the percentage of the samples being correctly clustered. 
 Both the measures, ARI and cluster purity, 
 have been used in \cite{Mukher19clustergan} for assessing the performance of Cluster GAN. 
 The comparison shows that Cluster EBGAN significantly outperforms Cluster GAN and classical clustering methods in both ARI and cluster purity. 

\paragraph{Seeds} The dataset is available at 
\url{https://archive.ics.uci.edu/ml/datasets/seeds}. 
The examined group comprised kernels belonging to three different varieties of wheat: Kama, Rosa and Canadian, 70 elements each, randomly selected for
the experiment. Seven geometric parameters of
wheat kernels were measured, including area, perimeter, compactness, length, width, asymmetry coefficient,  and length of kernel groove. Table \ref{purity-table} summarizes the performance of different methods on the dataset. The comparison indicates that Cluster EBGAN significantly outperforms others in both ARI and cluster purity.  For this dataset, DBSCAN is not available any more, as performing density estimation in a 7-dimensional space is hard.  


\paragraph{MNIST}

\begin{figure}[htbp]
\centering
\begin{tabular}{cc}
(a) & (b)  \\
\includegraphics[width=1.25in,height=1.25in]{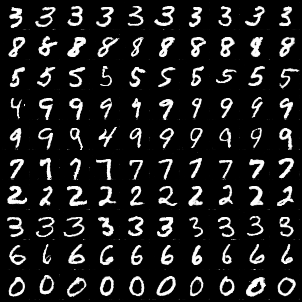} &
 \includegraphics[width=1.25in]{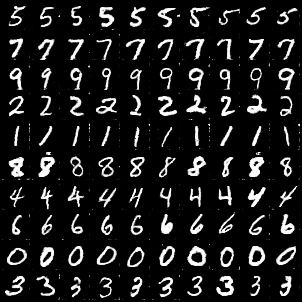}
\end{tabular} \caption{\label{clustergangen} 
Images generated by (a) Cluster GAN  and (b) Cluster EBGAN, where each row corresponds to a different $z_c$ index vector.}
 \end{figure}

The MNIST dataset consists of 70,000 images of digits ranging from 0 to 9. Each sample point is a $28\times 28$ grey scale image. Figure \ref{clustergangen} compares the images generated by Cluster GAN and Cluster EBGAN, each representing the 
 best result achieved by the corresponding method
 in 5 independent runs. It is remarkable that Cluster EBGAN can generate all digits from 0 to 9 and there is no confusion of digits between different generators.  However, Cluster GAN failed to generate the digit 1 and confused  the digits 4 and 9. 
 Table \ref{purity-table} summarizes the performance of different methods, which  indicates again the superiority of the Cluster EBGAN over Cluster GAN and classical nonparametric methods. 

\section{Experimental Settings}

In all training with the Adam algorithm \citep{adam}, 
we set the tuning parameters $(\alpha_1,\alpha_2)=(0.5,0.999)$.
In this paper, all the deep convolutional GANs (DCGANs) were trained using the Adam algorithm.

For Algorithm 1 (of the main text), the step size is chosen 
in the form $w_t=c_1(t+c_2)^{-\zeta_1}$, and the momentum smoothing factor $\alpha$ is re-denoted by $\alpha_1$ in Tables \ref{partab1}-\ref{partab_fashionmnist}. A constant learning rate $\epsilon$ and a constant momentum biasing factor $\rho=1$.


\subsection{A Gaussian Example}

Table \ref{partabKL} gives the parameter settings of GAN and EBGAN for the Gaussian example. 

\begin{table}[htbp]
\caption{\label{partabKL} Parameter settings for the 2D Gaussian dataset with $\phi_3(D)=log(D)$}
\begin{center}
\begin{tabular}{lccccc}
\toprule
Method & \multicolumn{2}{c}{Learning rate} & $\alpha_1$ & $\alpha_2$ & $\rho$ \\ \cline{2-3}
& Discriminator($\omega_t$) & Generator($\epsilon_t$)  \\
\midrule
\textcolor{black}{GAN}  & 0.00002 &  0.0002 & 0.5& 0.999 &   \\
EBGAN & $(c_1,c_2,\zeta_1)=(1,1000,0.75)$ & 0.01 & 0.9 & & 1 \\
\bottomrule
\end{tabular}
\end{center}
\end{table}

\subsection{A Mixture Gaussian Example}  

Tables \ref{partab1} and \ref{partab2} give the parameter settings of different methods for the minimax and non-saturating cases, respectively. For EBGAN, we set $\tau=0.01$.

\begin{table}[htbp]
\caption{\label{partab1} Parameter settings for the synthetic dataset: the minimax case with $\phi_3(D)=-\log(1-D)$}
\begin{center}
\begin{adjustbox}{width=\textwidth}
\begin{tabular}{lcccccc}
\toprule
Method & \multicolumn{2}{c}{Learning rate} & $\alpha_1$ & $\alpha_2$ & $\rho$ & $\lambda(\text{Lipshitz})$ \\ \cline{2-3}
& Discriminator & Generator  \\
\midrule
\textcolor{black}{GAN}  & 0.0002 &  0.0002 & 0.5& 0.999 & &  \\
BGAN & 0.001 & 0.001 & 0.9 &\\
Probgan &  0.0005& 0.0005 & 0.5 & \\
EBGAN &  $(c_1,c_2,\zeta_1)=(1,1000,0.75)$ & 0.5 & 0.9 & &1 & \\
Lipshitz-GAN &   0.0002 &  0.0002 & 0.5& 0.999 & & \textcolor{black}{5} \\
Lipshitz-EBGAN &  $(c_1,c_2,\zeta_1)=(1,1000,0.75)$ & 0.5 & 0.9 & &1 &\textcolor{black}{5}\\

\bottomrule
\end{tabular}
\end{adjustbox}
\end{center}
\end{table}

\begin{table}[htbp]
\caption{ \label{partab2} Parameter settings for the synthetic dataset: the non-saturating case with  $\phi_3(D)=\log(D)$}
\begin{center}
\begin{adjustbox}{width=\textwidth}
\begin{tabular}{lcccccc}
\toprule
Method & \multicolumn{2}{c}{Learning rate} & $\alpha_1$ & $\alpha_2$ & $\rho$  & $\lambda(\text{Lipshitz})$  \\ \cline{2-3}
& Discriminator & Generator  \\
\midrule
\textcolor{black}{GAN} & 0.0002 &  0.0002 & 0.5& 0.999 & &  \\
BGAN & 0.001 & 0.001 & 0.9 &\\
Probgan &  0.0005& 0.0005 & 0.9 & \\
EBGAN &  $(c_1,c_2,\zeta_1)=(1,1000,0.75)$ & 0.5  & 0.9 & &1 &\\
Lipshitz-GAN & 0.0002 &  0.0002 & 0.5& 0.999 && 5 \\
Lipshitz-EBGAN &  $(c_1,c_2,\zeta_1)=(1,1000,0.75)$ & 0.5  & 0.9 & &1 &5\\
\bottomrule
\end{tabular}
\end{adjustbox}
\end{center}
\end{table}

\subsection{Fashion MNIST}

The network structures of all models are typical DCGAN style. We set the mini-batch size to 300, set the total number of epochs to 200, and set the dimension of $z_n$ to 10. For training the inception model, we used Adam with a learning rate of $0.0003$, $(\alpha_1,\alpha_2)=(0.9,0.999)$, a mini-batch size of 50, and $5$ epochs. After training, the prediction accuracy on the test data set was $0.9304$. Table \ref{partab_fashionmnist} gives the parameter settings used by different methods. And we set $\zeta_2=\frac{1}{40}$ for EBGAN referring to \cite{kim2020stochastic}. In addition, we set $\tau=0.001$ for EBGAN, and set $k_g=10$ for EBGAN, BGAN and ProbGAN. 

\begin{table}[htbp]
\caption{ \label{partab_fashionmnist} Parameter settings for the Fashion MNIST}
\begin{center}
\begin{adjustbox}{width=\textwidth}
\begin{tabular}{lccccc}
\toprule
Method & \multicolumn{2}{c}{Learning rate} & $\alpha_1$ & $\alpha_2$ & $\rho$    \\ \cline{2-3}
& Discriminator & Generator  \\
\midrule
\textcolor{black}{GAN} & 0.0002 &  0.0002 & 0.5& 0.999 &  \\
BGAN & 0.005 & 0.005& 0.9\\
ProbGAN &  0.005& 0.005 & 0.9\\
EBGAN-KL &  $(c_1,c_2,\zeta_1)=(0.5,250,1)$ with Adam & 0.01  & 0.9 & &1 \\
EBGAN-Gaussian &  $(c_1,c_2,\zeta_1)=(1,500,1)$ with Adam & 0.01  & 0.9 & &1 \\

\bottomrule
\end{tabular}
\end{adjustbox}
\end{center}
\end{table}

\begin{table}[htbp]
  \caption{Model structure of EBGAN for Fashion MNIST}
  \centering
  \begin{tabular}{cc} \toprule

 Generator  & Discriminator \\ \midrule
     4$\times$ 4 conv, 512 stride 2 ReLU &  4$\times$ 4 conv, 64 stride 2 pad 1 LReLU \\ \midrule
     3$\times$ 3 conv, 256 stride 2 pad 1 ReLU  &  4$\times$ 4 conv, 128 stride 2 pad 1 LReLU \\ \midrule
   4$\times$ 4 conv, 128 stride 2 pad 1 ReLU & 3$\times$ 3 conv, 256 stride 2 pad 1 LReLU  \\ \midrule
   $4\times 4$ upconv 64 stride 2 pad 1 Tanh  & 4$\times$ 4 conv, 512 stride 2 LReLU  \\
  \bottomrule
    \end{tabular}
\end{table}

\subsection{HMNIST} 

We set the latent dimension as 20, and use the normal prior $N(0,\frac{1}{80})$ on 7 generator parameters with temperature $\tau=0.001$, with 200 batch size. Other parameter settings and the model structure are given in Table \ref{HMNIST:par} and 
Table \ref{HMNIST:structure}, respectively. 

\begin{table}[htbp]
\caption{ \label{partab_Hmnist} Parameter settings for the HMNIST}
\label{HMNIST:par}
\begin{center}
\begin{adjustbox}{width=\textwidth}
\begin{tabular}{lccccc}
\toprule
Method & \multicolumn{2}{c}{Learning rate} & $\alpha_1$ & $\alpha_2$ & $\rho$    \\ \cline{2-3}
& Discriminator & Generator  \\
\midrule
\textcolor{black}{GAN} & 0.0002 &  0.0002 & 0.5& 0.999 &  \\
EBGAN-Gaussian &  $(c_1,c_2,\zeta_1)=(0.05,250,1)$ with Adam & 0.001  & 0.9 & & 1 \\
\bottomrule
\end{tabular}
\end{adjustbox}
\end{center}
\end{table}

\begin{table}[htbp]
  \caption{Model structure of EBGAN for HMNIST}
  \label{HMNIST:structure}
  \centering
  \begin{tabular}{cc} \toprule

 Generator  & Discriminator \\ \midrule
     4$\times$ 4 conv, 512 stride 2 ReLU &  4$\times$ 4 conv, 64 stride 2 pad 1 LReLU \\ \midrule
     3$\times$ 3 conv, 256 stride 2 pad 1 ReLU  &  4$\times$ 4 conv, 128 stride 2 pad 1 LReLU \\ \midrule
   4$\times$ 4 conv, 128 stride 2 pad 1 ReLU & 3$\times$ 3 conv, 256 stride 2 pad 1 LReLU  \\ \midrule
   $4\times 4$ upconv 64 stride 2 pad 1 Tanh  & 4$\times$ 4 conv, 512 stride 2 LReLU  \\
  \bottomrule
    \end{tabular}
\end{table}

\subsection{Conditional independence test}

\paragraph*{Simulated Data} 

The network structures of all models we used are the same as in
 \cite{bellot2019CItest}. In short, the generator network has a structure of $(d+d/10)-(d/10)-1$ and the discriminator network has a structure of $(1+d)-(d/10)-1$, where $d$ is the dimension of the confounding vector $Z$. 
 All experiments for GCIT were implemented with the code given at \url{https://github.com/alexisbellot/GCIT/blob/master/GCIT.py}. For the functions $\phi_1$, $\phi_2$ and $\phi_3$, the nonsaturating settings were adopted, i.e., 
 we set  $(\phi_1,\phi_2,\phi_3)=(\log x, \log(1-x),\log x)$.
 For both cases of the synthetic data, EBGAN was run with
 a mini-batch size of 64, Adam optimization was used with learning rate 0.0001 for discriminator. A prior $p_g=N(0,100I_p)$, $p$ for dimension of parameters, and a constant learning rate of 0.005 were used for the generator. Lastly, we set $\tau=1$. Each run consisted of 1000 iterations for case 1, case 2 and case 3. KCIT and RcoT were run by R-package at \url{https://github.com/ericstrobl/RCIT}.


\paragraph*{CCLE Data} 
For the CCLE dataset, EBGAN was run for 1000 iterations and $(c_1,c_2,\eta_1,\alpha_1)=(1,1000,0.75,0.9)$ was used. Other parameters were set as above.

\subsection{Nonparametric Clustering}

\paragraph*{Two Circle} We set the dimension of 
$z_n$ to 3, set $(\beta_n,\beta_c)=(0.1,0.1)$,  set the mini-batch size to 500, and set a constant learning rate of 0.05 with $\tau=1$ for the generator. For optimization of the discriminator, we used Adam and set $(\alpha_1,\alpha_2)=(0.5,0.9)$ with a constant learning rate of 0.1. The total number of epochs was set to 2000.

\begin{table}[htbp]
 \caption{Model structure of Cluster-GAN and Cluster-EBGAN for two-circle data}
 \label{2-circle-structure}
 \centering
 \begin{tabular}{ccc} \toprule

Generator & Encoder & Discriminator \\ \midrule
     FC 20 LReLU &FC 20 LReLU &FC 30 LReLU  \\ \midrule
  FC 20 LReLU  & FC 20 LReLU & FC 30 LReLU \\ \midrule
 FC 2 linear Tanh & FC 5 linear & FC 1 linear\\
 \bottomrule
    \end{tabular}
\end{table}

\paragraph*{Iris} For the iris data, we used a simple feed-forward network structure for Cluster GAN and Cluster EBGAN. We set the dimension of $z_n$ to 20,  set $(\beta_n,\beta_c)=(10,10)$, set the mini-batch size to 32, and set a constant learning rate of 0.01 for the generator with $\tau=1$. For optimization of the discriminator, we used Adam and set $(\alpha_1,\alpha_2)=(0.5,0.9)$ with a learning rate of 0.0001. The hyperparameters of Cluster-GAN is set to the default values.  

\begin{table}[htbp]
 \caption{Model structure of Cluster-GAN and Cluster-EBGAN for Iris}
 \label{Seeds-structure}
 \centering
 \begin{tabular}{ccc} \toprule
Generator & Encoder & Discriminator \\ \midrule
     FC 5 LReLU &FC 5 LReLU &FC 5 LReLU  \\ \midrule
  FC 5 LReLU  & FC 5 LReLU & FC 5 LReLU \\ \midrule
 FC 4 linear Sigmoid & FC 23 linear & FC 1 linear\\
 \bottomrule
    \end{tabular}
\end{table}

\paragraph*{Seeds} For the seeds data, we used a  simple feed-forward network structure for Cluster-GAN and Cluster-EBGAN. We set the dimension of $z_n$ to 20, set $(\beta_n,\beta_c)=(5,5)$,  set the mini-batch size to 128, and set a constant learning rate of 0.01 for generator with $\tau=0.0001$. For optimization of the discriminator, we used Adam and set $(\alpha_1,\alpha_2)=(0.5,0.9)$ with a learning rate of  0.005. The hyperparameters of Cluster-GAN is set to the default values.  

\begin{table}[htbp]
 \caption{Model structure of Cluster-GAN and Cluster-EBGAN for Seeds}
 \label{iris-structure}
 \centering
 \begin{tabular}{ccc} \toprule
Generator & Encoder & Discriminator \\ \midrule
     FC 20 LReLU &FC 20 LReLU &FC 100 LReLU  \\ \midrule
  FC 20 LReLU  & FC 20 LReLU & FC 100 LReLU \\ \midrule
 FC 7 linear Tanh & FC 23 linear & FC 1 linear\\
 \bottomrule
    \end{tabular}
\end{table}

\paragraph*{MNIST}

For Cluster GAN, our implementation is based on the code given at \url{https://github.com/zhampel/clusterGAN}, with a a small modification on Encoder. 
The Structures of the generator, encoder and discriminator are given as follow. Cluster GAN was run with the same parameter setting as given in the original work \cite{Mukher19clustergan}. 

\begin{table}[htbp]
  \caption{Model structure of ClusterGAN for MNIST data}
  \centering
  \begin{adjustbox}{width=\textwidth}
  \begin{tabular}{ccc} \toprule
 Generator & Encoder & Discriminator \\ \midrule
    FC 1024 ReLU BN & 4$\times$ 4 conv, 64 stride 2 LReLU &  4$\times$ 4 conv, 64 stride 2 LReLU \\ \midrule
     FC $7\times 7 \times 128$ ReLU BN & 4$\times$ 4 conv, 128 stride 2 LReLU &  4$\times$ 4 conv, 64 stride 2 LReLU \\ \midrule
   $4\times 4$ upconv 64 stride 2 ReLU BN & 4$\times$ 4 conv, 256 stride 2 LReLU& FC1024 LReLU  \\ \midrule
   $4\times 4$ upconv 1 stride 2 Sigmoid  & FC 1024 LReLU & FC 1 linear  \\ \midrule
        & FC 40 &  \\
  \bottomrule
    \end{tabular}
\end{adjustbox}
\end{table}

For Cluster EBGAN, to accelerate  computation, we used the parameter sharing strategy as in \cite{hoang2018mgan}, where all generators share the parameters except for the first layer. We set the dimension of $z_n$ to 5,    $(c_1,c_2,\eta_1,\alpha_1)=(40,10000,0.75,0.9)$, set the  mini-batch size to 100, and set a constant learning rate of $0.005$ for the generator.  For the functions $\phi_1$, $\phi_2$ and $\phi_3$, the non-saturating settings were adopted, i.e., 
 we set  $(\phi_1,\phi_2,\phi_3)=(\log x, \log(1-x),\log x)$.


\begin{table}[htbp]
  \caption{Model structure of ClusterEBGAN for MNIST simulation}
  \label{10modelst}
  \centering
  \begin{adjustbox}{width=\textwidth}
  \begin{tabular}{ccc} \toprule

 Generator & Encoder & Discriminator \\ \midrule
     4$\times$ 4 conv, 512 stride 2 ReLU & 4$\times$ 4 conv, 64 stride 2 LReLU &  4$\times$ 4 conv, 64 stride 2 LReLU \\ \midrule
     3$\times$ 3 conv, 128 stride 2 pad 1 ReLU & 4$\times$ 4 conv, 128 stride 2 LReLU &  4$\times$ 4 conv, 64 stride 2 LReLU \\ \midrule
   4$\times$ 4 conv, 64 stride 2 pad 1 ReLU & 4$\times$ 4 conv, 256 stride 2 LReLU& FC1024 LReLU  \\ \midrule
   $4\times 4$ upconv 1 stride 2 pad 1 Sigmoid  & FC 1024 LReLU & FC 1 linear  \\ \midrule
        & FC 30 &  \\

  \bottomrule
    \end{tabular}
\end{adjustbox}
\end{table}

\bibliographystyle{chicago}
\bibliography{reference}

\end{document}


\title{Supplementary Material for ``A New Paradigm of Generative Adversarial Networks based on Randomized Decision Rules''}

\author{Sehwan Kim, Qifan Song, and Faming Liang
\thanks{To whom correspondence should be addressed: Faming Liang.
  F. Liang is Distinguished Professor (email: fmliang@purdue.edu),
  S. Kim is Graduate Student, and Q. Song is Associate Professor, Department of Statistics,
  Purdue University, West Lafayette, IN 47907.
 }
 }

\maketitle
 
 \setcounter{table}{0}
\renewcommand{\thetable}{S\arabic{table}}
\setcounter{figure}{0}
\renewcommand{\thefigure}{S\arabic{figure}}
\setcounter{equation}{0}
\renewcommand{\theequation}{S\arabic{equation}}
\setcounter{algorithm}{0}
\renewcommand{\thealgorithm}{S\arabic{algorithm}}
\setcounter{lemma}{0}
\renewcommand{\thelemma}{S\arabic{lemma}}
\setcounter{theorem}{0}
\renewcommand{\thetheorem}{S\arabic{theorem}}
\setcounter{remark}{0}
\renewcommand{\theremark}{S\arabic{remark}}

\section{More Numerical studies} 
 
 \subsection{Synthetic Example}
 
\begin{figure}[htbp]
\begin{center}
\begin{tabular}{cc|c}
 (a) & (b) & (c)\\
\includegraphics[width=2in]{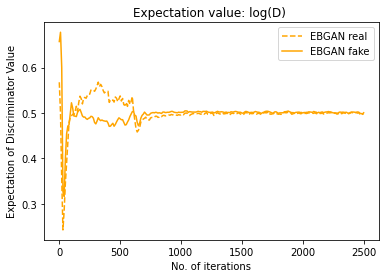} &
\includegraphics[width=2in]{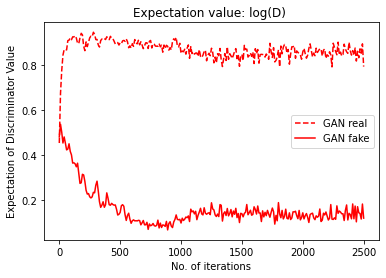} &
\includegraphics[width=2in]{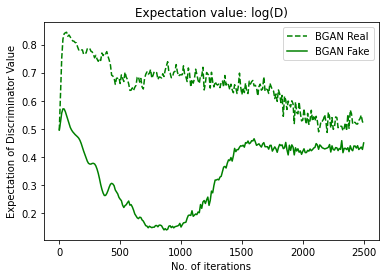} \\

(d) & (e) & (f)\\
\includegraphics[width=2in]{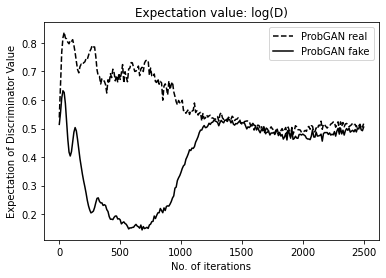} &
\includegraphics[width=2in]{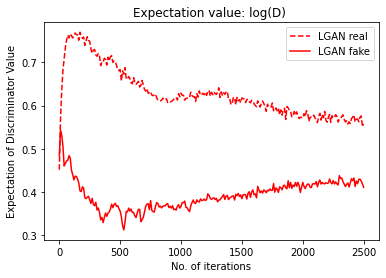} &
\includegraphics[width=2in]{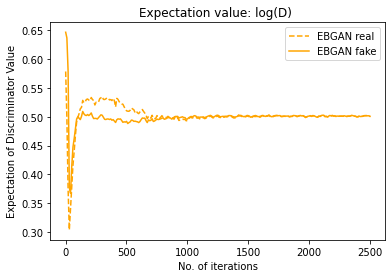}\\
\end{tabular}
\end{center}
\caption{\label{simFig3} Nash equilibrium convergence 
 plots with $\phi_3(D)=\log(D)$, which compare the empirical means of $D_{\theta_d^{(t)}}(x_i)$ and  
 $D_{\theta_d^{(t)}}(\tilde{x}_i)$ produced by 
  different methods along with iterations: (a) EBGAN with $\lambda=0$,
  (b) non-saturating GAN, (c) BGAN, (d) ProbGAN,
  (e) Lipschitz GAN, and (f) EBGAN with a Lipschitz penalty. }
\end{figure}

 We have tried the choice $\phi_3(D)=\log(D)$ of non-saturating GAN. 
 Under this non-saturating setting, the game is no longer of the minimax style. However, it helps to overcome the gradient vanishing issue suffered by the minimax GAN. 
Figure \ref{simFig3} shows the empirical means  $\mathbb{E}(D_{\theta_d^{(t)}}(x_i))$  and  $\mathbb{E}(D_{\theta_d^{(t)}}(\tilde{x}_i))$ produced by different methods along with iterations. 
The non-saturating GAN and Lipschitz GAN still 
failed to converge to the Nash equilibrium, 
but BGAN and ProbGAN nearly converged 
after about 2000 iterations. 
In contrast, EBGAN still worked very well: It can converge to the Nash equilibrium in either case, with or without a  Lipschitz penalty. It is interesting to point out that under this non-saturating setting, the Lipschitz penalty seems not helpful to the convergence of EBGAN any more.

\begin{figure}[htbp]
\begin{center}
\begin{tabular}{cc|c}
(a) & (b) &  (c) \\
\includegraphics[width=2in]{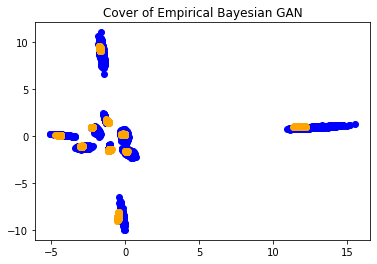} &
 \includegraphics[width=2in]{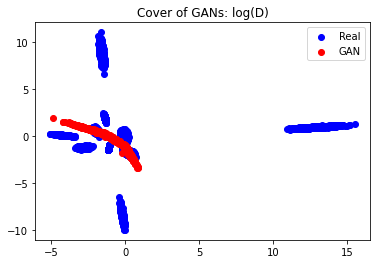} &  
  \includegraphics[width=2in]{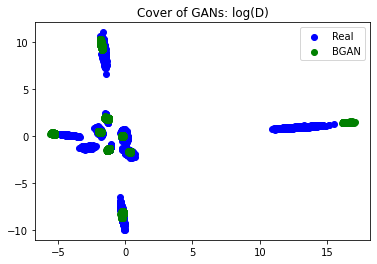} \\
 (d) & (e) & (f) \\
   \includegraphics[width=2in]{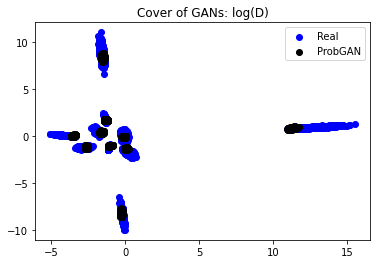} & 
   \includegraphics[width=2in]{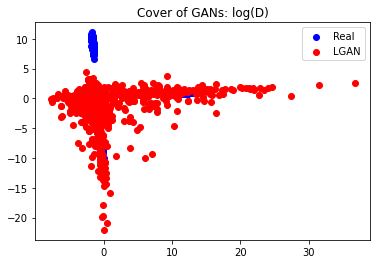} &
   \includegraphics[width=2in]{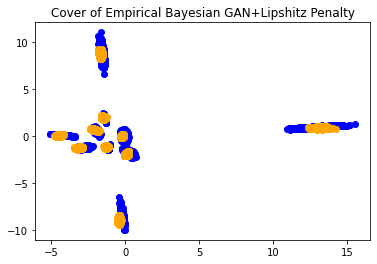}  \\
\end{tabular}
\end{center}
 \caption{\label{simFig4} Component recovery plots produced by different methods with $\phi_3(D)=\log(D)$:  (a) EBGAN with $\lambda=0$, (b) non-saturating GAN, (c) BGAN, (d) ProbGAN, (e) Lipschitz GAN, and (f) EBGAN with a Lipschitz penalty. }
\end{figure}

Figure \ref{simFig4} shows the plots of component recovery from the fake data. It indicates that EBGAN has recovered all 10 components of the true data in either case, with or without a Lipschitz penalty. 
Both ProbGAN and BGAN worked much better 
with this non-saturating choice than with the minimax choice of $\phi_3$:
The ProbGAN has even recovered all 10 components, 
although the coverage area is smaller than that by EBGAN; and BGAN just had one component missed in 
recovery. The non-saturating GAN and Lipschitz GAN still failed for this example, which is perhaps due to the 
the model collapse issue. Using a single generator is hard to generate data following a multi-modal distribution. 

\subsection{Image Generation with GAN}

This section illustrates the classical use of GAN for image generation using the skin cancer dataset (HMNIST). 
The dataset is available at \url{ https://www.kaggle.com/kmader/skin-cancer-mnist-ham10000}, 
which consists of a total of 10,015 dermatoscopic images of skin lesions classified to seven types of skin cancer. Unlike other benchmark computer vision datasets,
HMNIST has imbalanced group sizes. The largest 
group size is 6705, while the smallest one is 
115, which makes conventional neural network models hard to be learned for the dataset.  
A natural idea to tackle this difficulty is to do 
data augmentation via GANs. If we can train a GAN to generate data for the underrepresented groups, then the 
augmented dataset will help people for future 
cancer diagnosis. To achieve this goal, we tried the 
conditional GAN \citep{Mirza2014conditional} for this example. The conditional GAN has included the data class information as part of its input and thus can be used to to generate images for specific classes.  


The EBGAN, DCGAN, BGAN, and ProbGAN were applied to this example. Section \ref{hmnistsetup} of this material presents the structure information of the 
discriminator and generator as well as the training parameter settings of these methods. 
To evaluate the performance 
of these methods, we used the prediction accuracy which  
 evaluates whether the generated images with a class label are similar to the true images of that class. 

To calculate prediction accuracy, a CNN has been trained for the data with the structure information and training parameter settings
given in Section \ref{hmnistsetup} of this material. In prediction accuracy calculation, 7000 images 
were generated from each class  
using the conditional GANs learned 
at the last iteration. 
Table \ref{HMNISTtab} summarizes the results produced by each method in 5 independent runs. For the real images, we also made each class consist of 7000 images by sampling with replacement from the original dataset. The comparison indicates that EBGAN is significantly better than the 
existing methods in terms of prediction accuracy. 
 
\begin{table}[htbp]
  \caption{Prediction accuracy by different methods for HMNIST data, where the means and standard deviations were calculated based on 5 independent runs, and $p$-values were calculated for comparisons of EBGAN vs other methods with a  two-sided Gaussian test.}
  \label{HMNISTtab}
  \centering
  \begin{adjustbox}{width=1.0\textwidth}
  \begin{tabular}{cccccc} \toprule
  Method  & Real Images & DCGAN & BGAN & ProbGAN & EBGAN \\
\midrule
 Accuracy  & $72.248\pm 0.161$  & $71.081\pm0.109$ & 
    $70.106\pm 0.462$ & $70.176\pm 0.308$ & \textbf{74.398}$\pm0.376$ \\  
   $p$-value & 1.47e-7 & 0 & 5.79e-13 & 0 & --- \\
  \bottomrule
    \end{tabular}
\end{adjustbox}
\end{table}

\begin{figure}[htbp]
\begin{center}
    \begin{tabular}{cc}
    (a) & (b) \\
        \includegraphics[width=2.5in]{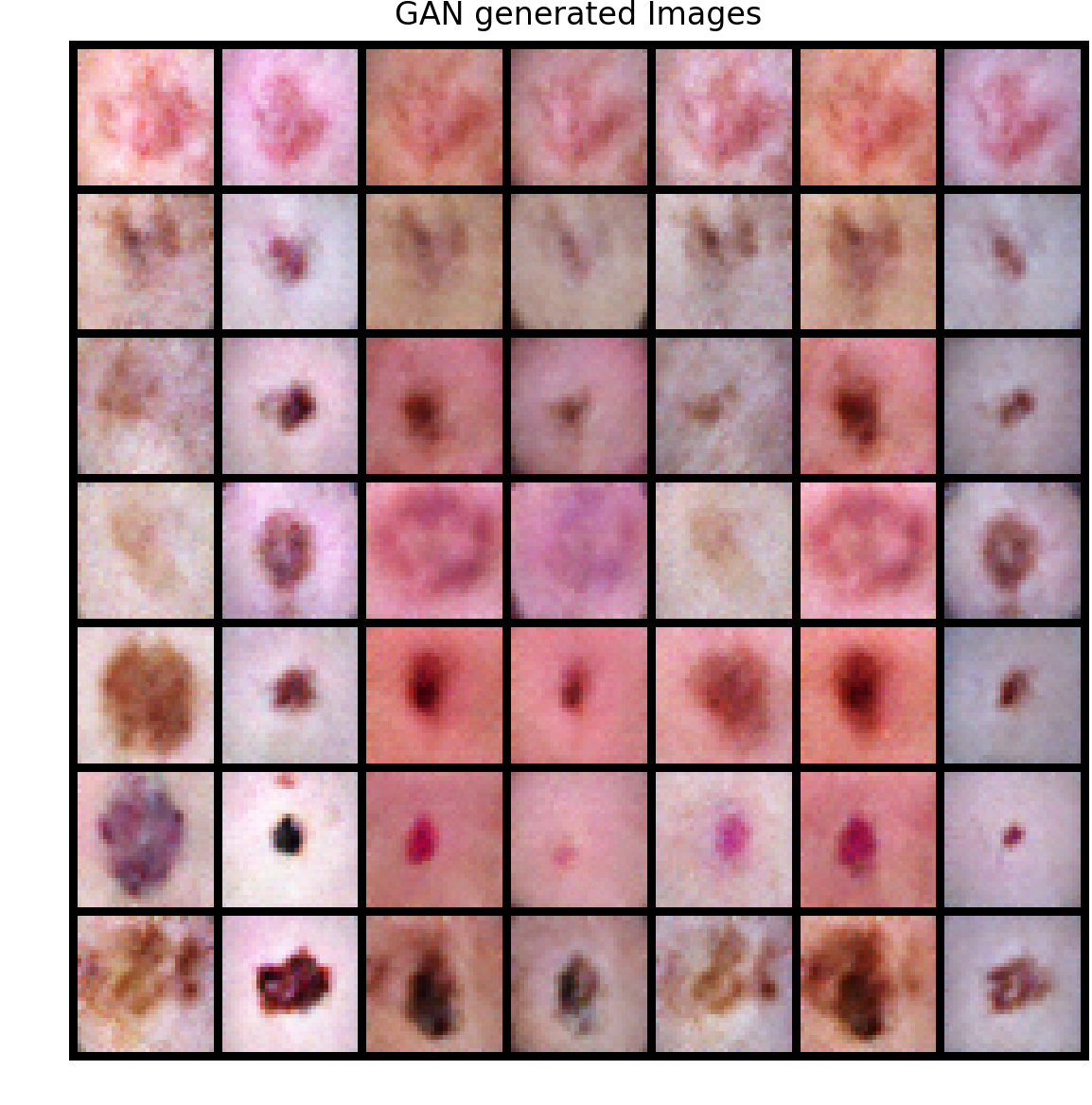} & 
        \includegraphics[width=2.5in]{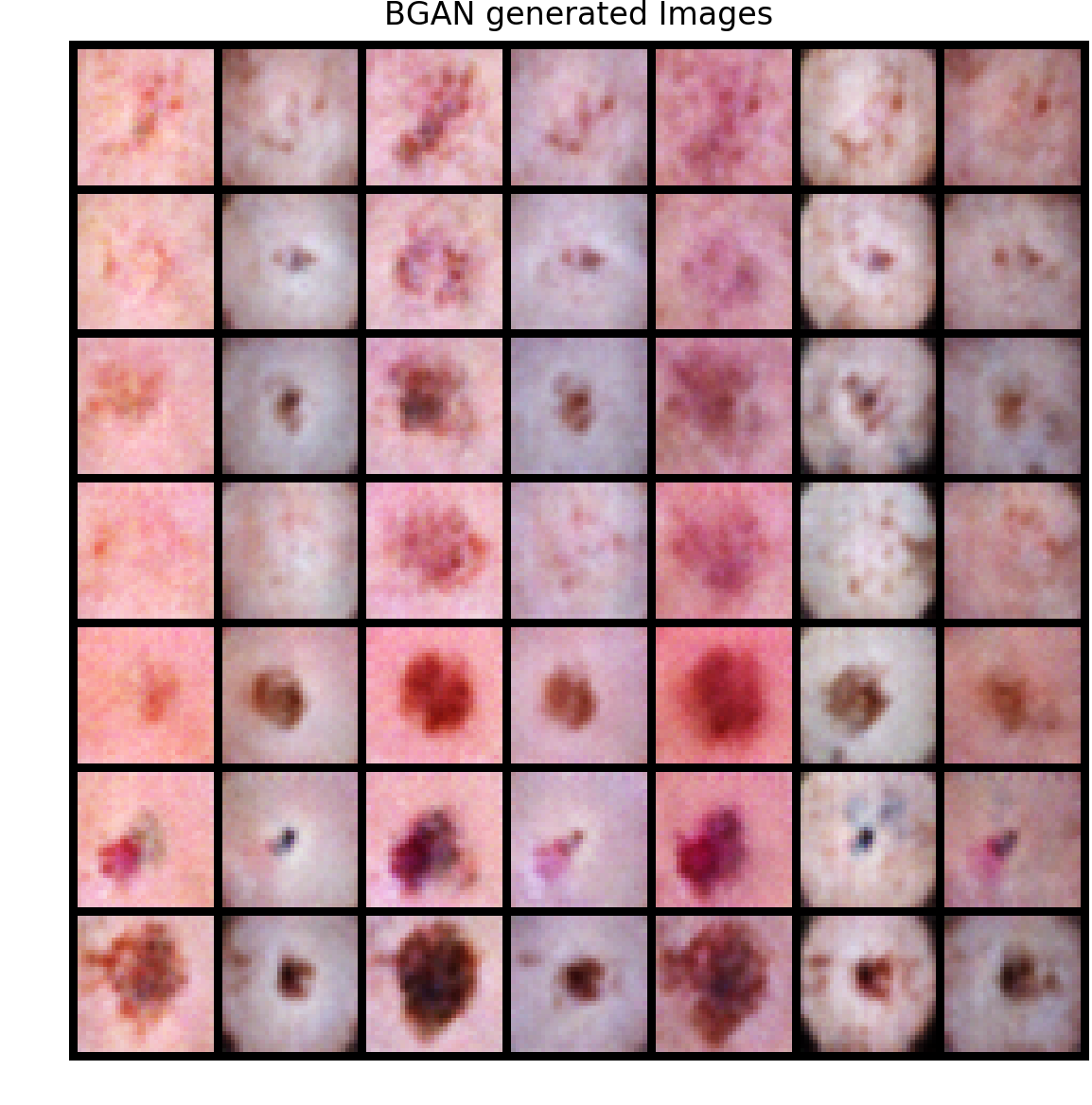} \\
    (c) & (d) \\
        \includegraphics[width=2.5in]{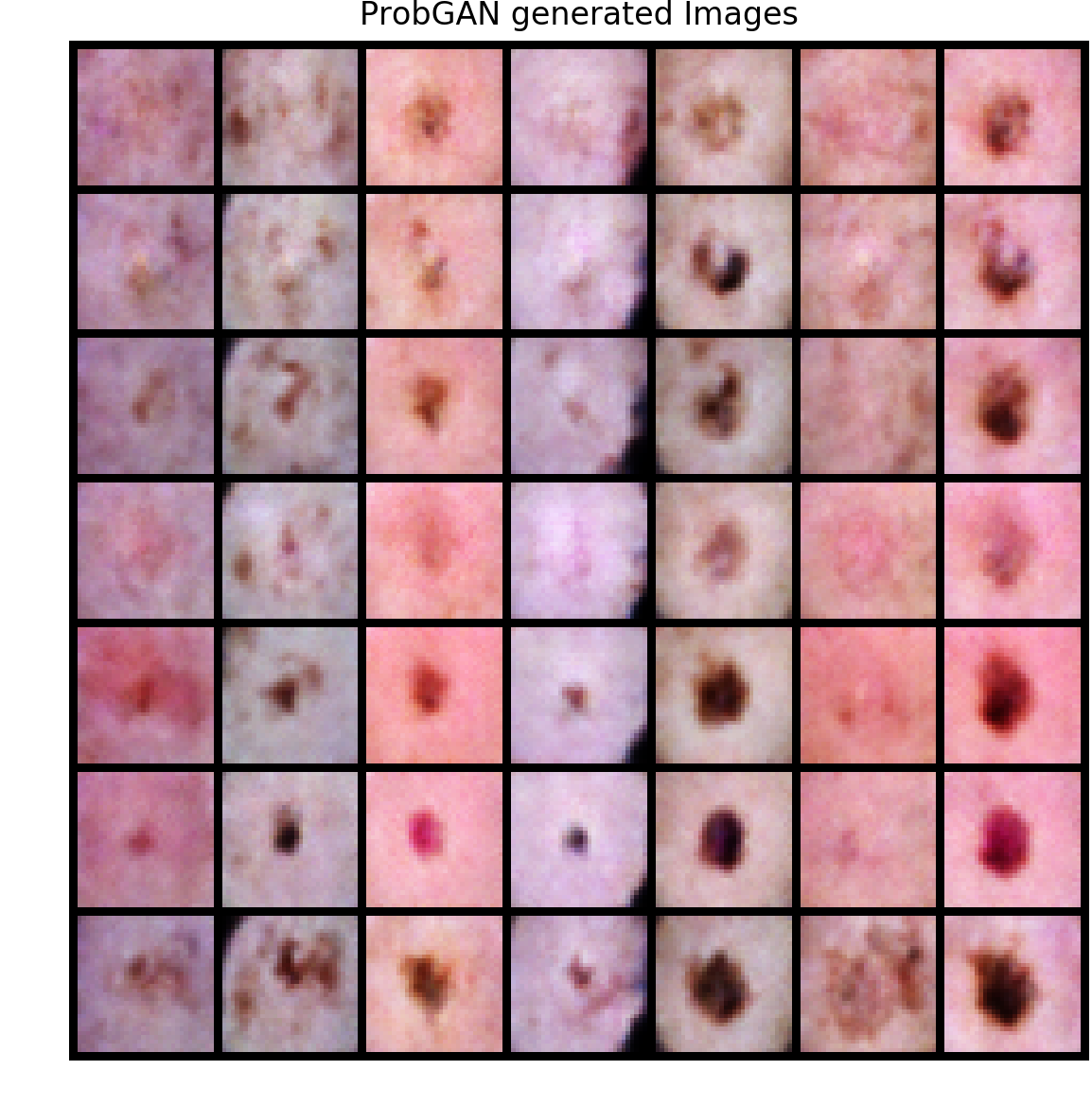} & 
        \includegraphics[width=2.5in]{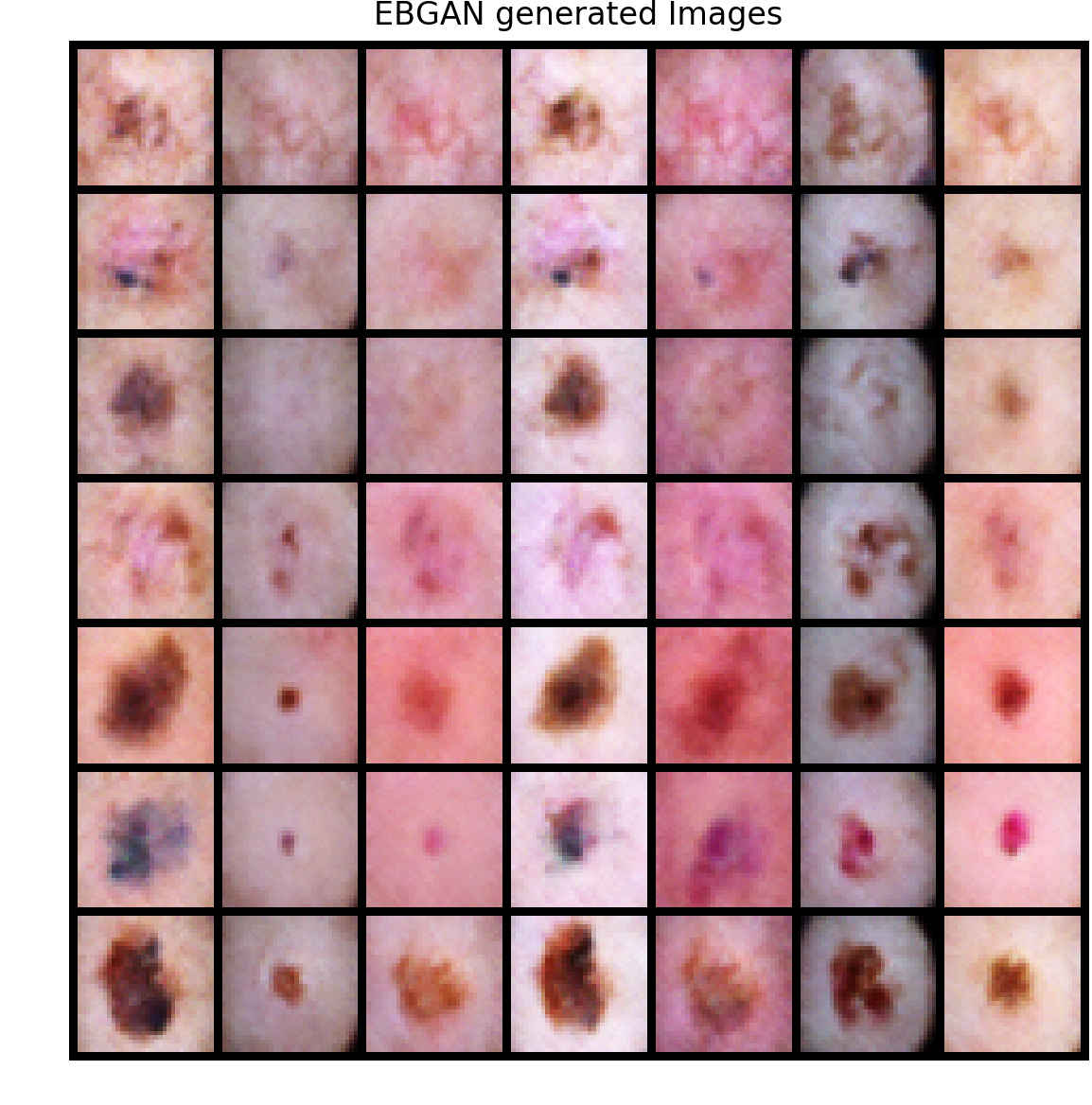} \\
 \end{tabular}
 \end{center}
    \caption{Images generated by the methods:
    (a) DCGAN, (b) BGAN,  (c) ProbGAN,  (d) EBGAN }\label{HMNISTfig}
\end{figure}

Figure \ref{HMNISTfig} shows images generated by 
different methods, where each row corresponds to 
 a different class. Figure \ref{HMNISTfig}(a) 
 indicates that DCGAN failed to generate images with right class labels. For example, row 1 and row 2 of Figure \ref{HMNISTfig}(a) show some non-similar shapes 
 to training data. 
 Both BGAN and ProbGAN worked less satisfactorily. For example, the images generated by 
 BGAN in rows 1-3 of Figure \ref{HMNISTfig}(b) are highly similar, and the images generated by ProbGAN in rows 1-3 of Figure \ref{HMNISTfig}(c) are also very similar.  EBGAN worked better than the other methods: It can generate different
 images according to class labels. 
 As shown by Figure \ref{HMNISTfig}(d), it is hard to find two highly similar rows there.

\section{Experimental Details}

In all training with the Adam algorithm\citep{adam}, 
we set the tuning parameters $(\alpha_1,\alpha_2)=(0.5,0.999)$.
In this paper, all the DCGANs were trained using the Adam algorithm.

For Algorithm 1 (of the main text), the step size is chosen 
in the form $w_t=c_1(t+c_2)^{-\zeta_1}$, and the momentum smoothing factor $\alpha$ is re-denoted by $\alpha_1$ 
in Tables \ref{partab1}-\ref{partab6}. 


\subsection{Synthetic Dataset}  

Tables \ref{partab1} and \ref{partab2} give the parameter settings of different methods for the minimax and non-saturating cases, respectively, with $\tau=0.01$ for EBGAN methods.

\begin{table}[htbp]
\caption{\label{partab1} Parameter settings for the synthetic dataset: the minimax setting with $\phi_3(D)=-\log(1-D)$}
\begin{center}
\begin{tabular}{lcccccc}
\toprule
Method & \multicolumn{2}{c}{Learning rate} & $\alpha_1$ & $\alpha_2$ & $\rho$ & $\lambda(\text{Lipshitz})$ \\ \cline{2-3}
& Discriminator & Generator  \\
\midrule
\textcolor{black}{GAN}  & 0.0002 &  0.0002 & 0.5& 0.999 & &  \\
BGAN & 0.001 & 0.001 & 0.9 &\\
Probgan &  0.0005& 0.0005 & 0.5 & \\
EBGAN &  $(c_1,c_2,\zeta_1)=(1,1000,0.75)$ & 0.5 & 0.9 & &1 & \\
Lipshitz-GAN &   0.0002 &  0.0002 & 0.5& 0.999 & & \textcolor{black}{5} \\
Lipshitz-EBGAN &  $(c_1,c_2,\zeta_1)=(1,1000,0.75)$ & 0.5 & 0.9 & &1 &\textcolor{black}{5}\\

\bottomrule
\end{tabular}
\end{center}
\end{table}

\begin{table}[htbp]
\caption{ \label{partab2} Parameter settings for the synthetic dataset: the non-saturating setting with  $\phi_3(D)=\log(D)$}
\begin{center}
\begin{tabular}{lcccccc}
\toprule
Method & \multicolumn{2}{c}{Learning rate} & $\alpha_1$ & $\alpha_2$ & $\rho$  & $\lambda(\text{Lipshitz})$  \\ \cline{2-3}
& Discriminator & Generator  \\
\midrule
\textcolor{black}{GAN} & 0.0002 &  0.0002 & 0.5& 0.999 & &  \\
BGAN & 0.001 & 0.001 & 0.9 &\\
Probgan &  0.0005& 0.0005 & 0.9 & \\
EBGAN &  $(c_1,c_2,\zeta_1)=(1,1000,0.75)$ & 0.5  & 0.9 & &1 &\\
Lipshitz-GAN & 0.0002 &  0.0002 & 0.5& 0.999 && 5 \\
Lipshitz-EBGAN &  $(c_1,c_2,\zeta_1)=(1,1000,0.75)$ & 0.5  & 0.9 & &1 &5\\
\bottomrule
\end{tabular}
\end{center}
\end{table}

\subsection{Fashion MNIST}

The network structures of all models are typical DCGAN style. We set the mini-batch size to 300 with total epochs 200, and set the dimension of $z_n$ to 10. For training inception model, we used Adam with learning rate $0.0003$, $(\alpha_1,\alpha_2)=(0.9,0.999)$. And mini-batch size was 50, with $5$ epochs. After training, accuracy on test data set was $0.9304$

\begin{table}[htbp]
\caption{ \label{partab_fashionmnist} Parameter settings for the Fashion MNIST}
\begin{center}
\begin{tabular}{lccccc}
\toprule
Method & \multicolumn{2}{c}{Learning rate} & $\alpha_1$ & $\alpha_2$ & $\rho$    \\ \cline{2-3}
& Discriminator & Generator  \\
\midrule
\textcolor{black}{GAN} & 0.0002 &  0.0002 & 0.5& 0.999 &  \\
BGAN & 0.005 & 0.005& 0.9\\
Probgan &  0.005& 0.005 & 0.9\\
EBGAN &  $(c_1,c_2,\zeta_1)=(1,500,0.1)$ with Adam & 0.01  & 0.9 & &1 \\
\bottomrule
\end{tabular}
\end{center}
\end{table}

\subsection{Conditional independence test}

\paragraph{Simulated Data} 

The network structures of all models we used are the same as in
 \cite{bellot2019CItest}. In short, the generator network has a structure of $(d+d/10)-(d/10)-1$ and the discriminator network has a structure of $(1+d)-(d/10)-1$, where $d$ is the dimension of the confounding vector $Z$. 
 All experiments for GCIT were implemented with the code given at \url{https://github.com/alexisbellot/GCIT/blob/master/GCIT.py}. For the functions $\phi_1$, $\phi_2$ and $\phi_3$, the nonsaturating settings were adopted, i.e., 
 we set  $(\phi_1,\phi_2,\phi_3)=(\log x, \log(1-x),\log x)$.
 For both cases of the synthetic data, EBGAN was run with
 a mini-batch size of 64, Adam optimization was used with learning rate 0.0001 for discriminator. A prior $p_g=N(0,100)$, and a constant learning rate of 0.005 were used for the generator. Each run consisted of 1000 iterations for case 1, case 2 and case 3. KCIT and RcoT were run by R-package at \url{https://github.com/ericstrobl/RCIT}.


\paragraph{CCLE Data} 
For the CCLE dataset, EBGAN was run for 1000 iterations and $(c_1,c_2,\eta_1,\alpha_1)=(1,1000,0.75,0.9)$ was used. Other parameters were set as above.

\subsection{Clustering}

\paragraph{Two Circle} We set the dimension of 
$z_n$ to 3, set $(\beta_n,\beta_c)=(0.1,0.1)$,  set the mini-batch size to 500, and set a constant learning rate of 0.05 for the generator. For optimization of the discriminator, we used Adam and set $(\alpha_1,\alpha_2)=(0.5,0.9)$ with a constant learning rate of 0.1. The total number of epochs was set to 2000.

\begin{table}[htbp]
 \caption{Model structure of Cluster-GAN and Cluster-EBGAN for two-circle data}
 \label{2-circle-structure}
 \centering
 \begin{tabular}{ccc} \toprule

Generator & Encoder & Discriminator \\ \midrule
     FC 20 LReLU &FC 20 LReLU &FC 30 LReLU  \\ \midrule
  FC 20 LReLU  & FC 20 LReLU & FC 30 LReLU \\ \midrule
 FC 2 linear Tanh & FC 5 linear & FC 1 linear\\
 \bottomrule
    \end{tabular}
\end{table}

\paragraph{Iris} For the iris data, we used a simple feed-forward network structure for Cluster GAN and Cluster EBGAN. We set the dimension of $z_n$ to 20,  set $(\beta_n,\beta_c)=(10,10)$, set the mini-batch size to 32, and set a constant learning rate of 0.01 for the generator with $\tau=1$. For optimization of the discriminator, we used Adam and set $(\alpha_1,\alpha_2)=(0.5,0.9)$ with a learning rate of 0.0001. The hyperparameters of Cluster-GAN is set to the default values.  

\begin{table}[htbp]
 \caption{Model structure of Cluster-GAN and Cluster-EBGAN for Iris}
 \label{Seeds-structure}
 \centering
 \begin{tabular}{ccc} \toprule
Generator & Encoder & Discriminator \\ \midrule
     FC 5 LReLU &FC 5 LReLU &FC 5 LReLU  \\ \midrule
  FC 5 LReLU  & FC 5 LReLU & FC 5 LReLU \\ \midrule
 FC 4 linear Sigmoid & FC 23 linear & FC 1 linear\\
 \bottomrule
    \end{tabular}
\end{table}

\paragraph{Seeds} For the seeds data, we used a  simple feed-forward network structure for Cluster-GAN and Cluster-EBGAN. We set the dimension of $z_n$ to 20, set $(\beta_n,\beta_c)=(5,5)$,  set the mini-batch size to 128, and set a constant learning rate of 0.01 for generator with $\tau=0.0001$. For optimization of the discriminator, we used Adam and set $(\alpha_1,\alpha_2)=(0.5,0.9)$ with a learning rate of  0.005. The hyperparameters of Cluster-GAN is set to the default values.  

\begin{table}[htbp]
 \caption{Model structure of Cluster-GAN and Cluster-EBGAN for Seeds}
 \label{iris-structure}
 \centering
 \begin{tabular}{ccc} \toprule
Generator & Encoder & Discriminator \\ \midrule
     FC 20 LReLU &FC 20 LReLU &FC 100 LReLU  \\ \midrule
  FC 20 LReLU  & FC 20 LReLU & FC 100 LReLU \\ \midrule
 FC 7 linear Tanh & FC 23 linear & FC 1 linear\\
 \bottomrule
    \end{tabular}
\end{table}

\paragraph{MNIST}

For Cluster GAN, our implementation is based on the code given at \url{https://github.com/zhampel/clusterGAN}, with a a small modification on Encoder. 
The Structures of the generator, encoder and discriminator are given as follow. Cluster GAN was run with the same parameter setting as given in the original work \cite{Mukher19clustergan}. 

\begin{table}[htbp]
  \caption{Model structure of ClusterGAN for MNIST data}
  \centering
  \begin{tabular}{ccc} \toprule
 Generator & Encoder & Discriminator \\ \midrule
    FC 1024 ReLU BN & 4$\times$ 4 conv, 64 stride 2 LReLU &  4$\times$ 4 conv, 64 stride 2 LReLU \\ \midrule
     FC $7\times 7 \times 128$ ReLU BN & 4$\times$ 4 conv, 128 stride 2 LReLU &  4$\times$ 4 conv, 64 stride 2 LReLU \\ \midrule
   $4\times 4$ upconv 64 stride 2 ReLU BN & 4$\times$ 4 conv, 256 stride 2 LReLU& FC1024 LReLU  \\ \midrule
   $4\times 4$ upconv 1 stride 2 Sigmoid  & FC 1024 LReLU & FC 1 linear  \\ \midrule
        & FC 40 &  \\
  \bottomrule
    \end{tabular}
\end{table}

For Cluster EBGAN, to accelerate  computation, we used the  parameter sharing strategy as in \cite{hoang2018mgan}, where all generators share the parameters except for the first layer. We set the dimension of $z_n$ to 5,    $(c_1,c_2,\eta_1,\alpha_1)=(40,10000,0.75,0.9)$, set the  mini-batch size to 100, and set a constant learning rate of $0.005$ for the generator.  For the functions $\phi_1$, $\phi_2$ and $\phi_3$, the non-saturating settings were adopted, i.e., 
 we set  $(\phi_1,\phi_2,\phi_3)=(\log x, \log(1-x),\log x)$.


\begin{table}[htbp]
  \caption{Model structure of ClusterEBGAN for MNIST simulation}
  \label{10modelst}
  \centering
  \begin{tabular}{ccc} \toprule

 Generator & Encoder & Discriminator \\ \midrule
     4$\times$ 4 conv, 512 stride 2 ReLU & 4$\times$ 4 conv, 64 stride 2 LReLU &  4$\times$ 4 conv, 64 stride 2 LReLU \\ \midrule
     3$\times$ 3 conv, 128 stride 2 pad 1 ReLU & 4$\times$ 4 conv, 128 stride 2 LReLU &  4$\times$ 4 conv, 64 stride 2 LReLU \\ \midrule
   4$\times$ 4 conv, 64 stride 2 pad 1 ReLU & 4$\times$ 4 conv, 256 stride 2 LReLU& FC1024 LReLU  \\ \midrule
   $4\times 4$ upconv 1 stride 2 pad 1 Sigmoid  & FC 1024 LReLU & FC 1 linear  \\ \midrule
        & FC 30 &  \\

  \bottomrule
    \end{tabular}
\end{table}

\subsection{Image Generation} \label{hmnistsetup}

\paragraph{HMNIST}
The discriminator has the architecture $[(3\times32\times 32),(7\times 32 \times 32)] \rightarrow 128\times16\times16 \rightarrow 256\times8\times8 \rightarrow 512\times4\times4 \rightarrow 1\times 1 \times 1$, where ``7 units'' in the input layer are reserved for the labels. The generator has the architecture $[(100\times1\times 1),(7\times 1\times 1)]\rightarrow 1024\times4\times4 \rightarrow 512\times8\times8 \rightarrow 256\times16\times16 \rightarrow 3\times32\times 32$. Remark that kernel size and stride are set to (4,1) for first deconvolutional layer and last convolutional layer. Other than that, (4,2) are set to all (de)convolutional layers. 
 
EBGAN was run for 50,000 iterations with $J_g=1$ and a mini-batch size of 128.
The other methods were also run for 50,000 iterations with the same mini-batch size 128. The learning rates and other parameters were given in Table \ref{partab6}.  \textcolor{black}{Again, the non-saturating setting was adopted with $(\phi_1,\phi_2,\phi_3)=(\log x, \log(1-x),\log x)$.    }
 
For prediction evaluation, a CNN with the structure $3\times32\times 32\rightarrow 128\times16\times16 \rightarrow 256\times8\times8 \rightarrow 512\times4\times4 \rightarrow 7\times1\times 1$ was used. The CNN was training using SGD with the momentum argument$=0.9$ and the learning rate decay argument$=5e-5$. The SGD was run for 50,000 iterations with the initial learning rate being set to 0.1, multiplying 0.1 by every 17,000 iteration, and the mini-batch size being set to 128. The networks learned at every 100 iterations were recorded, and the last 5 recorded networks were used for prediction in an ensemble averaging manner.

\begin{table}[htbp]
\caption{\label{partab6} Parameter settings for HMNIST data}
\begin{center}
\begin{tabular}{lcccccc}
\toprule
Method & \multicolumn{2}{c}{Learning rate} & $\alpha_1$ & $\alpha_2$ & $\rho$ \\ \cline{2-3}
& Discriminator & Generator  \\
\midrule
DCGAN & 0.0002 &  0.0002 & 0.5& 0.999  \\
BGAN & 0.0002 & 0.0002 & 0.5&\\
ProbGAN & 0.0002     &  0.0002      &   0.5 &  \\
EBGAN &  $(c_1,c_2,\alpha)=(5,5000,0.75)$& 0.0002 & 0.9 & &1\\
\bottomrule
\end{tabular}
\end{center}
\end{table}

\bibliographystyle{asa}
\bibliography{reference}